\documentclass[sigconf]{acmart}

\acmConference[MobiHoc 2025]{The 26th ACM International Symposium on Mobile Ad Hoc Networking and Computing}{June 2025}{Houston, USA}
\acmBooktitle{Proceedings of the 26th ACM International Symposium on Mobile Ad Hoc Networking and Computing (MobiHoc '25)}
\acmDOI{XX.XXXX/XXXXXXX.XXXXXXX}
\acmISBN{XXX-X-XXXX-XXXX-X}

\usepackage{microtype}
\usepackage{graphicx}
\usepackage{subfig}
\usepackage{booktabs}
\usepackage{url}
\usepackage{xcolor}
\usepackage{multirow}

\usepackage[skip=0pt]{caption}
\usepackage{hyperref}

\usepackage[capitalize,noabbrev]{cleveref}

\usepackage{algorithm}
\usepackage{algorithmic}

\newtheorem{theorem}{Theorem}
\newtheorem{assumption}{Assumption}
\newtheorem*{remark}{Remark}
\newtheorem{lemma}{Lemma}

\usepackage{xspace}

\newcommand{\alg}{\textsf{DPMixSGD}\xspace}
\newcommand{\algns}{\textsf{DPMixSGD}}

\usepackage{color, colortbl}

\definecolor{greyL}{RGB}{230,248,255}

\begin{document}

\title{Enhancing Privacy in Decentralized Min-Max Optimization: \\ A Differentially Private Approach}

\author{Yueyang Quan}
\affiliation{
	\institution{University of North Texas}
	\city{Denton}
        \country{USA}
}

\author{Chang Wang}
\authornote{Chang Wang conducted this research while he was an intern under the supervision of Zhuqing Liu.}
\affiliation{
	\institution{University of Nevada, Las Vegas}
	\city{Las Vegas}
        \country{USA}
}

\author{Shengjie Zhai}
\affiliation{
	\institution{University of Nevada, Las Vegas}
	\city{Las Vegas}
        \country{USA}
}

\author{Minghong Fang}
\affiliation{
	\institution{University of Louisville}
	\city{Louisville}
         \country{USA}
}

\author{Zhuqing Liu}
\affiliation{
	\institution{University of North Texas}
	\city{Denton}
        \country{USA}
}

\renewcommand{\shortauthors}{Yueyang Quan et al.}

\begin{abstract}
Decentralized min-max optimization allows multi-agent systems to collaboratively solve global min-max optimization problems by facilitating the exchange of model updates among neighboring agents, eliminating the need for a central server.
However, sharing model updates in such systems carry a risk of exposing sensitive data to inference attacks, raising significant privacy concerns. To mitigate these privacy risks, differential privacy (DP) has become a widely adopted technique for safeguarding individual data.
Despite its advantages, implementing DP in decentralized min-max optimization poses challenges, as the added noise can hinder convergence, particularly in non-convex scenarios with complex agent interactions in min-max optimization problems. In this work, we propose an algorithm called \alg (\underline{D}ifferential \underline{P}rivate \underline{Mi}nma\underline{x} Hybrid \underline{S}tochastic \underline{G}radient \underline{D}escent), a novel privacy-preserving algorithm specifically designed for non-convex decentralized min-max optimization. Our method builds on the state-of-the-art STORM-based algorithm, one of the fastest decentralized min-max solutions. We rigorously prove that the noise added to local gradients does not significantly compromise convergence performance, and we provide theoretical bounds to ensure privacy guarantees. To validate our theoretical findings, we conduct extensive experiments across various tasks and models, demonstrating the effectiveness of our approach.

\end{abstract}



\maketitle

% !TEX root = main.tex

\section{Introduction}\label{sec:intro}

Min-max optimization has been widely applied in various machine learning (ML) domains, such as in-context learning \citep{kim2024transformers,lee2024supervised}, generative adversarial networks (GANs) \citep{goodfellow2014generative,gulrajani2017improved,pmlr-v125-lin20a}, and adversarial reinforcement learning \citep{wai2019variance,han2022solution,he2023robust}.
Traditionally, ML models have been trained using high-performance clusters within large data centers. However, the growing range of ML applications has led to an increasing shift toward deploying models on edge computing networks. This change is driven by the need to process data from geographically distributed sources (e.g., smart devices, vehicles, and sensors) and the high costs or impracticality of transmitting raw training data to centralized cloud servers due to communication bandwidth limitations or privacy concerns \citep{nedic2009distributed,lian2017can}.
This paradigm is particularly beneficial in scenarios such as multi-agent pretraining and fine-tuning of large language models (LLMs) \citep{Poon2021SmoothBilevel, hashemi2024cobocollaborativelearningbilevel,Liu2021BOML,fang2020local,cao2020fltrust,fang2024byzantine}, where collaborative efforts are needed due to the sensitivity of fine-tuning data. 
It also benefits decentralized min-max optimization applications, such as decentralized AUC maximization \citep{gao2024decentralized}, multi-agent meta-learning \citep{Rajeswaran2019MetaLearning,Liu2021BOML}, and multi-agent reinforcement learning \citep{zhang2021taming,lu2022stochastic}, as the decentralized framework reduces communication overhead and mitigates privacy risks by limiting direct data sharing among agents.

Despite the perceived privacy advantages of decentralized learning, which limit direct data sharing among agents, recent research has shown that it remains vulnerable to privacy breaches due to indirect leakage through model updates or gradient information \citep{yang2019adversary,kasyap2021privacy,hallaji2024decentralized,regan2022federated}. An attacker can exploit shared model updates to infer sensitive information from agents, and in some cases, even reconstruct the original training data \citep{nasr2019comprehensive,hu2021source,fu2022label}. 
These vulnerabilities introduce substantial privacy risks, compromising the anticipated benefits of decentralized systems. To mitigate these challenges, researchers have employed differential privacy (DP)~\citep{cheng2019towards, lin2022towards,cyffers2022privacy,biswas2024noiseless}, a method that strengthens privacy by adding strategically designed noise to local updates before sharing, thereby offering increased protection against data breaches.
While existing works primarily focus on incorporating DP into centralized learning or standard decentralized frameworks such as federated learning, applying DP to decentralized stochastic min-max optimization remains largely unexplored and presents unique challenges:

\begin{list}{\labelitemi}{\leftmargin=1em \itemindent=0em \itemsep=.2em  \topsep=5pt
  \partopsep=1pt
  \parsep=0pt}

\item Adding noise for differential privacy in decentralized min-max optimization introduces randomness that degrades gradient accuracy, slowing convergence and destabilizing the optimization. In such problems, even slight noise can disrupt the delicate min-max balance and cause oscillations near saddle points. Privacy noise thus poses unique challenges by destabilizing complex saddle-point dynamics, complicating convergence in decentralized, privacy-preserving settings.

\item Decentralized min-max optimization faces additional difficulty due to non-IID data across agents, causing local gradients to diverge and hindering consensus on saddle points. Adding privacy noise worsens this by obscuring useful signals. The heterogeneity of non-IID data intensifies coordination challenges while preserving privacy.

\item Privacy analysis is especially challenging in decentralized min-max setups because each agent adds noise locally and communicates iteratively, complicating cumulative privacy accounting. Ensuring rigorous differential privacy without harming convergence remains difficult.

%\item \textcolor{blue}{Beyond privacy budget tracking, managing distributed loss across the network is crucial. Iterative multi-agent updates make controlling accumulated errors and steady progress toward a global saddle point challenging, especially with privacy noise and data heterogeneity.}

\end{list}

In this paper, we bridge a critical gap by introducing \alg, an innovative and efficient algorithm for differentially private decentralized min-max optimization. Our method leverages the STORM framework~\citep{cutkosky2019momentum} to reduce gradient variance, crucial for controlling noise under differential privacy. Its single-loop design eases implementation and privacy analysis. Unlike prior non-private decentralized min-max uses~\citep{xian2021faster}, we adapt STORM with privacy-preserving updates and a noise-aware convergence proof.
The core mechanism of \alg involves each agent perturbing its gradients with carefully calibrated noise to ensure differential privacy. These perturbed gradients are then shared with neighboring agents, enabling decentralized collaboration while maintaining privacy. To demonstrate the effectiveness of our \alg, we conduct a thorough convergence analysis and assess the privacy guarantees of \alg under practical and reasonable assumptions.

Our key contributions are as follows:

\begin{list}{\labelitemi}{\leftmargin=1em \itemindent=0em \itemsep=.2em  \topsep=5pt
  \partopsep=1pt
  \parsep=0pt}

\item We introduce \alg, a new algorithm that guarantees differential privacy in non-convex-strongly-concave decentralized min-max optimization. \alg is built upon STORM-based algorithms tailored for min-max problems, providing robust privacy protection while maintaining high optimization performance.

\item 
We establish rigorous theoretical 
convergence guarantee and privacy guarantees for our proposed algorithm, \alg.
Our proof shows that  even with Gaussian noise added to the communication process of local gradients, \alg can still maintain strict convergence.
 Meanwhile, by strategically designing the noise added to the communication of local gradients in decentralized min-max optimization framework, we achieve DP without significantly degrading the algorithm’s performance. This approach effectively balances privacy preservation with strong optimization results.

\item We empirically evaluate the \alg algorithm on logistic regression and AUROC min-max optimization tasks. 
To assess its performance, we compare \alg with several state-of-the-art methods, including DM-HSGD~\cite{xian2021faster}, SGDA~\cite{beznosikov2023stochastic}, and DP-SGDA~\cite{yang2022differentially}. 
The results show that \alg performs robustly, naturally preserving privacy while achieving results on par with other methods. Moreover, we conduct a comparative experiment between our algorithm and DM-HSGD to demonstrate that our method
significantly improves privacy robustness against Deep Leakage from Gradients (DLG) attacks.

\end{list}

% !TEX root = main.tex

\section{Related works} \label{sec:relatedworks}

\subsection{Decentralized min-max optimization}

Numerous methods have been proposed to address min-max optimization problems, including gradient descent techniques \citep{xu2024decentralized}, momentum-based approaches \citep{barazandeh2021decentralized, huang2023near}, mirror descent ascent methods \citep{huang2021efficient}, and stochastic gradient methods \citep{chen2024efficient, gao2022decentralized, luo2020stochastic}. Building upon these foundational techniques, various algorithms have been developed specifically for decentralized min-max optimization, such as those in \citep{koppel2015saddle, mateos2015distributed, liu2019decentralized, tsaknakis2020decentralized}. Our approach draws primarily from the decentralized min-max hybrid stochastic gradient descent (DM-HSGD) algorithm~\citep{xian2021faster}, which achieves a stochastic first-order oracle (SFO) complexity of \(\mathcal{O}(\kappa^3 \epsilon^{-3})\). Here, $\kappa = \frac{L}{\mu}$ denotes the condition number of the problem, defined as the ratio between the smoothness constant $L$ and the strong convexity constant $\mu$, and $\epsilon$ represents the target accuracy level for the optimization error. These developments have significantly expanded the applications of min-max optimization, especially in the context of machine learning.

\subsection{Differential privacy (DP)}
Differential Privacy (DP)~\citep{dwork2006differential} is a rigorous mathematical framework that ensures strong privacy guarantees when analyzing and sharing data. Many algorithms have been designed to provide these guarantees for minimization problems~\citep{wang2023decentralized,chaudhuri2011differentially,wang2018empirical,zhang2022net}, and some have been further adapted to handle min-max problems~\citep{kang2022stability,zhang2022bring}. Several recent works have explored differential privacy (DP) in the context of variational inequalities and saddle point problems. Boob et al.~\citep{boob2024optimal} investigated DP stochastic variational inequalities and saddle point problems, achieving optimal weak gap guarantees; González et al.~\citep{gonzalez2024mirror} proposed DP mirror descent methods with nearly dimension-independent utility guarantees for stochastic saddle-point problems. However, their analysis is primarily limited to centralized settings and does not extend to decentralized regime. Bassily et al.~\citep{bassily2023differentially} refined strong gap analysis using recursive regularization techniques, but their methods require strong convexity and are not directly applicable to min-max formulations in decentralized environments. Zhou et al.~\citep{zhou2024differentially} addressed worst-group risk minimization through a stability-based lens, but did not consider interactive or game-theoretic settings such as saddle-point optimization. In the nonconvex-strongly-concave case, Zhao et al.~\citep{zhao2023differentially} introduced a DP temporal difference learning algorithm; nonetheless, their focus lies in reinforcement learning rather than generic decentralized optimization. Our algorithm focuses on addressing the current limitations of these prior works.

Privacy concerns are particularly prominent in distributed systems, such as federated learning~\citep{wei2020federated, hu2020personalized, truex2020ldp, adnan2022federated} and multi-party computation, where nodes often need to exchange sensitive information. The main challenge in these scenarios is to safeguard individual data privacy while ensuring effective model training and robust performance, especially in the face of potential threats from malicious actors that could compromise the integrity of the learning process. Unlike centralized settings, decentralized optimization lacks a trusted central authority and requires nodes to frequently communicate gradients or model updates over untrusted networks, significantly increasing privacy risks. For instance, adversaries can reconstruct training data from shared gradients, as demonstrated in~\citep{zhu2019deep}, making privacy preservation even more challenging in decentralized settings. To address this, it is crucial to ensure differential privacy (DP) at each node. While existing work such as DP-SGDA~\citep{yang2022differentially} has explored DP in centralized min-max optimization, the decentralized case remains largely unaddressed. In this paper, we propose a novel approach that enhances privacy by injecting noise directly into local gradients at each node in a decentralized setting. While DP-SGDA~\citep{yang2022differentially} ensures privacy in centralized min-max optimization through gradient perturbation, it is not directly applicable to decentralized scenarios due to its reliance on centralized data access and coordination. In contrast, our method introduces noise locally in a distributed network, eliminating the need for central coordination and enabling efficient and privacy-preserving updates even in multi-agent systems. This design is naturally compatible with decentralized architectures and leads to improved scalability and communication efficiency. Following the strategy in~\citep{cyffers2022muffliato}, our algorithm operates by exchanging model variables $\mathbf{x}_t$ and $\mathbf{y}_t$, which depend on gradients. However, instead of perturbing the variables themselves, we perturb the local gradients to preserve privacy while maintaining optimization performance. This design makes our method particularly suitable for privacy-sensitive decentralized applications, such as federated adversarial training~\cite{zizzo2020fat, shah2021adversarial}. We provide the Table \ref{table_comparision} to intuitively illustrate the differences between our algorithm and the baselines we use in our paper. Our work thus takes a critical step toward bridging the gap between differential privacy and decentralized min-max optimization.

\begin{table}[h]
\centering
\caption{Comparison of DPMixSGD with baselines.}
\resizebox{\columnwidth}{!}{%
\begin{tabular}{|l|c|c|c|}
\hline
\textbf{Method} & \textbf{DP Guarantee} & \textbf{Variance Reduction} & \textbf{Min-Max Setting} \\
\hline
\textbf{DM-HSGD} & $\times$ & \checkmark (STORM) & \checkmark \\
\textbf{SGDA}  & $\times$ & $\times$ & \checkmark \\
\textbf{DP-SGDA} & \checkmark & $\times$ & \checkmark \\
\textbf{DPMixSGD (ours)} & \checkmark & \checkmark (STORM) & \checkmark \\
\hline
\end{tabular}
}

\label{table_comparision}
\end{table}

% !TEX root = main.tex

\section{Problem Formulation and Motivation }\label{sec:probformulation}

\subsection{Problem Formulation}
Before presenting our problem formulation, we first introduce the mixing matrix $\mathbf{W}$, which represents the averaging weights in the communication network. The matrix $\mathbf{W}=\left\{\mathbf{w}_{i j}\right\} \in \mathbb{R}^{m \times m}$ is doubly stochastic and satisfies the following conditions:
\begin{equation}
\mathbf{W} \mathbf{1} = \mathbf{W}^{\top} \mathbf{1} = \mathbf{1}
\end{equation}
where $\mathbf{1}$ is an all-ones matrix, and $\mathbf{W}^{\top}$ is the transpose of $\mathbf{W}$.
Note that in this paper $\mathbf{W}$ is required to be symmetric, allowing the communication network to represent undirected graphs.

Decentralized min-max problems are typically formulated in a multi-agent environment, where each agent has access only to its local data and collaborates with other agents via limited communication to optimize a non-convex strongly concave min-max objective function. The mathematical form of a decentralized min-max problem can be expressed as follows:
\begin{equation}
\begin{aligned}
\min _{\mathbf{x} \in \mathcal{X}} \max _{\mathbf{y} \in \mathcal{Y}} f(\mathbf{x, y}) = \frac{1}{m} \sum_{i=1}^m f_i(\mathbf{x, y}), \\
f_i(\mathbf{x, y}) := \mathbb{E}_{\mathbf{z}^{(i)} \sim D_i} F_i\left(\mathbf{x, y}; \mathbf{z}^{(i)}\right)
\end{aligned}
\end{equation}
where $m$ is the total number of agents; $\mathcal{X} \subseteq \mathbb{R}^{d_1}$ and $\mathcal{Y} \subseteq \mathbb{R}^{d_2}$ represent the decision spaces for $\mathbf{x}$ and $\mathbf{y}$, respectively; The local objective function $F_i(\mathbf{x, y}; \mathbf{z}^{(i)})$ is $L$-smooth, non-convex with respect to $\mathbf{x}$, and strongly concave with respect to $\mathbf{y}$;
$D_i$ denotes the data distribution on the $i$-th agent; $\mathbf{z}^{(i)}$ is a random vector sampled from the local dataset $\mathcal{Z}$.

In order to simplify the min-max problem, we often introduce $\Phi(\mathbf{x})=\max _{\mathbf{y} \in \mathcal{Y}} f(\mathbf{x}, \mathbf{y})$, reducing the problem to one that involves optimizing only over $\mathbf{x}$:
\begin{equation}
\min _{\mathbf{x} \in \mathcal{X}} \Phi(\mathbf{x})=\min _{\mathbf{x} \in \mathcal{X}} \max _{\mathbf{y} \in \mathcal{Y}} f(\mathbf{x}, \mathbf{y}).
\end{equation}
Additionally, we introduce noise to the local gradients. To establish the privacy guarantees of our algorithm, we now present the definition of differential privacy in the context of stochastic decentralized min-max problems.

\newcounter{definition}
\textbf{Definition \refstepcounter{definition}\label{definition 1}1.} [Differential Privacy~\citep{DworkRoth2014}]  
An algorithm $\mathcal{A}: \mathcal{Z}^n \rightarrow \mathbb{R}^{d_1}\times \mathbb{R}^{d_2}$ is said to be $(\theta, \gamma)$-differentially private if, for any adjacent datasets $\mathbf{z}^{(i)} \sim \mathbf{z}^{(i)\prime}$ (on the $i$-th agent) and for all output events $O \subseteq \operatorname{range}(\mathcal{A})$, the following holds:
\begin{equation}
\mathbb{P}\left[\mathcal{A}\left( \mathbf{z}^{(i)}\right) \in O\right] \leq e^\theta \mathbb{P}\left[\mathcal{A}\left( \mathbf{z}^{(i)\prime}\right) \in O\right] + \gamma,
\end{equation}
where $\mathcal{A}\left(\mathbf{z}^{(i)}\right)$ is the output of the decentralized algorithm based on $i$-th agent datasets $\mathbf{z}^{(i)}$. $\mathbb{P}$ denotes the probability of the algorithm's output in the corresponding event. Note that two datasets are said to be adjacent if they differ in at most one data sample.

Empirical risk plays a crucial role in differential privacy. By optimizing the empirical risk across all agents, the algorithm can effectively train a global model while protecting individual data privacy. Furthermore, empirical risk minimization helps evaluate the impact of privacy-preserving mechanisms on the overall system, ensuring that the algorithm can still converge correctly and produce a meaningful model even after the addition of noise. By combining STORM with gradient tracking, our algorithm effectively mitigates consensus errors and ensures convergence in decentralized settings with non-identical data. This makes it more robust and suitable for complex distributed scenarios. In this paper, we define the average empirical risk as the mean of the local gradients estimators $\bar{\mathbf{g}}_t$ across all agents, expressed as follows:
\begin{equation}
\begin{aligned}
&\nabla_\mathbf{x}f_S(\bar{\mathbf{x}},\bar{\mathbf{y}})  = \bar{\mathbf{g}}_t  = \frac{1}{m} \sum_{i=1}^{m} \nabla_{\mathbf{x}} F_i\left(\mathbf{x}_t^{(i)}, \mathbf{y}_t^{(i)};\mathbf{z}_t^{(i)}\right)  \\
&+ \left(1 - \beta_{\mathbf{x}}\right)\left(\bar{\mathbf{g}}_{t-1} - \frac{1}{m} \sum_{i=1}^{m} \nabla_{\mathbf{x}} F_i\left(\mathbf{x}_{t-1}^{(i)}, \mathbf{y}_{t-1}^{(i)};\mathbf{z}_t^{(i)}\right)\right)   
\end{aligned}
\end{equation}
The empirical risk is constructed by $F_i\left(\mathbf{x}_t^{(i)}, \mathbf{y}_t^{(i)}; \mathbf{z}^{(i)}\right)$, which is the local loss functions across all $m$ agents, where $\mathbf{z}^{(i)}$ denotes the data of the $i$-th agent.

\subsection{Motivating Applications}

Our motivation stems from concerns about privacy leaks in the real-world applications of decentralized learning. Here, we present two motivating applications to illustrate the practical relevance of our work:

\begin{list}{\labelitemi}{\leftmargin=1em \itemindent=0em \itemsep=.2em}

\item \textbf{Decentralized Min-Max Learning in Healthcare}: Decentralized learning is widely used in healthcare to enable collaborative model training across institutions without sharing sensitive patient data~\cite{shiranthika2023decentralized, kasyap2021privacy, tedeschini2022decentralized, kuo2018modelchain}. Each hospital trains a model locally using its own records and communicates model updates—such as gradients—with its neighbors. However, this process introduces privacy risks, as shared updates may leak confidential information. Beyond privacy, healthcare itself presents intrinsic min-max structures: resource allocation problems—such as distributing ICU beds, vaccines, or staff—often aim to optimize system-wide performance under limited capacity by minimizing the worst-case delay or maximizing the earliest service availability \cite{ninh2024stochastic}. Motivated by this, we propose a decentralized min-max optimization framework, where each hospital solves a local min-max problem that captures both learning objectives and operational constraints inherent to healthcare. To ensure patient confidentiality during collaboration, we incorporate DP by injecting calibrated noise into local updates. This mechanism prevents sensitive information from being inferred from shared gradients, enabling secure and privacy-preserving model training across institutions.
\item \textbf{Decentralized Min-Max Learning for Financial Systems}: In financial systems~\cite{zhou2023sok, ren2024lookahead}, institutions often need to collaboratively train models—for tasks like risk assessment or market forecasting—without sharing sensitive data. Many of these problems naturally follow a min-max structure, as each agent seeks to minimize risk or loss under worst-case scenarios or regulatory constraints. Since financial data is highly sensitive and regulated, differential privacy (DP) is critical to prevent leakage of proprietary or customer information through shared model updates. By introducing noise into local computations, DP enables institutions to collaborate securely without compromising data confidentiality.

\end{list}

% !TEX root = main.tex
\section{Solution Approach}\label{sec:algorithm}

In this section, we first outline the necessary preparations for the algorithm and then proceed to present the algorithm along with detailed explanations.

\subsection{Preliminaries}

Before detailing the proposed algorithms, we define the notations and key concepts used throughout this paper. Let $\mathbf{x}_t^{(i)}$ and $\mathbf{y}_t^{(i)}$ represent the column vector parameters on the $i$-th agent at $t$-th iteration. The matrices $X_t$ and $Y_t$ are defined by stacking the vectors $\mathbf{x}_t^{(i)}$ and $\mathbf{y}_t^{(i)}$ across all $m$ agents, i.e., $X_t = \left[\mathbf{x}_t^{(1)}, \mathbf{x}_t^{(2)}, \dots, \mathbf{x}_t^{(m)}\right], \quad Y_t = \left[\mathbf{y}_t^{(1)}, \mathbf{y}_t^{(2)}, \dots, \mathbf{y}_t^{(m)}\right].$ The gradient estimators $\mathbf{g}_t^{(i)}$ and $\mathbf{h}_t^{(i)}$ are the local gradient estimators for $\mathbf{x}$ and $\mathbf{y}$ at the $i$-th agent, respectively, while $\mathbf{v}_t^{(i)}$ and $\mathbf{u}_t^{(i)}$ denote their aggregated counterparts across the network. The matrices $G_t, H_t, V_t$, and $U_t$ are constructed by stacking the corresponding column vectors $\mathbf{g}_t^{(i)}, \mathbf{h}_t^{(i)}, \mathbf{v}_t^{(i)}$, and $\mathbf{u}_t^{(i)}$ from all agents. Additionally, the matrices $G_t^*$ and $H_t^*$ represent the gradient estimators with noise components included in $\mathbf{g}_t^{(i)}$ and $\mathbf{h}_t^{(i)}$, respectively.

For the mean vectors, we denote the lower-case variable with a bar to represent it, and the upper-case variables with a bar to represent matrices where each column is the corresponding mean vector. Specifically, the mean of $\mathbf{x}_t^{(i)}$ is given by $
\bar{\mathbf{x}}_t = \frac{1}{m} \sum_{i=1}^m \mathbf{x}_t^{(i)},$ and the  matrix $\bar{X}_t$ is defined as $ \bar{X}_t = \left[\bar{\mathbf{x}}_t, \bar{\mathbf{x}}_t, \dots, \bar{\mathbf{x}}_t\right].
$ Meanwhile, the added noise terms $n_{\mathbf{x},t}^{(i)} \sim \mathcal{N}(0, \sigma_{\mathbf{x}}^2 I_{d_1})$ and $n_{\mathbf{y},t}^{(i)} \sim \mathcal{N}(0, \sigma_{\mathbf{y}}^2 I_{d_2})$ for $\forall{i}$ are applied to the respective gradients. We define their mean values as follows:
\begin{equation}
\mathcal{N}_{\mathbf{x}, t}=\frac{\sum_{i=1}^m n_{\mathbf{x}, t}^{(i)}}{m}, \quad \mathcal{N}_{\mathbf{y}, t}=\frac{\sum_{i=1}^m n_{\mathbf{y}, t}^{(i)}}{m}.
\end{equation}
Next, we define the optimal solution for $\mathbf{y}$ as:
\begin{equation}
\mathbf{y}^*(\cdot) = \arg \max_{\mathbf{y} \in \mathcal{Y}} f(\cdot, \mathbf{y}),
\quad \hat{\mathbf{y}}_t = \arg \max_{\mathbf{y} \in \mathcal{Y}} f\left(\bar{\mathbf{x}}_t, \mathbf{y}\right), 
\end{equation}
where, under the condition that $f$ is strongly concave in $\mathbf{y}$, $\hat{\mathbf{y}}_t$ is unique. We further define the deviation as
\begin{equation}
\delta_t = \left\|\hat{\mathbf{y}}_t - \bar{\mathbf{y}}_t\right\|^2.
\end{equation}

The vectors $\mathbf{0}$ and $\mathbf{1}$ denote $m \times 1$ column vectors of all zeros and ones, respectively. The frobenius norm is denoted by $\|\cdot\|_F$, and the spectral norm by $\|\cdot\|_2$. Partial derivatives with respect to $\mathbf{x}$ and $\mathbf{y}$ are represented by $\nabla_{\mathbf{x}}$ and $\nabla_{\mathbf{y}}$.
\begin{algorithm}[htbp]
		\caption{\alg on the $i$-th agent. \label{algorithm}}
		\label{alg:AOA}
		\renewcommand{\algorithmicrequire}{\textbf{Initialize:}}
		\renewcommand{\algorithmicensure}{\textbf{Output:}}
		\begin{algorithmic}[1]
			\REQUIRE  Mixing\ matrix $\mathbf{W}$, initial  value $\mathbf{x}_0^{(i)}=\mathbf{x}_0, \mathbf{y}_0^{(i)}=\mathbf{y}_0, \mathbf{v}_{-1}^{(i)}=\mathbf{g}_{-1}^{(i)*}=\mathbf{0}, \mathbf{u}_{-1}^{(i)}=\mathbf{h}_{-1}^{(i)*}=\mathbf{0}$, when the algorithm can reach the optimal solution with 0 iteration. We set $\mathbf{g}_0^{(i)}=\nabla_\mathbf{x} F_i\left(\mathbf{x}_0^{(i)}, \mathbf{y}_0^{(i)} ; \mathbf{z}_0^{(i)}\right)$ and $\mathbf{h}_0^{(i)}=\nabla_\mathbf{y} F_i\left(\mathbf{x}_0^{(i)}, \mathbf{y}_0^{(i)} ; \mathbf{z}_0^{(i)}\right), \quad\left|\mathbf{z}_0^{(i)}\right|=b_0$.
			
			\hspace*{-\leftmargin}\textbf{Parameter: } Privacy budgets $\theta, \gamma$, learning rate $\eta_\mathbf{x}, \eta_\mathbf{y}$, weight  $\beta_\mathbf{x}, \beta_\mathbf{y}$, batch size $b_0$, epoch $T$.
			
			\ENSURE $\bar{x}_\zeta$, where $\zeta$ is chosen randomly from $\{1,2, \cdots, T\}$  %%output
			
			\FOR{each $t = 1, \dots, T-1$ }
			\STATE \hypertarget{line2}
            $\begin{aligned}
             \mathbf{g}_t^{(i)} = & \left(1 - \beta_\mathbf{x}\right) \left(\mathbf{g}_{t-1}^{(i)} - \nabla_\mathbf{x} F_i\left(\mathbf{x}_{t-1}^{(i)}, \mathbf{y}_{t-1}^{(i)} ; \mathbf{z}_t^{(i)}\right)\right)+\\& \quad \nabla_\mathbf{x} F_i\left(\mathbf{x}_t^{(i)}, \mathbf{y}_t^{(i)} ; \mathbf{z}_t^{(i)}\right)   
            \end{aligned}$ 
            \STATE \hypertarget{line3}
            $\begin{aligned}\mathbf{h}_t^{(i)} = &\left(1 - \beta_\mathbf{y}\right) \left(\mathbf{h}_{t-1}^{(i)} - \nabla_\mathbf{y} F_i\left(\mathbf{x}_{t-1}^{(i)}, \mathbf{y}_{t-1}^{(i)} ; \mathbf{z}_t^{(i)}\right)\right)+\\&\quad \nabla_\mathbf{y} F_i\left(\mathbf{x}_t^{(i)}, \mathbf{y}_t^{(i)} ; \mathbf{z}_t^{(i)}\right)
            \end{aligned}$ 
			\STATE   {//Encrypt gradients when communicating with other agents.}
           \STATE \hypertarget{line5} Sample noise $n_{\mathbf{x},t}^{(i)}\sim \mathcal{N}(0, \sigma_\mathbf{x}^2 I_{d_1})$ and $n_{\mathbf{y},t}^{(i)} \sim \mathcal{N}(0, \sigma_\mathbf{y}^2 I_{d_2})$.

			\STATE $\mathbf{g}_t^{(i)*} = \mathbf{g}_t^{(i)} +  n_{\mathbf{x},t}^{(i)}$
			\STATE \hypertarget{line7}$\mathbf{h}_t^{(i)*} = \mathbf{h}_t^{(i)} +  n_{\mathbf{y},t}^{(i)}$
			\STATE   {//At the $i$-th agent, the encrypted gradient is received and calculated.}
			\STATE \hypertarget{line9}$\mathbf{v}_t^{(i)}=\sum_{j=1}^m w_{ij}\left(\mathbf{v}_{t-1}^{(j)}+\mathbf{g}_t^{(j)*}-\mathbf{g}_{t-1}^{(j)*}\right)$
			\STATE \hypertarget{line10}$\mathbf{u}_t^{(i)}=\sum_{j=1}^m w_{ij}\left(\mathbf{u}_{t-1}^{(j)}+\mathbf{h}_t^{(j)*}-\mathbf{h}_{t-1}^{(j)*}\right)$
			\STATE   {//Send the computation result to the respective agent and perform the mixed information calculation at that agent.} 
			\STATE $\mathbf{x}_{t+1}^{(i)}=\sum_{j=1}^m w_{i j}\left(\mathbf{x}_t^{(j)}-\eta_\mathbf{x} \mathbf{v}_t^{(j)}\right)$\\
			\STATE $\mathbf{y}_{t+1}^{(i)}=\sum_{j=1}^m w_{i j}\left(\mathbf{y}_t^{(j)}+\eta_\mathbf{y} \mathbf{u}_t^{(j)}\right)$\\
			\ENDFOR
		\end{algorithmic}
	\end{algorithm}

\subsection{DP in decentralized min-max problem}
In this subsection, we will explain our new algorithm step by step. The overall procedure is similar to the STORM-based algorithm, however, it is important to note that we introduce gradient perturbation in the algorithm to ensure privacy protection.

The original values of the parameters at all agents are set to be identical, that is, $\mathbf{x}_0^{(i)} = \mathbf{x}_0$ and $\mathbf{y}_0^{(i)} = \mathbf{y}_0$ for every agent $i$. The quantities $\mathbf{g}_t^{(i)}$ and $\mathbf{h}_t^{(i)}$ represent the gradient estimators at the $i$-th agent with respect to $\mathbf{x}$ and $\mathbf{y}$, respectively. These estimators are computed following the STORM~\citep{cutkosky2019momentum} method used on DM-HSGD~\citep{xian2021faster}. Specifically, at $t=0$, a large batch size of $b_0$ is used to estimate the stochastic gradient (see Initialize in Algorithm \ref{algorithm}). For $t > 0$, the gradient estimators can be computed using either a single data point or a mini-batch (refer to lines \hyperlink{line2}{2} and \hyperlink{line3}{3} in Algorithm \ref{algorithm}).

The update rule for the gradient estimator $\mathbf{g}_t^{(i)}$ across all agents can be expressed as follows:
\begin{equation}
\begin{aligned}
&\bar{\mathbf{g}}_t = \frac{1}{m} \sum_{i=1}^{m} \nabla_{\mathbf{x}} F_i\left(\mathbf{x}_t^{(i)}, \mathbf{y}_t^{(i)}, \mathbf{z}_t^{(i)}\right)\\
&+ \left(1 - \beta_{\mathbf{x}}\right)\left(\bar{\mathbf{g}}_{t-1} - \frac{1}{m} \sum_{i=1}^{m} \nabla_{\mathbf{x}} F_i\left(\mathbf{x}_{t-1}^{(i)}, \mathbf{y}_{t-1}^{(i)}, \mathbf{z}_t^{(i)}\right)\right)
\end{aligned}
\label{eq_contin}
\end{equation}
where the mean gradient $\bar{\mathbf{g}}_t$ is updated by combining the current and previous gradients weighted by $\beta_{\mathbf{x}}$.

Similarly, the computation for $\mathbf{h}_t^{(i)}$ follows the same procedure as for $\mathbf{g}_t^{(i)}$. Once the local gradient estimators $\mathbf{g}_t^{(i)}$ and $\mathbf{h}_t^{(i)}$ are computed, each agent communicates with its neighboring agents to aggregate the estimates and compute the new gradient estimators $\mathbf{u}_t^{(i)}$ and $\mathbf{v}_t^{(i)}$.
To ensure differential privacy, noise is added to the local gradients during communication with neighboring agents (see lines \hyperlink{line5}{5} to \hyperlink{line7}{7} in Algorithm \ref{algorithm}). This guarantees that our algorithm meets privacy requirements. To mitigate the consensus error, gradient tracking is employed (see lines \hyperlink{line9}{9} and \hyperlink{line10}{10} in Algorithm \ref{algorithm}).
After obtaining the updated gradient estimators $\mathbf{u}_t^{(i)}$ and $\mathbf{v}_t^{(i)}$, each agent communicates with its neighbors again to update the model parameters $\mathbf{x}$ and $\mathbf{y}$.  
% !TEX root = main.tex
\section{Theoretical Analysis}
\label{sec:theorem}

In this section, we present the convergence analysis and discuss the privacy guarantees of our \alg algorithm under certain mild assumptions. All relevant proofs are provided in Appendix. We begin by reviewing some essential assumptions and definitions.

\subsection{Convergence analysis}

In our proposed \alg, each agent introduces noise to the local gradients to ensure privacy during agent communication. Therefore, before presenting the privacy guarantees, we first provide a rigorous proof of convergence to demonstrate that the added noise does not affect the original algorithm's convergence. To support this proof, we introduce several mild assumptions.

\begin{assumption} 
\label{assumption 1}
\textit{(Lipschitz continuity of the gradient)} Each local function $F_i\left(\mathbf{x}, \mathbf{y} ; \mathbf{z}^{(i)}\right)$ is lipschitz smooth, meaning there exists a constant $L$ such that for any two pairs $(\mathbf{x}, \mathbf{y})$ and $\left(\mathbf{x}^{\prime}, \mathbf{y}^{\prime}\right)$, we have:
\begin{equation}
\begin{aligned}
&\left\|\nabla F_i(\mathbf{x, y} ; \mathbf{z}^{(i)})-\nabla F_i\left(\mathbf{x}^{\prime}, \mathbf{y}^{\prime} ; \mathbf{z}^{(i)}\right)\right\|^2 \\
&\leq L^2\left(\left\|\mathbf{x}-\mathbf{x}^{\prime}\right\|^2+\left\|\mathbf{y}-\mathbf{y}^{\prime}\right\|^2\right).
\end{aligned}
\end{equation}
\end{assumption}

\begin{assumption} 
\label{assumption 2}
\textit{(Bounded gradient variance)} The gradient of each local function $F_i\left(\mathbf{x}, \mathbf{y} ; \mathbf{z}^{(i)}\right)$ is an unbiased estimate of $\nabla f_i(\mathbf{x}, \mathbf{y})$ and has bounded variance, i.e.,
\begin{equation}
\begin{aligned}
\mathbb{E}\left\|\nabla F_i(\mathbf{x, y} ; \mathbf{z}^{(i)})-\nabla f_i(\mathbf{x, y})\right\|^2 \leq \sigma<+\infty.
\end{aligned}
\end{equation}
\end{assumption}

\begin{assumption} 
\label{assumption 3}
\textit{(Lower bound of the objective)} The global objective function $\Phi(\cdot)$ is lower bounded, i.e., $\inf _\mathbf{x} \Phi(\mathbf{x})=\Phi^*>-\infty$.
\end{assumption}

\begin{remark} 
All the aforementioned assumptions are standard assumptions in optimization analysis  \citep{khanduri2021stem, cutkosky2019momentum, karimireddy2020mime, zhang2022understanding, fang2018spider}.
\end{remark}

\begin{assumption} 
\label{assumption 4}
\textit{(Spectral gap of the mixing matrix)} The doubly stochastic mixing matrix $\mathbf{W}$ satisfies the following spectral gap condition: $\left\|\mathbf{W}-\frac{\mathbf{1 1}^{\top}}{n}\right\|_2=\lambda \in[0,1)$.
\end{assumption}

\begin{remark} 
The spectral gap assumption plays a crucial role in ensuring effective information transfer across the network, allowing each agent to achieve global convergence by communicating with its neighboring agents, as highlighted in prior works~\citep{xin2021hybrid, wang2023decentralized}. A typical spectral gap assumption requires the mixing matrix $\mathbf{W}$ to be symmetric and doubly stochastic, with eigenvalues $\lambda_1 \geq \lambda_2 \geq \dots \geq \lambda_n$ such that $\left|\lambda_2\right|<1$ and $\left|\lambda_n\right|<1$. This condition guarantees that the communication graph remains connected, preventing both excessive diffusion and slow propagation of information within the system.

We adopt this symmetric and doubly stochastic matrix setting because undirected graphs are standard and widely adopted in decentralized learning. In many practical scenarios, such as decentralized federated learning over peer-to-peer networks~\cite{behera2021federated}, sensor networks~\cite{giridhar2006toward}, or cooperative robotics systems~\cite{bullo2009distributed}, communication between agents is naturally bidirectional—each agent can both send and receive information from its neighbors. This symmetric communication structure simplifies the design and analysis of algorithms and has been shown to yield stable and efficient convergence in numerous studies. Furthermore, undirected graphs with symmetric weight matrices allow for well-established consensus-based protocols and spectral methods to be employed, making them a natural choice for studying theoretical properties such as convergence and privacy guarantees in decentralized settings.
\end{remark}

\begin{assumption} 
\label{assumption 5}
\textit{(Strong concavity)} The function $f_i(\mathbf{x}, \mathbf{y})$ is $\mu$-strongly concave in $\mathbf{y}$. That is, there exists a constant $\mu>0$ such that for any $\mathbf{x}, \mathbf{y}$ and $\mathbf{y}^{\prime}$, we have:
\begin{equation}
\begin{aligned}
f_i(\mathbf{x, y}) \leq f_i\left(\mathbf{x, y}^{\prime}\right)+\left\langle\nabla_\mathbf{y} f_i\left(\mathbf{x, y}^{\prime}\right), \mathbf{y}-\mathbf{y}^{\prime}\right\rangle  
-\frac{\mu}{2}\left\|\mathbf{y}-\mathbf{y}^{\prime}\right\|^2.
\end{aligned}
\end{equation}
\end{assumption}

\begin{remark} 
The assumption of strong concavity in $\mathbf{y}$ is crucial for ensuring the well-posedness of the min-max problem. Specifically, $\mu$-strong concavity ensures the uniqueness of the solution during the $\mathbf{y}$-update step, thereby preventing ambiguity in the optimization process. This assumption is standard in the analysis of min-max optimization problems and is essential for deriving theoretical guarantees related to convergence rates and stability. Many previous works in the field of decentralized min-max optimization \citep{chen2024efficient} adopt this assumption to enhance their algorithms' convergence, stability, and efficiency.
\end{remark}

Similar to standard nonconvex-strongly-concave problems, we continue to use the $\epsilon$-stationary point as the convergence criterion, i.e., $\|\nabla \Phi(\mathbf{x})\| \leq \epsilon$. From the Lemma 4.3 in \citep{lin2020gradient}, it is established that the function $\Phi(\mathbf{x})$ is differentiable and satisfies the $(L + \kappa L)$-smoothness condition. And it also mention that $\mathbf{y}^{*}(\cdot)$ is $\kappa$-Lipschitz continuous, meaning for any $\mathbf{x}_1, \mathbf{x}_2 \in \mathbb{R}^{d_1}$, the inequality $\left\|\mathbf{y}^{*}(\mathbf{x}_1) - \mathbf{y}^{*}(\mathbf{x}_2)\right\| \leq \kappa \left\|\mathbf{x}_1 - \mathbf{x}_2\right\|$ holds. This indicates that the variation of $\mathbf{y}(\cdot)$ is bounded by $\kappa$. Consequently, we have:
\begin{equation}
\nabla \Phi\left(\bar{\mathbf{x}}_t\right)=\nabla_\mathbf{x} f\left(\bar{\mathbf{x}}_t, \hat{\mathbf{y}}_t\right)+\nabla_\mathbf{y} f\left(\bar{\mathbf{x}}_t, \hat{\mathbf{y}}_t\right) \cdot \partial \mathbf{y}^*\left(\bar{\mathbf{x}}_t\right),
\end{equation}
where we use $\nabla_\mathbf{y} f\left(\bar{\mathbf{x}}_t, \hat{\mathbf{y}}_t\right)=0$ as defined earlier. Thus we have $\nabla \Phi\left(\bar{\mathbf{x}}_t\right)=\nabla_\mathbf{x} f\left(\bar{\mathbf{x}}_t, \hat{\mathbf{y}}_t\right)$. 
With this, we now present the main theorem that show our algorithm maintains convergence despite the added noise for privacy preservation.

\begin{theorem}
\label{theorem 1}
Let Assumptions \ref{assumption 1} to \ref{assumption 5} hold, our Algorithm~\ref{algorithm} satisfies
\begin{equation}
\label{eq14}
\begin{aligned}
&\frac{1}{T} \sum_{t=0}^{T-1} \mathbb{E}\left\|\nabla \Phi\left(\bar{\mathbf{x}}_t\right)\right\|^2 = \mathcal{O}\left(\epsilon^2\right)+\mathcal{O}\left(m\epsilon^2\right)+\mathcal{O}\left(\sigma_\mathbf{x}^2d_1+\sigma_\mathbf{y}^2d_2\right),\notag
\end{aligned}
\end{equation}
when we set $T=\frac{1500 \kappa^3 }{(1-\lambda)^2 \epsilon\beta_\mathbf{x}}$ and the other parameters satisfy $\beta_\mathbf{y}=\frac{\beta_\mathbf{x}}{25\kappa^2}$, $ \eta_\mathbf{x}=\frac{(1-\lambda)^2 \beta_\mathbf{x}}{750 \kappa^3 L\epsilon}$, $\eta_\mathbf{y}=\frac{(1-\lambda)^2 \beta_\mathbf{x}}{75 \kappa L\epsilon}$, $ b_0=\frac{20 \kappa\epsilon}{\beta_\mathbf{x}}$, $\beta_x=\frac{\epsilon \min \{1, m \epsilon\}}{20}$. And we have,
\begin{equation}
\label{eq15}
\begin{aligned}
&\frac{1}{T} \sum_{t=0}^{T-1} \mathbb{E}\left\|\nabla \Phi\left(\bar{\mathbf{x}}_t\right)\right\|^2 =\mathcal{O}\left(\frac{1}{\left(m T_0\right)^{2 / 3}}\right)+\mathcal{O}\left(\frac{1}{T_0}\right)\\
&+\mathcal{O}\left(\frac{m^{1/3}}{T_0^{2/3}}\right)+\mathcal{O}\left(\sigma_\mathbf{x}^2d_1+\sigma_\mathbf{y}^2d_2\right), 
\end{aligned}
\end{equation}
when we set $T = \frac{30000\kappa^3 T_0}{(1-\lambda)^2}$, and the other parameters satisfy $\beta_\mathbf{y}=\frac{\beta_\mathbf{x}}{25\kappa^2}$, $ \eta_\mathbf{x}=\frac{(1-\lambda)^2 \beta_\mathbf{x}}{750 \kappa^3 L\epsilon}$, $\eta_\mathbf{y}=\frac{(1-\lambda)^2 \beta_\mathbf{x}}{75 \kappa L\epsilon}$, $ b_0=\frac{20 \kappa\epsilon}{\beta_\mathbf{x}}$, $T_0 \geq 10m^2$, and $\beta_x=\frac{m^{1 / 3}}{20 T_0^{2 / 3}}$.
 \end{theorem}

\begin{remark} 
We build upon the convergence analysis from Xian's work~\citep{xian2021faster}, however, by introducing noise into the local gradients during communication with neighboring agents, additional terms, namely $n_{\mathbf{x},t}^{(i)}$ and $n_{\mathbf{y},t}^{(i)}$, are introduced into $\mathbf{g}_t$ in our analysis. These additional terms are amplified during the proof process and require adjustments to the entire proof. As a result, we recalculated the bounds for all theorems and lemmas involved. To simplify our results, we applied novel scaling and bounding techniques. (e.g., for $\frac{1}{T} \sum_{t=0}^{T-1} \mathbb{E}\left\|\nabla \Phi\left(\bar{\mathbf{x}}_t\right)\right\|^2 $, we adopt a similar approach and found that our results, including the additional terms, are no greater than twice the original results in Xian's work. Since this already provides a tight bound, we opt to double some constants in the related terms for simplicity.) From our results, we conclude that the added noise does not impact the SFO complexity of the DM-HSGD algorithm. Specifically, when $T$ is determined by $\epsilon$ which is shown in Eq.~(\ref{eq14}), if $m \leq O\left(\epsilon^{-1}\right)$, the SFO complexity of Algorithm \ref{algorithm} is $O\left(\kappa^3 \epsilon^{-3}\right)$. For $m > O\left(\epsilon^{-1}\right)$, the SFO complexity is $O\left(\kappa^3 m \epsilon^{-2}\right)$. When $T$ is independent of $\epsilon$ as we show in Eq.~(\ref{eq15}), the leading term in the convergence rate remains $O\left(\frac{1}{\left(m T_0\right)^{2 / 3}}\right)$, thus preserving the linear speedup characteristic of the original algorithm.
Detailed proof of Theorem~\ref{theorem 1} is provided in Appendix~\ref{sec:apendixa}.
\end{remark}

\subsection{Privacy analysis}

\begin{assumption} 
\label{assumption 6} 
\citep{kang2022stability} For the min-max problem, we say $f$ is $\rho$-strongly-convex-strongly-concave ( $\rho-S C-S C$ ) if for each fixed $\mathbf{y} \in \mathcal{Y}$, the function $f_i(\mathbf{x, y}; \cdot)$ is $\rho$-strongly-convex in $\mathbf{x}$ for all $i$. And for each fixed $\mathbf{x} \in \mathcal{X}$, the function $f_i(\mathbf{x, y} ; \cdot)$ is $\rho$-strongly-concave in $\mathbf{y}$ for all $i$. In this paper, we focus on the $\rho-S C-S C$ problem.
\end{assumption}

\begin{remark}
While this assumption may appear restrictive, it captures a number of important practical
scenarios where strong convexity can be induced through regularization. Examples include robust
federated learning, adversarial training, and resource allocation. We will clarify this in the revised
version and explicitly mention it as a key direction for future work, to either relax this assumption
or extend our analysis to broader settings.
\end{remark}

\begin{assumption}[Bounded Gradient] 
\label{assumption 7} 
There exists a constant $L_g>0$ such that, for any $\mathbf{x}, \mathbf{y}$ and $\mathbf{z}$,
\begin{equation}
\left\|\nabla_{\mathbf{x}} F_i(\mathbf{x}, \mathbf{y} ; \mathbf{z})\right\|_2 \leq L_g, \quad
\left\|\nabla_{\mathbf{y}} F_i(\mathbf{x}, \mathbf{y} ; \mathbf{z})\right\|_2 \leq L_g.   
\end{equation}
\end{assumption}

Now let's review the proof for convergence part (Appendix \ref{sec:apendixa}), we have already known that $\bar{\mathbf{v}}_t=\bar{\mathbf{g}}_t=\left(1-\beta_\mathbf{x}\right)\left(\bar{\mathbf{g}}_{t-1}-\frac{1}{m} \right.$
$\left.\sum_{i=1}^m \nabla_\mathbf{x} F_i \left(\mathbf{x}_{t-1}^{(i)}, \mathbf{y}_{t-1}^{(i)}, \mathbf{z}_t^{(i)}\right)\right)+\frac{1}{m} \sum_{i=1}^m 
\nabla_\mathbf{x} F_i\left(\mathbf{x}_t^{(i)}, \mathbf{y}_t^{(i)}, \mathbf{z}_t^{(i)}\right)$. 
And by the definition of $\mathbf{g}_t^{(i)}$, we can obtain this recursively:
\begin{equation}
\begin{aligned}
&\bar{\mathbf{g}}_t=\frac{1}{m}  \sum_{j=1}^{m}\bigg(\sum_{k=0}^{t}\left(1-\beta_\mathbf{x}\right)^{t-k}\left[\nabla_\mathbf{x} F_j\left(\mathbf{x}_k^{(i)}, \mathbf{y}_k^{(i)} ; \mathbf{z}_k^{(i)}\right)\right.\\
&\left.-\nabla_\mathbf{x} F_j\left(\mathbf{x}_k^{(i)}, \mathbf{y}_k^{(i)} ; \mathbf{z}_{k+1}^{(i)}\right)\right]+\nabla_\mathbf{x} F_j\left(\mathbf{x}_t^{(i)}, \mathbf{y}_t^{(i)} ; \mathbf{z}_{t+1}^{(i)}\right)\bigg). 
\end{aligned}
\end{equation}

From the definition of $\bar{\mathbf{x}}_t$, we know $\bar{\mathbf{x}}_{t+1}=\bar{\mathbf{x}}_t-\eta_\mathbf{x} \bar{\mathbf{v}}_t$, so we have:
\begin{equation}
\begin{aligned}
\bar{\mathbf{x}}_{t+1}=\bar{\mathbf{x}}_t-\eta_\mathbf{x} \left(\bar{\mathbf{g}}_t + \mathcal{N}_{\mathbf{x}, t}\right),
\end{aligned}
\end{equation}
where 
$\mathcal{N}_{\mathbf{x}, t} \sim \mathcal{N}\left(0, \frac{\sigma_\mathbf{x}^2}{m} I_{d_1}\right)$.

\begin{lemma} 
\label{lemma 1}
\citep{wang2017differentially} In single parameter DP-GD paradigm whose model updates as $\bar{\mathbf{x}}_{t+1}=\bar{\mathbf{x}}_t-\eta_\mathbf{x} \left(\bar{\mathbf{g}}_t + \mathcal{N}_{\mathbf{x}, t}\right)$, meanwhile the loss function is $G$-lipschitz, for $\theta, \gamma>0$, for some constant $c$, it is $(\theta, \gamma)$-DP if the random noise is zero mean gaussian noise, i.e., $\mathcal{N}_{\mathbf{x}, t} \sim \mathcal{N}\left(0, \frac{\sigma_\mathbf{x}^2}{m} I_{d_1}\right)$, and  $\sigma_\mathbf{x}^2=c \frac{G^2 T \log (1 / \gamma)}{m \theta^2}$.
\end{lemma}

\begin{table*}[]
\centering
\caption{AUROC score of each algorithm over epochs during the robust logistic regression experiments on `a8a', `a9a' and CIFAR-10 datasets.}
\begin{small}
\subfloat[Impact of total number of agents $m$.]
{
\begin{tabular}{|c|c|c|c|c|c|c|c|c|c|c|c|c|}
\hline
\multirow{1}{*}{$m$} & \multicolumn{3}{c|}{$m = 5$} & \multicolumn{3}{c|}{$m = 10$} & \multicolumn{3}{c|}{$m = 15$} & \multicolumn{3}{c|}{$m = 20$}  \\ 
\hline
\multirow{1}{*}{Method}
 & a8a & a9a & CIFAR-10 & a8a & a9a & CIFAR-10 & a8a & a9a & CIFAR-10 & a8a & a9a & CIFAR-10 \\ \hline
SGDA & 0.7590 & 0.7164 & 0.6648 & 0.7801 & 0.7029 & 0.6705 & 0.7626 & 0.6968 & 0.6762 & 0.7625 & 0.6887 & 0.6778\\ \hline
DP-SGDA & 0.7383 & 0.7037 & 0.5910 & 0.7417 & 0.7047 & 0.6041 & 0.7453 & 0.7453 & 0.6161 & 0.7302 & 0.6945 & 0.6332\\ \hline
DM-HSGD & 0.7519 & 0.6708 & 0.6644 & 0.7420 & 0.7169 & 0.6695 & 0.7853 & 0.7053 & 0.6767 & 0.7702 & 0.6977 & 0.6772\\ \hline
\rowcolor{greyL}
\algns & 0.8094 & 0.7003 & 0.5927 & 0.7392 & 0.6692 & 0.6099 & 0.7751 & 0.6970 & 0.6259 & 0.7457 & 0.6926 & 0.6367 \\ \hline
\end{tabular}
}
% \vskip 0.1in
\\
\subfloat[Impact of sparsity level $p$.]
{
\begin{tabular}{|c|c|c|c|c|c|c|c|c|c|c|c|c|}
\hline
\multirow{1}{*}{$p$} & \multicolumn{3}{c|}{$t = 0.2$} & \multicolumn{3}{c|}{$p = 0.5$} & \multicolumn{3}{c|}{$p = 0.8$} & \multicolumn{3}{c|}{$p = 1$}  \\ 
\hline
\multirow{1}{*}{Method}
 & a8a & a9a & CIFAR-10 & a8a & a9a & CIFAR-10 & a8a & a9a & CIFAR-10 & a8a & a9a & CIFAR-10 \\ \hline
SGDA & 0.7388 & 0.6778 & 0.6624 & 0.7270 & 0.6357 & 0.6648 & 0.7373 & 0.6965 & 0.6612 & 0.7276 & 0.6971 & 0.6601 \\ \hline
DP-SGDA & 0.7500 & 0.6582 & 0.5914 & 0.7374 & 0.6591 & 0.5910 & 0.7181 & 0.7096 & 0.5937 & 0.7380 & 0.7178 & 0.5969\\ \hline
DM-HSGD & 0.7674 & 0.6588 & 0.6632 & 0.7018 & 0.7059 & 0.6644 & 0.7247 & 0.6753 & 0.6622 & 0.6888 & 0.6716 & 0.6619 \\ \hline
\rowcolor{greyL}
\algns & 0.7272 & 0.5971 & 0.5987 & 0.7825 & 0.7039 & 0.5927 & 0.7696 & 0.6504 & 0.5910 & 0.7666 & 0.6758 & 0.5906\\ \hline
\end{tabular}
}
\\
\subfloat[Impact of $\theta$.]
{
\centering
\begin{tabular}{|c|c|c|c|c|c|c|c|c|c|c|c|c|}
\hline
\multirow{1}{*}{$\theta$} & \multicolumn{3}{c|}{$\theta= 0.005$ 	
} & \multicolumn{3}{c|}{$\theta= 0.01$ 	
} & \multicolumn{3}{c|}{$\theta= 0.05$ 	
} & \multicolumn{3}{c|}{$\theta= 0.1$ 	
}  \\ 
\hline
\multirow{1}{*}{Method}
 & a8a & a9a & CIFAR-10 & a8a & a9a & CIFAR-10 & a8a & a9a & CIFAR-10 & a8a & a9a & CIFAR-10 \\ \hline
SGDA & 0.7719 & 0.6957 & 0.6648 & 0.7719 & 0.6957 & 0.6648 & 0.7719 & 0.6957 & 0.6648 & 0.7719 & 0.6957 & 0.6648 
 \\ \hline
DP-SGDA & 0.7595 & 0.6691 & 0.5918 & 0.7555 & 0.6673 & 0.5910 & 0.7257 & 0.6778 & 0.5965 & 0.7321 & 0.6750 & 0.6127 \\ \hline
DM-HSGD & 0.7941 & 0.7142 & 0.6644 & 0.7941 & 0.7142 & 0.6644 & 0.7941 & 0.7142 & 0.6644 & 0.7941 & 0.7142 & 0.6644
 \\ \hline
\rowcolor{greyL}
\algns & 0.6653 & 0.5644 & 0.5932 & 0.6991 & 0.6026 & 0.5927 & 0.7651 & 0.7011 & 0.6000 & 0.7658 & 0.6170 & 0.5978 \\ \hline
\end{tabular}
}
\\
\subfloat[Impact of $\gamma$.]
{
\centering
\begin{tabular}{|c|c|c|c|c|c|c|c|c|c|c|c|c|}
\hline
\multirow{1}{*}{$\gamma$} & \multicolumn{3}{c|}{$\gamma= 1/60000$ 	
	
} & \multicolumn{3}{c|}{$\gamma= 1/30000	$ 
	
} & \multicolumn{3}{c|}{$\gamma = 1/5000	$
	
} & \multicolumn{3}{c|}{$\gamma= 1/1000	$ 
	
}  \\ 
\hline
\multirow{1}{*}{Method}
 & a8a & a9a & CIFAR-10 & a8a & a9a & CIFAR-10 & a8a & a9a & CIFAR-10 & a8a & a9a & CIFAR-10 \\ \hline
SGDA & 0.7719 & 0.6957 & 0.6644 & 0.7719 & 0.6957 & 0.6644 & 0.7719 & 0.6957 & 0.6644 & 0.7719 & 0.6957 & 0.6644 \\ \hline
DP-SGDA & 0.7325 & 0.6507 & 0.5922 & 0.7564 & 0.7112 & 0.5910 & 0.7383 & 0.6990 & 0.6168 & 0.7757 & 0.7102 & 0.6007 \\ \hline
DM-HSGD & 0.7941 & 0.7142 & 0.6644 & 0.7941 & 0.7142 & 0.6644 & 0.7941 & 0.7142 & 0.6644 & 0.7941 & 0.7142 & 0.6644 \\ \hline
\rowcolor{greyL}
\algns & 0.7741 & 0.6927 & 0.5962 & 0.7979 & 0.6692 & 0.5927 & 0.7444 & 0.7023 & 0.5948 & 0.7719 & 0.6859 & 0.5927\\ \hline
\end{tabular}
}
\label{table1}
%\end{normalsize}
\end{small}
\end{table*}

However, we use a momentum gradient descent method, so the parameter updates involve $\bar{\mathbf{g}}_t$ instead of just $\nabla_\mathbf{x} F_i\left(\cdot, \cdot ; \cdot\right)$, according to Assumption \ref{assumption 6}, since each gradient term $\nabla_\mathbf{x} F_i\left(\cdot, \cdot ; \cdot\right)$  is $L$-lipschitz, the weighted sum operation does not change the lipschitz constant. Therefore, $\mathbf{g}_t$ is $G$-lipschitz where $G$ is derived from $L$, we clarify that the Lipschitz constant \( G \) follows from the \( L \)-Lipschitz continuity of \( \nabla F \) and the structure of Eq.~(\ref{eq_contin}), where \( g_t \) is a linear combination of Lipschitz-smooth gradients. Using the recursion \( G \leq L + (1 - \beta_{\mathbf{x}}) G \), we get \( G \leq \frac{L}{\beta_{\mathbf{x}}} \).

The authors in~\citep{wang2017differentially} provide a tight noise bound for differentially private gradient descent under a single-parameter condition. However, in the min-max paradigm,  privacy leakage also arises from the gradient information, regardless of whether it is used for minimization or maximization. Since the updates for $\mathbf{y}$ share the same structure as those for $\mathbf{x}$, the noise variance derived in \citep{wang2017differentially} can be symmetrically applied to $\mathbf{y}$. Notably, the privacy cost is independent of whether the process involves minimization or maximization. Therefore, by injecting the noise proposed in \citep{wang2017differentially} into both $\mathbf{x}$ and $\mathbf{y}$, the DP guarantee can still be ensured. Since the proof process is nearly identical (with the only difference being its application to $\mathbf{y}$ as well), we directly adopt the result in our theorem. Therefore, by Lemma \ref{lemma 1} we give the privacy guarantees of \alg.

\begin{theorem} 
\label{theorem 2}
If $F_i(\cdot, \cdot ; \cdot)$ satisfies Assumption \ref{assumption 6} then for some privacy budget 
 $\theta = \Omega\left(\frac{L_g d^{1/2} \log(1/\gamma)^{1/2}}{m^{1/2} \epsilon^4} \right)$,$\gamma>0$, we get a utility for \alg to be $(\theta, \gamma)$-DP if
\begin{equation}
\begin{aligned}
\sigma_{x}, \sigma_{y}=\mathcal{O}\left(\frac{L_g \sqrt{\left(\frac{8 T(T+1)(2 T+1)}{3}+4 T\right) \log (1 / \gamma)}}{2 \theta \sqrt{m}}\right).
\end{aligned}
\end{equation}
\end{theorem}

\begin{remark}
While $\sigma_x$ and $\sigma_y$ scale with $T$ as given in Theorem~\ref{theorem 2}, the additional noise-induced error remains controlled under an appropriate choice of $\theta$. Specifically, by ensuring
\begin{align}
\theta = \Omega\left(\frac{L_g d^{1/2} \log(1/\gamma)^{1/2}}{m^{1/2} \epsilon^4} \right)    
\end{align}

we obtain:
\begin{align}
\frac{L_g^2 d \log(1/\gamma)}{\theta^2 m \epsilon^6} = O(\epsilon^2)
\end{align}

Then we can get an optimization error bound of
\begin{align}
\frac{1}{T} \sum_{t=0}^{T-1}\mathbb{E} \|\nabla \Phi(\bar{\mathbf{x}}_t)\|^2 = O(\epsilon^2)   
\end{align}

which ensures that our algorithm maintains the desired convergence rate without being dominated by noise.
\end{remark}

% !TEX root = main.tex

\section{Experiments}\label{sec:experiment}

\subsection{Robust logistic regression in decentralized min-max problem }
In this section, we conduct the experiment of decentralized robust logistic regression 
based on ``a8a''~\cite{chang2011libsvm} ,``a9a''~\cite{chang2011libsvm}, and CIFAR-10~\cite{krizhevsky2009learning} datasets.
In these experiments, we compare the \alg, DM-HSGD~\cite{xian2021faster}, SGDA~\cite{beznosikov2023stochastic}, and DP-SGDA~\cite{yang2022differentially} algorithms.
We partition the given dataset as $\left\{\left(a_i, b_i\right)\right\}_{i=1}^m$, where each feature vector $a_i \in \mathbb{R}^d$ and each label $b_i \in \{-1, 1\}$. 
Robust logistic regression is formulated as the following min-max problem:
\begin{align}
    \min _{\mathbf{x} \in \mathbb{R}^d} \max _{\mathbf{y} \in \Delta_m} f(\mathbf{x}, \mathbf{y})=\sum_{i=1}^m \mathbf{y}_i l_i(\mathbf{x})-V(\mathbf{y})+g(\mathbf{x}),
\end{align}
where $m$ is the total number of agents, $\mathbf{y}_i$ represents the $i$-th component of the variable $\mathbf{y}$. The logistic loss function is defined by
$
        l_i(\mathbf{x}) = \log \left(1 + \exp\left(-b_i a_i^\top \mathbf{x}\right)\right).
$
The divergence measure $V(\mathbf{y})$ is given by
$
    V(\mathbf{y}) = \frac{1}{2} \lambda_1 \|m \mathbf{y} - \mathbf{1}\|^2.
$
The simplex $\Delta_m$ in $\mathbb{R}^m$ is defined as
$
    \Delta_m = \left\{ \mathbf{y} \in \mathbb{R}^m \,\middle|\, 0 \leq \mathbf{y}_i \leq 1 \text{ for all } i, \ \sum_{i=1}^m \mathbf{y}_i = 1 \right\}.
$
Additionally, the nonconvex regularization term $g(\mathbf{x})$ is formulated as
$
    g(\mathbf{x}) = \lambda_2 \sum_{i=1}^d \frac{\alpha \mathbf{x}_i^2}{1 + \alpha \mathbf{x}_i^2}.
$
Following the experimental configurations outlined here, we set the parameters to $\lambda_1 = \frac{1}{m^2}$, $\lambda_2 = 0.001$, and $\alpha = 10$ in our experiments.

For the evaluation of the \alg, DM-HSGD, SGDA, and DP-SGDA algorithms, we show the results of our experiment in Table \ref{table1}. Regarding the optimization parameters within the neural network, the learning rates for the model parameters $\mathbf{x}$ and their dual variables $\mathbf{y}$ are selected from the set $\{1.0, 0.1, 0.01, 0.001\}$. The mini-batch size is fixed at $20$. Specifically for the \alg and DM-HSGD algorithms, the batch size for the initial iteration is set to $b_0 = 10,\!000$. Additionally, the gradient weight adjustment parameters $\beta_x$ and $\beta_y$ are chosen from the set $\{0.5, 0.1, 0.01\}$.  

\begin{remark}
It is worth noting that our experiments did not apply gradient clipping, although gradient clipping is common in differential privacy (DP) training but not strictly required. In our case, the DP noise level is moderate, and a well-tuned learning rate ensures stable convergence without clipping. Experimental results show no signs of instability. Additional experiment on gradient clipping is provided in Appendix \ref{sec:appendixC}, showing similar trends. 
\end{remark}

In the experiment, the communication topology among agents is modeled using an Erdős–Rényi random graph  $\mathcal{G}(m, p)$ , where $m$ is the number of agents and $p \in [0,1]$ denotes the sparsity level, i.e., the probability that an edge exists between any two agents. A higher $p$ implies a denser communication network. Formally, each edge is included in the graph independently with probability:
\begin{align}
\mathbb{P}[(i,j) \in \mathcal{E}] = p, \quad \forall i \ne j,
\end{align}
where $\mathcal{E}$ denotes the edge set of the communication graph. The expected degree of each node is $(m - 1)p$, and the total expected number of edges is $\frac{p m(m - 1)}{2}$. Therefore, we have the definition of sparsity level $p$:
\begin{align}
p = \frac{2|\mathcal{E}|}{m(m - 1)},
\end{align}
where $|\mathcal{E}|$ is the total number of edges or links in the system.

We conduct control group experiments on robust logistic regression, examined the impact of several factors. These include the number of agents in the network, the sparsity level $p$ of the connectivity matrix, and the adding noise is affected by $\theta$ and $\gamma$. 
Table~\ref{table1} illustrates the AUROC score of each algorithm over epochs during the robust logistic regression experiments. 

From the Table~\ref{table1}, we observe that our algorithm performs better with a small number of agents. This can be attributed to the injection of noise into the local gradients before the model updates. However, as the communication frequency increases, the performance inevitably declines. Furthermore, in the experiments regarding sparsity levels, our algorithm exhibits superior performance compared to other algorithms under high sparsity conditions. The experimental data on the two parameters affecting noise show that our algorithm demonstrates greater stability when adding noise at different levels. Therefore, this indicates that its theoretical design is effective in practical applications, outperforming existing methods such as DP-SGDA, and in certain cases, approaching or even surpassing non-private mechanisms like SGDA and DM-HSGD.
To further assess the applicability of our algorithm in non-convex deep learning settings, we also conduct experiments on image classification tasks using a multilayer perceptron (MLP) with the Fashion-MNIST dataset. These results, which demonstrate consistent advantages of our method over baselines, are provided in Appendix~\ref{sec:appendixC}.

\subsection{Robustness to DLG Attacks}

\begin{figure}[h!]
\vspace{-.15in}
    \centering
    \subfloat[MNIST dataset]{%
        \includegraphics[width=0.48\columnwidth]{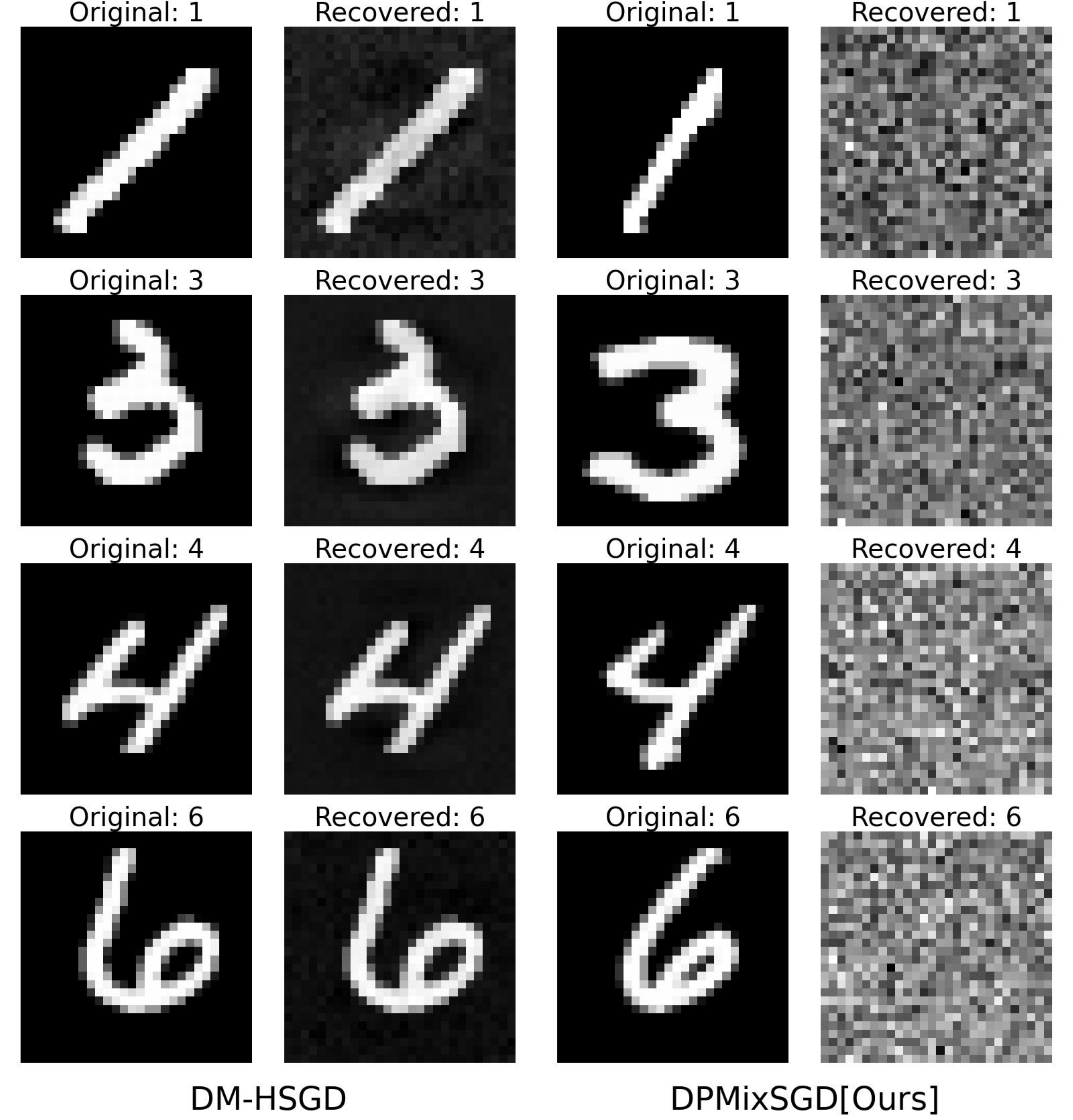}%
    }\hfill
    \subfloat[Fashion-MNIST dataset]{%
        \includegraphics[width=0.48\columnwidth]{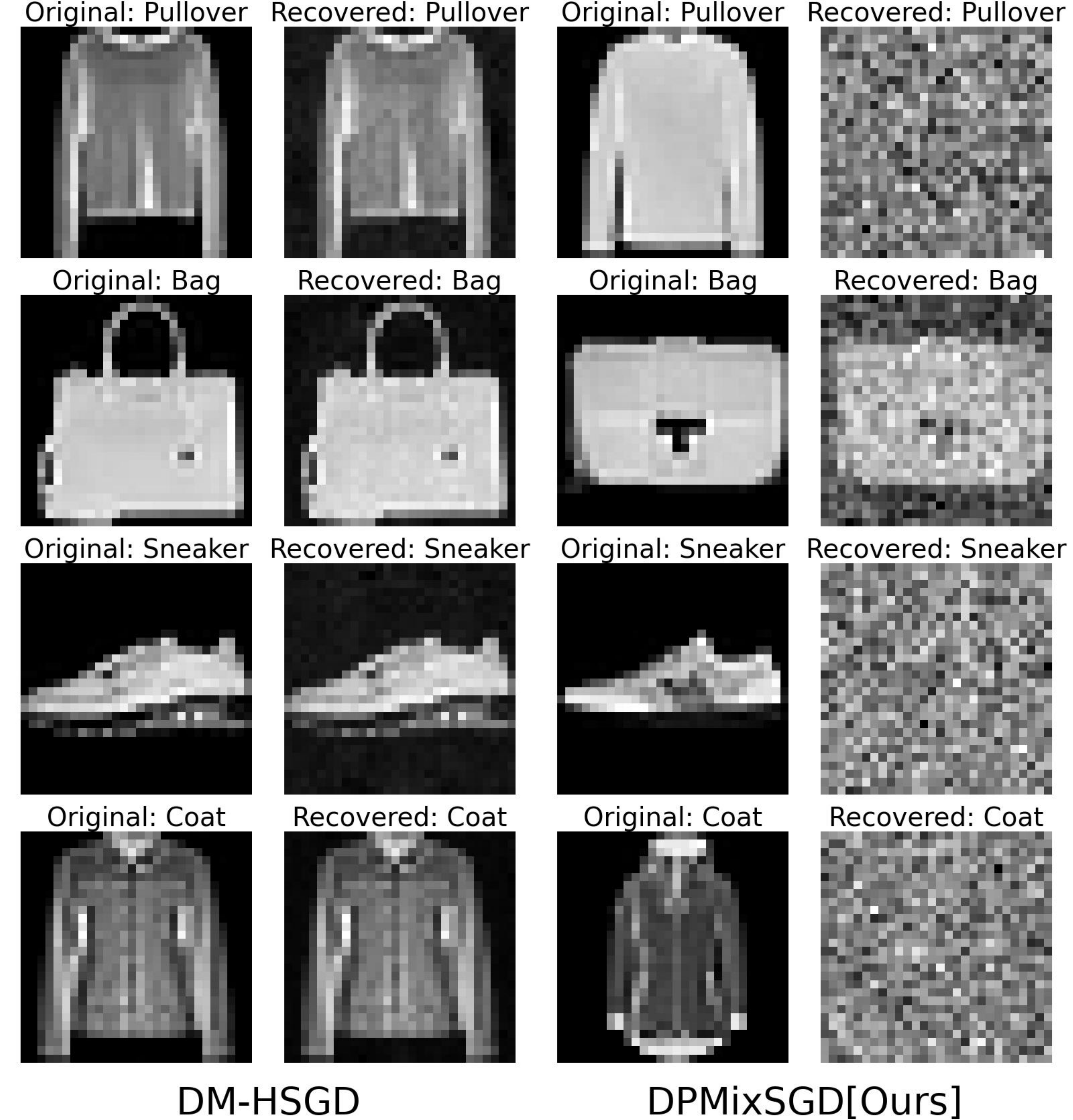}%
    }
    \caption{DLG Attack Reconstruction Results.}
    \label{Fig1}
    \vspace{-.15in}
\end{figure}

To further evaluate the privacy protection capabilities of our proposed \alg algorithm, we conduct additional experiments on the MNIST and Fashion-MNIST datasets using a multi-layer perceptron (MLP) model. These experiments focus on the algorithm’s robustness against Deep Leakage from Gradients (DLG) attacks ~\cite{zhu2019deep}, a gradient inversion method that can potentially recover private training data from shared gradients.
We assess the reconstruction quality of the DLG attack under a given noise level $\sigma = 1$. The results shown in Figure \ref{Fig1} demonstrate that our differentially private algorithm effectively mitigates the risk of visual identity recovery, significantly enhancing privacy robustness.
For transparency and reproducibility, the experimental settings are specified as follows: the learning rate is set to 0.01 for the primal variable $\mathbf{x}$ and 0.001 for the dual variable $\mathbf{y}$, and the mini-batch size is fixed at 128.  The evaluation confirms that \alg not only maintains strong privacy guarantees but also resists gradient leakage attacks in practical decentralized learning scenarios.

% !TEX root = main.tex

\section{Conclusion}
\label{sec:conlusion}

In this paper, we addressed the challenges of privacy protection in decentralized min-max learning problem by proposing a novel \alg algorithm. Our theoretical analysis proves that \alg ensures rigorous privacy guarantees while maintaining provable convergence.Empirical results demonstrate that out proposed method DPMixSGD not only upholds strong privacy guarantees but also effectively resists gradient leakage in practical decentralized learning scenarios. This work contributes to advancing decentralized learning by effectively balancing the need for privacy and efficient communication in distributed systems, providing a robust framework for future applications in privacy-sensitive domains.

\begin{acks}
    We thank the anonymous reviewers for their helpful comments.
\end{acks}

\bibliographystyle{ACM-Reference-Format}
\bibliography{Reference}

\newpage
\appendix
\onecolumn
% !TEX root = main.tex

\section{Proof of convergence}\label{sec:apendixa}
From the algorithm, we obtain that:
\begin{equation}
\begin{aligned}
\bar{\mathbf{g}}_t^*&=\bar{\mathbf{g}}_t+\frac{1}{m} \sum_{i=1}^m n_{\mathbf{x}, t}^{(i)}\\
&=\frac{1}{m} \sum_{i=1}^m \nabla_\mathbf{x} F_i\left(\mathbf{x}_t^{(i)}, \mathbf{y}_t^{(i)}, \mathbf{z}_t^{(i)}\right)+\left(1-\beta_\mathbf{x}\right)\left(\bar{\mathbf{g}}_{t-1}-\frac{1}{m} \sum_{i=1}^m \nabla_\mathbf{x} F_i\left(\mathbf{x}_{t-1}^{(i)}, \mathbf{y}_{t-1}^{(i)}, \mathbf{z}_t^{(i)}\right)\right)+\mathcal{N}_{\mathbf{x}, t}
\end{aligned}
\end{equation}
Where $\mathcal{N}_{\mathbf{x}, t}=\frac{\sum_{i=1}^m n_{\mathbf{x}, t}^{(i)}}{m}$

\subsection{Basic Lemmas and Important Conclusions}
First, we introduce following basic lemmas, which are broadly used in the convergence analysis of optimization algorithms.

\textbf{Lemma \refstepcounter{lemma}\label{lemma 2}2}. Let vector $X$ be a stochastic variable. Then we have
\begin{equation}
\begin{aligned}
    0 \leq \mathbb{E}\|X-\mathbb{E} X\|^2=\mathbb{E}\|X\|^2-\|\mathbb{E} X\|^2 \leq \mathbb{E}\|X\|^2
\end{aligned}
\end{equation}
\textbf{Lemma \refstepcounter{lemma}\label{lemma 3}3}. Let $X_1, X_2, \cdots, X_n$ be $m$ independent stochastic variables of which the means are 0. Then we have
\begin{equation}
\begin{aligned}
\mathbb{E}\left\|\sum_{i=1}^m X_i\right\|^2=\sum_{i=1}^m \mathbb{E}\left\|X_i\right\|^2
\end{aligned}
\end{equation}
\textbf{Lemma\refstepcounter{lemma}\label{lemma 4} 4}. Suppose $A$ and $B$ are two matrices. Then it satisfies
\begin{equation}
\begin{aligned}
\|A B\|_F \leq\|A\|_2\|B\|_F
\end{aligned}
\end{equation}
\textbf{Lemma\refstepcounter{lemma}\label{lemma 5} 5.} (Lemma 4.3 from \cite{lin2020gradient}) $\Phi(\mathbf{x})$ is $(L+\kappa L)$-smooth and $\mathbf{y}^*(\cdot)$ is $\kappa$-Lipschitz, which means $\left\|\mathbf{y}^*\left(\mathbf{x}_1\right)-\mathbf{y}^*\left(\mathbf{x}_2\right)\right\| \leq \kappa\left\|\mathbf{x}_1-\mathbf{x}_2\right\|$ for any $\mathbf{x}_1, \mathbf{x}_2 \in \mathbb{R}^{d_1}$.

\textbf{Lemma \refstepcounter{lemma}\label{lemma 6}6}. When $\eta_\mathbf{y} \leq \frac{1}{5 L}$ we have following estimation for $\delta_t$.
\begin{equation}
\begin{aligned}
&\sum_{t=0}^{T-1} \delta_t 
\leq  \frac{4 \kappa}{L \eta_\mathbf{y}} \delta_0+\frac{10 \eta_\mathbf{y}}{\mu} \sum_{t=1}^{T-1}\left(1-\frac{\mu \eta_\mathbf{y}}{4}\right)^{T-t-1} \sum_{s=0}^{t-1}\left\|\bar{\mathbf{u}}_s-\frac{1}{m} \sum_{i=1}^m \nabla f_i\left(\mathbf{x}_s^{(i)}, \mathbf{y}_s^{(i)}\right)\right\|^2+\frac{40 \kappa^2}{m} \sum_{t=0}^{T-1}\\
&\left(\left\|X_t-\bar{X}_t\right\|_F^2+\left\|Y_t-\bar{Y}_t\right\|_F^2\right)+\frac{20 \kappa^4 \eta_\mathbf{x}^2}{L^2 \eta_\mathbf{y}^2} \sum_{t=0}^{T-1}\left\|\bar{\mathbf{v}}_t\right\|^2-\frac{14 \eta_\mathbf{y}}{5 \mu} \sum_{t=0}^{T-1}\left(1-\left(1-\frac{\mu \eta_\mathbf{y}}{4}\right)^{T-t}\right)\left\|\bar{\mathbf{u}}_t\right\|^2
\end{aligned}
\end{equation}
\textit{Proof:} As we defined before, $\hat{\mathbf{y}}_t=\arg \max _{\mathbf{y} \in \mathcal{Y}} f\left(\bar{\mathbf{x}}_t, \mathbf{y}\right)$, we obatin:
\begin{equation}
\label{eq6.3}
\left\|\bar{\mathbf{y}}_{t+1}-\hat{\mathbf{y}}_t\right\|^2=\left\|\bar{\mathbf{y}}_t+\eta_\mathbf{y} \bar{\mathbf{u}}_t-\hat{\mathbf{y}}_t\right\|^2=\left\|\bar{\mathbf{y}}_t-\hat{\mathbf{y}}_t\right\|^2+\eta_\mathbf{y}^2\left\|\bar{\mathbf{u}}_t\right\|^2+2 \eta_\mathbf{y}\left\langle\bar{\mathbf{y}}_t-\hat{\mathbf{y}}_t, \bar{\mathbf{u}}_t\right\rangle
\end{equation}
As function $f$ is strongly-concave in $\mathbf{y}$, we get:
\begin{equation}
\begin{aligned}
\label{eq6.1}
f\left(\bar{\mathbf{x}}_t, \hat{\mathbf{y}}_t\right) &\leq  f\left(\bar{\mathbf{x}}_t, \bar{\mathbf{y}}_t\right)+\left\langle\nabla_\mathbf{y} f\left(\bar{\mathbf{x}}_t, \bar{\mathbf{y}}_t\right), \hat{\mathbf{y}}_t-\bar{\mathbf{y}}_t\right\rangle-\frac{\mu}{2}\left\|\hat{\mathbf{y}}_t-\bar{\mathbf{y}}_t\right\|^2 \\
&=f\left(\bar{\mathbf{x}}_t, \bar{\mathbf{y}}_t\right)-\frac{\mu}{2}\left\|\hat{\mathbf{y}}_t-\bar{\mathbf{y}}_t\right\|^2 +\left\langle\bar{\mathbf{u}}_t, \hat{\mathbf{y}}_t-\bar{\mathbf{y}}_{t+1}\right\rangle+\left\langle\nabla_{\mathbf{y}} f\left(\bar{\mathbf{x}}_t, \bar{\mathbf{y}}_t\right)-\bar{\mathbf{u}}_t, \hat{\mathbf{y}}_t-\bar{\mathbf{y}}_{t+1}\right\rangle\\
&\quad +\left\langle\nabla_{\mathbf{y}} f\left(\bar{\mathbf{x}}_t, \bar{\mathbf{y}}_t\right), \bar{\mathbf{y}}_{t+1}-\bar{\mathbf{y}}_t\right\rangle
\end{aligned}
\end{equation}
From assumption \ref{assumption 1}, and let $L \eta_\mathbf{y} \leq \frac{1}{5}$ we know:
\begin{equation}
\begin{aligned}
\label{eq6.2}
 -\frac{1}{10\eta_\mathbf{y}}\left\|\bar{\mathbf{y}}_{t+1}-\bar{\mathbf{y}}_t\right\|^2 &\leq-\frac{L}{2}\left\|\bar{\mathbf{y}}_{t+1}-\bar{\mathbf{y}}_t\right\|^2  \\
&\leq f\left(\bar{\mathbf{x}}_t, \mathbf{y}_{t+1}\right)-f\left(\bar{\mathbf{x}}_t, \bar{\mathbf{y}}_t\right)-\left\langle\nabla_{\mathbf{y}} f\left(\bar{\mathbf{x}}_t, \bar{\mathbf{y}}_t\right), \bar{\mathbf{y}}_{t+1}-\bar{\mathbf{y}}_t\right\rangle
\end{aligned}
\end{equation}
Adding Eq.(\ref{eq6.1}) and (\ref{eq6.2}) together, and from the algorithm, we know that $\bar{\mathbf{y}}_{t+1}-\bar{\mathbf{y}}_t = \mathbf{u}_t$.
\begin{equation}
\begin{aligned}
&f\left(\bar{\mathbf{x}}_t, \hat{\mathbf{y}}_t\right)-f\left(\bar{\mathbf{x}}_t, \mathbf{y}_{t+1}\right) +\frac{\mu}{2}\left\|\hat{\mathbf{y}}_t-\bar{\mathbf{y}}_t\right\|^2 \\
&\leq  \left\langle\bar{\mathbf{u}}_t, \hat{\mathbf{y}}_t-\bar{\mathbf{y}}_{t+1}\right\rangle+\left\langle\nabla_{\mathbf{y}} f\left(\bar{\mathbf{x}}_t, \bar{\mathbf{y}}_t\right)-\bar{\mathbf{u}}_t, \hat{\mathbf{y}}_t-\bar{\mathbf{y}}_{t+1}\right\rangle+\frac{1}{10\eta_\mathbf{y}}\left\|\bar{\mathbf{y}}_{t+1}-\bar{\mathbf{y}}_t\right\|^2\\
&= \left\langle\bar{\mathbf{u}}_t, \bar{\mathbf{y}}_t-\bar{\mathbf{y}}_t\right\rangle+\left\langle\bar{\mathbf{u}}_t, \hat{\mathbf{y}}_t-\bar{\mathbf{y}}_{t+1}\right\rangle+\left\langle\nabla_{\mathbf{y}} f\left(\bar{\mathbf{x}}_t, \bar{\mathbf{y}}_t\right)-\bar{\mathbf{u}}_t, \hat{\mathbf{y}}_t-\bar{\mathbf{y}}_{t+1}\right\rangle+\frac{\eta_\mathbf{y}}{10}\left\|\bar{\mathbf{u}}_t\right\|^2\\
&= \left\langle\bar{\mathbf{u}}_t, \hat{\mathbf{y}}_t-\bar{\mathbf{y}}_t\right\rangle-\eta_\mathbf{y}\left\|\bar{\mathbf{u}}_t\right\|^2+\left\langle\nabla_{\mathbf{y}} f\left(\bar{\mathbf{x}}_t, \bar{\mathbf{y}}_t\right)-\bar{\mathbf{u}}_t, \hat{\mathbf{y}}_t-\bar{\mathbf{y}}_{t+1}\right\rangle+\frac{\eta_\mathbf{y}}{10}\left\|\bar{\mathbf{u}}_t\right\|^2\\
&=\left\langle\bar{\mathbf{u}}_t, \hat{\mathbf{y}}_t-\bar{\mathbf{y}}_t\right\rangle+\left\langle\nabla_{\mathbf{y}} f\left(\bar{\mathbf{x}}_t, \bar{\mathbf{y}}_t\right)-\bar{\mathbf{u}}_t, \hat{\mathbf{y}}_t-\bar{\mathbf{y}}_{t+1}\right\rangle-\frac{9\eta_\mathbf{y}}{10}\left\|\bar{\mathbf{u}}_t\right\|^2\\
&\leq \left\langle\bar{\mathbf{u}}_t, \hat{\mathbf{y}}_t-\bar{\mathbf{y}}_t\right\rangle + \frac{2}{\mu}\left\|\nabla_{\mathbf{y}} f\left(\bar{\mathbf{x}}_t, \bar{\mathbf{y}}_t\right)-\bar{\mathbf{u}}_t\right\|^2 + \frac{\mu}{8}\left\|\hat{\mathbf{y}}_t-\bar{\mathbf{y}}_{t+1}\right\|^2-\frac{9\eta_\mathbf{y}}{10}\left\|\bar{\mathbf{u}}_t\right\|^2\\
&\leq \left\langle\bar{\mathbf{u}}_t, \hat{\mathbf{y}}_t-\bar{\mathbf{y}}_t\right\rangle + \frac{2}{\mu}\left\|\nabla_{\mathbf{y}} f\left(\bar{\mathbf{x}}_t, \bar{\mathbf{y}}_t\right)-\bar{\mathbf{u}}_t\right\|^2 + \frac{\mu}{4}\left\|\hat{\mathbf{y}}_t-\bar{\mathbf{y}}_t\right\|^2+\frac{\mu}{4}\left\|\bar{\mathbf{y}}_t-\bar{\mathbf{y}}_{t+1}\right\|^2-\frac{9\eta_\mathbf{y}}{10}\left\|\bar{\mathbf{u}}_t\right\|^2\\
&= \left\langle\bar{\mathbf{u}}_t, \hat{\mathbf{y}}_t-\bar{\mathbf{y}}_t\right\rangle + \frac{2}{\mu}\left\|\nabla_{\mathbf{y}} f\left(\bar{\mathbf{x}}_t, \bar{\mathbf{y}}_t\right)-\bar{\mathbf{u}}_t\right\|^2 + \frac{\mu}{4}\left\|\hat{\mathbf{y}}_t-\bar{\mathbf{y}}_t\right\|^2-\left(\frac{9\eta_\mathbf{y}}{10}-\frac{\mu\eta_\mathbf{y}^2}{4}\right)\left\|\bar{\mathbf{u}}_t\right\|^2
\end{aligned}
\end{equation}
Where in the second inequality, we use Young's inequality, and in the last inequality we use Cauchy-Schwartz inequality. As we defined $\hat{\mathbf{y}_t}$, so $f\left(\bar{\mathbf{x}}_t,\hat{\mathbf{y}_t}\right) \geq f\left(\bar{\mathbf{x}}_t,\bar{\mathbf{y}}_{t+1}\right)$.
\begin{equation}
\label{eq6.4}
\frac{\mu \eta_\mathbf{y}}{2}\left\|\hat{\mathbf{y}}_t-\bar{\mathbf{y}}_t\right\|^2 \leq 2 \eta_\mathbf{y} \left\langle\bar{\mathbf{u}}_t, \hat{\mathbf{y}}_t-\bar{\mathbf{y}}_t\right\rangle+\frac{4\eta_\mathbf{y}}{\mu}\left\|\nabla_{\mathbf{y}} f\left(\bar{\mathbf{x}}_t, \bar{\mathbf{y}}_t\right)-\bar{\mathbf{u}}_t\right\|^2-\left(\frac{9 \eta_{\mathbf{y}^2}}{5}-\frac{\mu \eta_{\mathbf{y}}^3}{2}\right)\left\|\bar{\mathbf{u}}_t\right\|^2
\end{equation}
Combining Eq.(\ref{eq6.3}) and (\ref{eq6.4}), and we set $\mu\eta_\mathbf{y}\leq L\eta_\mathbf{y} \leq \frac{1}{5}$ we can get:
\begin{equation}
\begin{aligned}
\label{eq6.5}
\left\|\bar{\mathbf{y}}_{t+1}-\hat{\mathbf{y}}_t\right\|^2 \leq \left(1-\frac{\mu \eta_\mathbf{y}}{2}\right)\left\|\hat{\mathbf{y}}_t-\bar{\mathbf{y}}_t\right\|^2+\frac{4\eta_\mathbf{y}}{\mu}\left\|\nabla_{\mathbf{y}} f\left(\bar{\mathbf{x}}_t, \bar{\mathbf{y}}_t\right)-\bar{\mathbf{u}}_t\right\|^2 - \frac{7\eta_\mathbf{y}^2}{10}\left\|\bar{\mathbf{u}_t}\right\|^2
\end{aligned}
\end{equation}
By Young's inequality we have:
\begin{equation}
\begin{aligned}
\label{eq6.6}
&\left\|\bar{\mathbf{y}}_{t+1}-\hat{\mathbf{y}}_{t+1}\right\|^2 \\ &\leq\left(1+\frac{\mu \eta_\mathbf{y}}{4}\right)\left\|\bar{\mathbf{y}}_{t+1}-\hat{\mathbf{y}}_t\right\|^2+\left(1+\frac{4}{\mu \eta_\mathbf{y}}\right)\left\|\hat{\mathbf{y}}_{t+1}-\hat{\mathbf{y}}_t\right\|^2 \\
& \leq\left(1-\frac{\mu \eta_\mathbf{y}}{4}-\frac{\mu^2\eta_\mathbf{y}^2}{8}\right)\left\|\bar{\mathbf{y}}_t-\hat{\mathbf{y}}_t\right\|^2+\left(\frac{4 \eta_\mathbf{y}}{\mu}+\eta_\mathbf{y}^2\right)\left\|\nabla_\mathbf{y} f\left(\bar{\mathbf{x}}_t, \bar{\mathbf{y}}_t\right)-\bar{\mathbf{u}}_t\right\|^2\\
&\quad+\frac{\mu\eta_\mathbf{y}+4}{\mu\eta_\mathbf{y}}\left\|\hat{\mathbf{y}}_{t+1}-\hat{\mathbf{y}}_t\right\|^2-\left(1+\frac{\mu\eta_\mathbf{y}}{4}\right)\frac{7 \eta_\mathbf{y}^2}{10 }\left\|\bar{\mathbf{u}}_t\right\|^2 \\
& \leq\left(1-\frac{\mu \eta_\mathbf{y}}{4}\right)\left\|\bar{\mathbf{y}}_t-\hat{\mathbf{y}}_t\right\|^2+\frac{5 \eta_\mathbf{y}}{\mu}\left\|\nabla_\mathbf{y} f\left(\bar{\mathbf{x}}_t, \bar{\mathbf{y}}_t\right)-\bar{\mathbf{u}}_t\right\|^2+\frac{5 \kappa^3 \eta_\mathbf{x}^2}{L \eta_\mathbf{y}}\left\|\bar{\mathbf{v}}_t\right\|^2-\frac{7 \eta_\mathbf{y}}{10 }^2\left\|\bar{\mathbf{u}}_t\right\|^2
\end{aligned}
\end{equation}
Using Eq. (\ref{eq6.5}) and the fact that $L \eta_\mathbf{y} \leq \frac{1}{5}$ in the calculation of the second inequality, we obtain $-\frac{\mu^2\eta_\mathbf{y}^2}{8} \geq 0$. Additionally, $\frac{4 \eta_\mathbf{y}}{\mu} + \eta_\mathbf{y}^2 \leq \frac{4 \eta_\mathbf{y}}{\mu} + \frac{1 \eta_\mathbf{y}}{5 \mu} \leq \frac{5 \eta_\mathbf{y}}{\mu}$, and $-\left(1 + \frac{\mu \eta_\mathbf{y}}{4}\right) \leq -1$. We simplify the inequality using an approximation method, and the last inequality holds because the function $\mathbf{y}^*(\cdot)$ is $\kappa$-Lipschitz, therefore, we have $\left\|\hat{\mathbf{y}}_{t+1}-\hat{\mathbf{y}}_t\right\|^2 \leq \kappa^2 \eta_\mathbf{x}^2 \left\|\bar{\mathbf{v}}_t^2\right\|$ In combination with the previously provided conditions, we have $\frac{\mu\eta_\mathbf{y}+4}{\mu\eta_\mathbf{y}} \leq \frac{5}{\mu\eta_\mathbf{y}} = \frac{5\kappa}{L \eta_\mathbf{y}}$. By the Cauchy-Schwarz inequality and Assumption 1, we also have:
\begin{equation}
\label{eq6.7}
\left\|\nabla_\mathbf{y} f\left(\bar{\mathbf{x}}_t, \bar{\mathbf{y}}_t\right)-\bar{\mathbf{u}}_t\right\|^2 \leq 2\left\|\bar{\mathbf{u}}_t-\frac{1}{m} \sum_{i=1}^m \nabla f_i\left(\mathbf{x}_t^{(i)}, \mathbf{y}_t^{(i)}\right)\right\|^2+\frac{2 L^2}{m}\left(\left\|X_t-\bar{X}_t\right\|_F^2+\left\|Y_t-\bar{Y}_t\right\|_F^2\right)
\end{equation}
By definition of $\delta_t$ and the recursion in Eq.(\ref{eq6.6}) we obtain:
\begin{equation}
\begin{aligned}
\delta_t \leq & \left(1-\frac{\mu \eta_\mathbf{y}}{4}\right)^t \delta_0+\frac{5 \eta_\mathbf{y}}{\mu} \sum_{s=0}^{t-1}\left(1-\frac{\mu \eta_\mathbf{y}}{4}\right)^{t-s-1}\left\|\bar{\mathbf{u}}_s-\nabla_\mathbf{y} f\left(\bar{\mathbf{x}}_t, \bar{\mathbf{y}}_t\right)\right\|^2\\
& +\frac{5 \kappa^3 \eta_\mathbf{x}^2}{L \eta_\mathbf{y}} \sum_{s=0}^{t-1}\left(1-\frac{\mu \eta_\mathbf{y}}{4}\right)^{t-s-1}\left\|\bar{\mathbf{v}}_s\right\|^2 -\frac{7 \eta_\mathbf{y}^2}{10 } \sum_{s=0}^{t-1}\left(1-\frac{\mu \eta_\mathbf{y}}{4}\right)^{t-s-1}\left\|\bar{\mathbf{u}}_s\right\|^2
\end{aligned}
\end{equation}
Using Eq.(\ref{eq6.7}) to sum above equation we have:
\begin{equation}
\begin{aligned}
&\sum_{t=0}^{T-1} \delta_t 
\leq  \frac{4 \kappa}{L \eta_\mathbf{y}} \delta_0+\frac{10 \eta_\mathbf{y}}{\mu} \sum_{t=1}^{T-1}\left(1-\frac{\mu \eta_\mathbf{y}}{4}\right)^{T-t-1} \sum_{s=0}^{t-1}\left\|\bar{\mathbf{u}}_s-\frac{1}{m} \sum_{i=1}^m \nabla f_i\left(\mathbf{x}_s^{(i)}, \mathbf{y}_s^{(i)}\right)\right\|^2+\frac{40 \kappa^2}{m} \sum_{t=0}^{T-1}\\
&\left(\left\|X_t-\bar{X}_t\right\|_F^2+\left\|Y_t-\bar{Y}_t\right\|_F^2\right)+\frac{20 \kappa^4 \eta_\mathbf{x}^2}{L^2 \eta_\mathbf{y}^2} \sum_{t=0}^{T-1}\left\|\bar{\mathbf{v}}_t\right\|^2-\frac{14 \eta_\mathbf{y}}{5 \mu} \sum_{t=0}^{T-1}\left(1-\left(1-\frac{\mu \eta_\mathbf{y}}{4}\right)^{T-t}\right)\left\|\bar{\mathbf{u}}_t\right\|^2
\end{aligned}
\end{equation}
\textbf{Lemma \refstepcounter{lemma}\label{lemma 7}7}. For all $t \in\{0,1, \cdots, T\}$ we have $\bar{\mathbf{v}}_t=\bar{\mathbf{g}}_t^*$ and $\bar{\mathbf{u}}_t=\bar{\mathbf{h}}_t^*$.

\textit{Proof:} As matrix W is doubly stochastic, we have:
\begin{equation}
\begin{aligned}
\bar{\mathbf{v}}_t=\bar{\mathbf{v}}_{t-1}+\bar{\mathbf{g}}_t^*-\bar{\mathbf{g}}_{t-1}^*
\end{aligned}
\end{equation}
which is equivalent to $\bar{\mathbf{v}}_t-\bar{\mathbf{g}}_t^*=\bar{\mathbf{v}}_{t-1}-\bar{\mathbf{g}}_{t-1}^*$. Additionally, $\bar{\mathbf{v}}_{-1}=\bar{\mathbf{g}}_{-1}^*$, so $\bar{\mathbf{v}}_t=\bar{\mathbf{g}}_t^*$.
Thus, from the above: $\bar{\mathbf{v}}_t = \bar{\mathbf{g}}_t+ \mathcal{N}_{\mathbf{x}, t}$.

\textbf{Lemma \refstepcounter{lemma}\label{lemma 8}8}. Let $A_t, B_t$ be positive sequences satisfying
\begin{equation}
\begin{aligned}
\label{eq.8.1}
A_{t+1} \leq(1-c) A_t+B_t
\end{aligned}
\end{equation}
for some constant $c \in(0,1)$. Then for any positive integer $T$ we have
\begin{equation}
\begin{aligned}
\label{eq8.2}
\sum_{t=0}^T A_t \leq \frac{1}{c} A_0+\frac{1}{c} \sum_{t=0}^{T-1} B_t
\end{aligned}
\end{equation}
\textit{Proof}: Using recursion on Eq.(\ref{eq.8.1}) we can obtain
\begin{equation}
\begin{aligned}
A_t \leq(1-c)^t A_0+\sum_{s=0}^{t-1}(1-c)^{t-s-1} B_s
\end{aligned}
\end{equation}
for $\forall t \geq 0$. Sum above inequality and we achieve the desired conclusion Eq.(\ref{eq8.2}), where we use the condtion $A_t, B_t$ are positive and the fact that $\sum_{t=0}^{\infty}(1-c)^t=\frac{1}{c}$.

\textbf{Lemma \refstepcounter{lemma}\label{lemma 9}9}. We can prove the following bound for gradient estimator $\bar{\mathbf{v}}_t$ and $\bar{\mathbf{u}}_t$.
\begin{equation}
\begin{aligned}
&\sum_{s=0}^{t-1} \mathbb{E}\left\|\bar{\mathbf{v}}_s-\frac{1}{m} \sum_{i=1}^m \nabla_\mathbf{x} f_i\left(\mathbf{x}_s^{(i)}, \mathbf{y}_s^{(i)}\right)\right\|^2 \\
& \leq \frac{2 \sigma^2}{m b_0 \beta_\mathbf{x}}+\frac{2 \beta_\mathbf{x} \sigma^2 t}{m}+\frac{12 L^2}{m^2 \beta_\mathbf{x}} \sum_{s=0}^{t-1}\left(\mathbb{E}\left\|X_s-\bar{X}_s\right\|_F^2+\mathbb{E}\left\|Y_s-\bar{Y}_s\right\|_F^2\right)+\frac{6 L^2}{m \beta_\mathbf{x}} \sum_{s=0}^{t-2}\\
&\quad \left(\eta_\mathbf{x}^2 \mathbb{E}\left\|\bar{\mathbf{v}}_s\right\|^2+\eta_\mathbf{y}^2 \mathbb{E}\left\|\bar{\mathbf{u}}_s\right\|^2\right) +2 \sum_{s=0}^{t-1} \mathbb{E}\left\|\mathcal{N}_{\mathbf{x},s}-\mathcal{N}_{\mathbf{x},s-1}\right\|^2\\
&\sum_{s=0}^{t-1} \mathbb{E}\left\|\bar{\mathbf{u}}_s-\frac{1}{m} \sum_{i=1}^m \nabla_\mathbf{y} f_i\left(\mathbf{x}_s^{(i)}, \mathbf{y}_s^{(i)}\right)\right\|^2 \\
& \leq \frac{2 \sigma^2}{m b_0 \beta_\mathbf{y}}+\frac{2 \beta_\mathbf{y} \sigma^2 t}{m}+\frac{12 L^2}{m^2 \beta_\mathbf{y}} \sum_{s=0}^{t-1}\left(\mathbb{E}\left\|X_s-\bar{X}_s\right\|_F^2+\mathbb{E}\left\|Y_s-\bar{Y}_s\right\|_F^2\right)+\frac{6 L^2}{m \beta_\mathbf{y}}\sum_{s=0}^{t-2} \\
&\quad \left(\eta_\mathbf{x}^2 \mathbb{E}\left\|\bar{\mathbf{v}}_s\right\|^2+\eta_\mathbf{y}^2 \mathbb{E}\left\|\bar{\mathbf{u}}_s\right\|^2\right) +2 \sum_{s=0}^{t-1} \mathbb{E}\left\|\mathcal{N}_{\mathbf{y},s}-\mathcal{N}_{\mathbf{y},s-1}\right\|^2
\end{aligned}
\end{equation}
for all $t \in\{1,2, \cdots, T\}$.\\\\
\textit{Proof:} By the definition of $\mathbf{g}_t^{(i)}$ and Lemma \ref{lemma 7}, now we have
\begin{equation}
\begin{aligned}
&\bar{\mathbf{v}}_t - \mathcal{N}_{\mathbf{x}, t} - \frac{1}{m} \sum_{i=1}^m \nabla_\mathbf{x} f_i\left(\mathbf{x}_t^{(i)}, \mathbf{y}_t^{(i)}\right) \\
&= \left(1 - \beta_\mathbf{x}\right) \left(\bar{\mathbf{v}}_{t-1} -\mathcal{N}_{\mathbf{x},t-1}- \frac{1}{m} \sum_{i=1}^m \nabla_\mathbf{x} f_i\left(\mathbf{x}_{t-1}^{(i)}, \mathbf{y}_{t-1}^{(i)}\right)\right)  + \frac{\beta_\mathbf{x}}{m} \sum_{i=1}^m \left(\nabla_\mathbf{x} F_i\left(\mathbf{x}_t^{(i)}, \mathbf{y}_t^{(i)} ; \mathbf{z}_t^{(i)}\right) \right.\\
&\left. \quad - \nabla_\mathbf{x} f_i\left(\mathbf{x}_t^{(i)}, \mathbf{y}_t^{(i)}\right)\right) + \left(1 - \beta_\mathbf{x}\right) \frac{1}{m} \sum_{i=1}^m \left(\nabla_\mathbf{x} F_i\left(\mathbf{x}_t^{(i)}, \mathbf{y}_t^{(i)} ; \mathbf{z}_t^{(i)}\right) - \nabla_\mathbf{x} F_i\left(\mathbf{x}_{t-1}^{(i)}, \mathbf{y}_{t-1}^{(i)} ; \mathbf{z}_t^{(i)}\right) \right. \\
&\quad \left. + \nabla_\mathbf{x} f_i\left(\mathbf{x}_{t-1}^{(i)}, \mathbf{y}_{t-1}^{(i)}\right) - \nabla_\mathbf{x} f_i\left(\mathbf{x}_t^{(i)}, \mathbf{y}_t^{(i)}\right)\right)
\end{aligned}
\end{equation}
Taking expectation $\left(\mathbf{z}_t^{(i)}\right)$ , the last two terms of Equation above are 0. Therefore, By using Cauchy-Schwarz inequality, we obtain:
\begin{equation}
\begin{aligned}
&\mathbb{E} \left\|\bar{\mathbf{v}}_t - \frac{1}{m} \sum_{i=1}^m \nabla_\mathbf{x} f_i\left(\mathbf{x}_t^{(i)}, \mathbf{y}_t^{(i)}\right)\right\|^2 \\
&\leq \mathbb{E}\left\|\left(1-\beta_\mathbf{x}\right)\left(\bar{\mathbf{v}}_{t-1} - \frac{1}{m} \sum_{i=1}^m \nabla_\mathbf{x} f_i\left(\mathbf{x}_{t-1}^{(i)}, \mathbf{y}_{t-1}^{(i)}\right)\right)+ \mathcal{N}_{\mathbf{x}, t} - \left(1 - \beta_\mathbf{x}\right)\mathcal{N}_{\mathbf{x}, t-1} \right\|^2 \\
&\quad + \mathbb{E} \bigg\| \frac{\beta_\mathbf{x}}{m} \sum_{i=1}^m \left(\nabla_\mathbf{x} F_i\left(\mathbf{x}_t^{(i)}, \mathbf{y}_t^{(i)}; \mathbf{z}_t^{(i)}\right) - \nabla_\mathbf{x} f_i\left(\mathbf{x}_t^{(i)}, \mathbf{y}_t^{(i)}\right)\right) \\
& \quad + 2\left(1-\beta_\mathbf{x}\right) \frac{1}{m} \sum_{i=1}^m \bigg(\nabla_\mathbf{x} F_i\left(\mathbf{x}_t^{(i)}, \mathbf{y}_t^{(i)}; \mathbf{z}_t^{(i)}\right) - \nabla_\mathbf{x} F_i\left(\mathbf{x}_{t-1}^{(i)}, \mathbf{y}_{t-1}^{(i)}; \mathbf{z}_t^{(i)}\right)  \\
& \quad + \nabla_\mathbf{x} f_i\left(\mathbf{x}_{t-1}^{(i)}, \mathbf{y}_{t-1}^{(i)}\right) - \nabla_\mathbf{x} f_i\left(\mathbf{x}_t^{(i)}, \mathbf{y}_t^{(i)}\right)\bigg)\bigg\|^2 \\
&\leq  2\left(1 - \beta_\mathbf{x}\right)^2 \mathbb{E} \left\|\bar{\mathbf{v}}_{t-1} - \frac{1}{m} \sum_{i=1}^m \nabla_\mathbf{x} f_i\left(\mathbf{x}_{t-1}^{(i)}, \mathbf{y}_{t-1}^{(i)}\right)\right\|^2 + \frac{2 \beta_\mathbf{x}^2}{m^2} \sum_{i=1}^m \mathbb{E} \left\| \nabla_\mathbf{x} F_i\left(\mathbf{x}_t^{(i)}, \mathbf{y}_t^{(i)}; \mathbf{z}_t^{(i)}\right) \right. \\
& \left.\quad - \nabla_\mathbf{x} f_i\left(\mathbf{x}_t^{(i)}, \mathbf{y}_t^{(i)}\right)\right\|^2 + \frac{2 \left(1 - \beta_\mathbf{x}\right)^2}{m^2} \sum_{i=1}^m \mathbb{E} \left\| \nabla_\mathbf{x} F_i\left(\mathbf{x}_t^{(i)}, \mathbf{y}_t^{(i)}; \mathbf{z}_t^{(i)}\right) \right.  - \nabla_\mathbf{x} F_i\left(\mathbf{x}_{t-1}^{(i)}, \mathbf{y}_{t-1}^{(i)}; \mathbf{z}_t^{(i)}\right)\\
& \left.\quad+ \nabla_\mathbf{x} f_i\left(\mathbf{x}_{t-1}^{(i)}, \mathbf{y}_{t-1}^{(i)}\right) - \nabla_\mathbf{x} f_i\left(\mathbf{x}_t^{(i)}, \mathbf{y}_t^{(i)}\right)\right\|^2  + 2 \mathbb{E} \left\|\mathcal{N}_{\mathbf{x}, t} -  \left(1 - \beta_\mathbf{x}\right)\mathcal{N}_{\mathbf{x}, t-1}\right\|^2
\end{aligned}
\end{equation}
The first inequality is obtained by Cauchy-Schwartz inequality. Then we use Lemma \ref{lemma 3} on the last two terms, and then use Assumption \ref{assumption 2}, Lemma \ref{lemma 2} and Assumption \ref{assumption 1}, we can obtain.
\begin{equation}
\begin{aligned}
\label{eq9.1}
&\mathbb{E} \left\|\bar{\mathbf{v}}_t - \frac{1}{m} \sum_{i=1}^m \nabla_\mathbf{x} f_i\left(\mathbf{x}_t^{(i)}, \mathbf{y}_t^{(i)}\right)\right\|^2\\
&\leq  2 \left(1 - \beta_\mathbf{x}\right)^2 \mathbb{E} \left\|\bar{\mathbf{v}}_{t-1} - \frac{1}{m} \sum_{i=1}^m \nabla_\mathbf{x} f_i\left(\mathbf{x}_{t-1}^{(i)}, \mathbf{y}_{t-1}^{(i)}\right)\right\|^2 + 2 \mathbb{E} \left\|\mathcal{N}_{\mathbf{x}, t} -  \left(1 - \beta_\mathbf{x}\right)\mathcal{N}_{\mathbf{x}, t-1}\right\|^2 \\
&\quad + \frac{2 \beta_\mathbf{x}^2 \sigma^2}{m} + \frac{2 L^2 \left(1 - \beta_\mathbf{x}\right)^2}{m^2} \left(\mathbb{E} \left\|X_t - X_{t-1}\right\|_F^2 + \mathbb{E} \left\|Y_t - Y_{t-1}\right\|_F^2 \right)
\end{aligned}
\end{equation}
 At the same time, by using Cauchy-Schwarz inequality, we have a rewritten form for $X_t - X_{t-1}$, $Y_t - Y_{t-1}$.
\begin{equation}
\begin{aligned}
\label{eq9.2}
& \left\|X_t-X_{t-1}\right\|_F^2 \leq 3\left\|X_t-\bar{X}_{t}\right\|_F^2+3 m \eta_\mathbf{x}^2\left\|\bar{\mathbf{v}}_{t-1}\right\|^2+3\left\|X_{t-1}-\bar{X}_{t-1}\right\|_F^2 \\
& \left\|Y_t-Y_{t-1}\right\|_F^2 \leq 3\left\|Y_t-\bar{Y}_t\right\|_F^2+3 m \eta_\mathbf{y}^2\left\|\bar{\mathbf{u}}_{t-1}\right\|^2+3\left\|Y_{t-1}-\bar{Y}_{t-1}\right\|_F^2 \\
\end{aligned}
\end{equation}
Combining above two inequalities with Eq.(\ref{eq9.1}) and Lemma \ref{lemma 8}, we have:
\begin{equation}
\begin{aligned}
& \sum_{s=0}^{t-1} \mathbb{E}\left\|\bar{\mathbf{v}}_s - \frac{1}{m} \sum_{i=1}^m \nabla_\mathbf{x} f_i\left(\mathbf{x}_s^{(i)}, \mathbf{y}_s^{(i)}\right)\right\|^2 \\
&\leq \frac{2}{\beta_\mathbf{x}} \mathbb{E}\left\|\bar{\mathbf{v}}_0 - \nabla_\mathbf{x} f\left(\mathbf{x}_0, \mathbf{y}_0\right)\right\|^2 + \frac{2 \beta_\mathbf{x} \sigma^2 t}{m}  + \frac{12 L^2}{m^2 \beta_\mathbf{x}} \sum_{s=0}^{t-1} \left( \mathbb{E}\left\|X_s - \bar{X}_s\right\|_F^2 + \mathbb{E}\left\|Y_s - \bar{Y}_s\right\|_F^2\right) \\
& \quad+ \frac{6 L^2}{m \beta_\mathbf{x}} \sum_{s=0}^{t-2} \left(\eta_\mathbf{x}^2 \mathbb{E}\left\|\bar{\mathbf{v}}_s\right\|^2 + \eta_\mathbf{y}^2 \mathbb{E}\left\|\bar{\mathbf{u}}_s\right\|^2\right) +2 \sum_{s=0}^{t-1} \mathbb{E}\left\|\mathcal{N}_{\mathbf{x},s} -  \left(1 - \beta_\mathbf{x}\right)\mathcal{N}_{\mathbf{x},s-1}\right\|^2 \\
& \leq \frac{2 \sigma^2}{m b_0 \beta_\mathbf{x}} + \frac{2 \beta_\mathbf{x} \sigma^2 t}{m} + \frac{12 L^2}{m^2 \beta_\mathbf{x}} \sum_{s=0}^{t-1} \left( \mathbb{E}\left\|X_s - \bar{X}_s\right\|_F^2 + \mathbb{E}\left\|Y_s - \bar{Y}_s\right\|_F^2\right)  \\
&\quad + \frac{6 L^2}{m \beta_\mathbf{x}} \sum_{s=0}^{t-2}\left(\eta_\mathbf{x}^2 \mathbb{E}\left\|\bar{\mathbf{v}}_s\right\|^2 + \eta_\mathbf{y}^2 \mathbb{E}\left\|\bar{\mathbf{u}}_s\right\|^2\right) + 2 \sum_{s=0}^{t-1} \mathbb{E}\left\|\mathcal{N}_{\mathbf{x},s} -  \left(1 - \beta_\mathbf{x}\right)\mathcal{N}_{\mathbf{x},s-1}\right\|^2
\end{aligned}
\end{equation}
for all $t \in\{1,2, \cdots, T\}$. In the first inequality we use the fact $\frac{1}{1-\left(1-\beta_\mathbf{x}\right)^2} \leq \frac{1}{\beta_\mathbf{x}}$ when $\beta_\mathbf{x} \leq 1$. Where $\mathbb{E}\left\|\bar{\mathbf{v}}_0-\nabla_\mathbf{x} f\left(\mathbf{x}_0, \mathbf{y}_0\right)\right\|^2 \leq \frac{\sigma^2}{m b_0}$ holds because of Assumption \ref{assumption 2} and Lemma \ref{lemma 3}. Mimic above steps we can also prove the second conclusion.

\textbf{Lemma\refstepcounter{lemma}\label{lemma 10} 10}. The consensus error satisfies the following recursive relation
\begin{equation}
\begin{aligned}
\left\|X_{t+1}-\bar{X}_{t+1}\right\|_F^2 & \leq \frac{1+\lambda^2}{2}\left\|X_t-\bar{X}_t\right\|_F^2+\frac{2 \lambda^2 \eta_\mathbf{x}^2}{1-\lambda^2}\left\|V_t-\bar{V}_t\right\|_F^2 \\
\left\|Y_{t+1}-\bar{Y}_{t+1}\right\|_F^2 & \leq \frac{1+\lambda^2}{2}\left\|Y_t-\bar{Y}_t\right\|_F^2+\frac{2 \lambda^2 \eta_\mathbf{y}^2}{1-\lambda^2}\left\|U_t-\bar{U}_t\right\|_F^2
\end{aligned}
\end{equation}
\textit{Proof.} As we set $J=\frac{11^T}{n}$, then we obtain:
\begin{equation}
\begin{aligned}
\label{eq10.1}
\left\|X_{t+1}-\bar{X}_{t+1}\right\|_F^2 
& = \left\|\left(X_t-\eta_\mathbf{x} V_t\right) W-\left(\bar{X}_t-\eta_\mathbf{x} \bar{V}_t\right)\right\|_F^2\\
&=\left\|\left(X_t-\bar{X}_t\right)(W-J)-\eta_\mathbf{x}\left(V_t-\bar{V}_t\right)(W-J)\right\|_F^2 \\
\end{aligned}
\end{equation}
Now, we use Lemma \ref{lemma 4} and Assumption \ref{assumption 4} on Eq.(\ref{eq10.1})
\begin{equation}
\begin{aligned}
\label{eq10.2}
&\left\|X_{t+1}-\bar{X}_{t+1}\right\|_F^2 \\
&\leq  \lambda^2\left\|X_t-\bar{X}_t\right\|_F^2+\lambda^2 \eta_\mathbf{x}^2\left\|V_t-\bar{V}_t\right\|_F^2-2\left\langle\left(X_t-\bar{X}_t\right)(W-J), \eta_\mathbf{x}\left(V_t-\bar{V}_t\right)(W-J)\right\rangle
\end{aligned}
\end{equation}
We use the Young's inequality to eliminate the last term above, and we set the constant $\mathbf{\alpha}=\frac{1-\lambda^2}{2 \lambda^2}$.
\begin{equation}
\begin{aligned}
\left\|X_{t+1}-\bar{X}_{t+1}\right\|_F^2 \leq & \left(\lambda^2+\mathbf{\alpha}\lambda^2\right)\left\|X_t-\bar{X}_t\right\|_F^2+\left(\frac{\lambda^2 \eta_\mathbf{x}^2}{\mathbf{\alpha}}+\lambda^2 \eta_\mathbf{x}^2\right)\left\|V_t-\bar{V}_t\right\|_F^2 \\
\leq & \frac{1+\lambda^2}{2}\left\|X_t-\bar{X}_t\right\|_F^2+\frac{2 \lambda^2 \eta_\mathbf{x}^2}{1-\lambda^2}\left\|V_t-\bar{V}_t\right\|_F^2
\end{aligned}
\end{equation}
Mimic the steps above, we can also get:
\begin{equation}
\begin{aligned}
\left\|Y_{t+1}-\bar{Y}_{t+1}\right\|_F^2 \leq \frac{1+\lambda^2}{2}\left\|Y_t-\bar{Y}_t\right\|_F^2+\frac{2 \lambda^2 \eta_\mathbf{y}^2}{1-\lambda^2}\left\|U_t-\bar{U}_t\right\|_F^2
\end{aligned}
\end{equation}

\textbf{Lemma\refstepcounter{lemma}\label{lemma 11} 11}. For all $t^\prime \in\{0,1, \cdots, T-1\}$ we have 
\begin{equation}
\begin{aligned}
&\sum_{s=0}^{t^{\prime}} \mathbb{E}\left\|V_s-\bar{V}_s\right\|_F^2 \\
&\leq \frac{2}{1-\lambda^2} \mathbb{E}\left\|V_0-\bar{V}_0\right\|_F^2+\frac{48 \lambda^2 L^2}{\left(1-\lambda^2\right)^2} \sum_{s=0}^{t^{\prime}}\left(\mathbb{E}\left\|X_s-\bar{X}_s\right\|_F^2+\mathbb{E}\left\|Y_s-\bar{Y}_s\right\|_F^2\right)\\
&\quad +\frac{24 m \lambda^2 L^2}{\left(1-\lambda^2\right)^2} \sum_{s=0}^{t^{\prime}-1} \eta_\mathbf{y}^2 \mathbb{E}\left\|\bar{\mathbf{u}}_s\right\|^2+\frac{24 m \lambda^2 L^2}{\left(1-\lambda^2\right)^2} \sum_{s=0}^{t^{\prime}-1} \eta_\mathbf{x}^2 \mathbb{E}\left\|\bar{\mathbf{v}}_s\right\|^2 +\frac{8 m \lambda^2 \beta_\mathbf{x}^2 \sigma^2 t^{\prime}}{1-\lambda^2}\\
&\quad +\frac{8 \lambda^2 \beta_\mathbf{x}^2}{\left(1-\lambda^2\right)^2} \sum_{s=0}^{t^{\prime}-1} \sum_{i=1}^m \mathbb{E}\left\|\mathbf{g}_s^{(i)}-\nabla_\mathbf{x} f_i\left(\mathbf{x}_s^{(i)}, \mathbf{y}_s^{(i)}\right)\right\|^2
+\frac{8 \lambda^2 m}{1-\lambda^2} \sum_{s=0}^{t^{\prime}}\mathbb{E}\left\|\mathcal{N}_{\mathbf{x},s}-\mathcal{N}_{\mathbf{x},s-1}\right\|^2\\
&\sum_{s=0}^{t^{\prime}} \mathbb{E}\left\|U_s-\bar{U}_s\right\|_F^2 \\
&\leq \frac{2}{1-\lambda^2} \mathbb{E}\left\|U_0-\bar{U}_0\right\|_F^2+\frac{48 \lambda^2 L^2}{\left(1-\lambda^2\right)^2} \sum_{s=0}^{t^{\prime}}\left(\mathbb{E}\left\|X_s-\bar{X}_s\right\|_F^2+\mathbb{E}\left\|Y_s-\bar{Y}_s\right\|_F^2\right)\\
&\quad +\frac{24 m \lambda^2 L^2}{\left(1-\lambda^2\right)^2} \sum_{s=0}^{t^{\prime}-1} \eta_\mathbf{y}^2 \mathbb{E}\left\|\bar{\mathbf{u}}_s\right\|^2+\frac{24 m \lambda^2 L^2}{\left(1-\lambda^2\right)^2} \sum_{s=0}^{t^{\prime}-1} \eta_\mathbf{x}^2 \mathbb{E}\left\|\bar{\mathbf{v}}_s\right\|^2 +\frac{8 m \lambda^2 \beta_\mathbf{y}^2 \sigma^2 t^{\prime}}{1-\lambda^2}\\
&\quad +\frac{8 \lambda^2 \beta_\mathbf{y}^2}{\left(1-\lambda^2\right)^2} \sum_{s=0}^{t^{\prime}-1} \sum_{i=1}^m \mathbb{E}\left\|\mathbf{h}_s^{(i)}-\nabla_\mathbf{y} f_i\left(\mathbf{x}_s^{(i)}, \mathbf{y}_s^{(i)}\right)\right\|^2
+\frac{8 \lambda^2 m}{1-\lambda^2} \sum_{s=0}^{t^{\prime}}\mathbb{E}\left\|\mathcal{N}_{\mathbf{y},s}-\mathcal{N}_{\mathbf{y},s-1}\right\|^2
\end{aligned}
\end{equation}
\textit{Proof}: Similar as Eq.(\ref{eq10.1}) and (\ref{eq10.2}), by definition of $V_t$, we obtain:
\begin{equation}
\begin{aligned}
\label{eq11.1}
&\left\|V_{t+1}-\bar{V}_{t+1}\right\|_F^2 \\
&\leq  \lambda^2\left\|V_t-\bar{V}_t\right\|_F^2+\lambda^2\left\|G_{t+1}^*-G_t^*\right\|_F^2+2\left\langle\left(V_t-\bar{V}_t\right)(W-J),\left(G_{t+1}^*-G_t^*\right)(W-J)\right\rangle
\end{aligned}
\end{equation}
By the definition of $\mathbf{g}_t^{(i)*}$:
\begin{equation}
\begin{aligned}
\label{eq11.2}
&\mathbf{g}_{t+1}^{(i)*}-\mathbf{g}_t^{(i)*}\\
&=  \nabla_\mathbf{x} F_i\left(\mathbf{x}_{t+1}^{(i)}, \mathbf{y}_{t+1}^{(i)} ; \mathbf{z}_{t+1}^{(i)}\right)-\nabla_\mathbf{x} F_i\left(\mathbf{x}_t^{(i)}, \mathbf{y}_t^{(i)} ; \mathbf{z}_{t+1}^{(i)}\right)-\beta_\mathbf{x}\left(\mathbf{g}_t^{(i)}-\nabla_\mathbf{x} f_i\left(\mathbf{x}_t^{(i)}, \mathbf{y}_t^{(i)}\right)\right) \\
& \quad+\beta_\mathbf{x}\left(\nabla_\mathbf{x} F_i\left(\mathbf{x}_t^{(i)}, \mathbf{y}_t^{(i)} ; \mathbf{z}_{t+1}^{(i)}\right)-\nabla_\mathbf{x} f_i\left(\mathbf{x}_t^{(i)}, \mathbf{y}_t^{(i)}\right)\right)+n_{\mathbf{x},t+1}^{(i)}-n_{\mathbf{x},t}^{(i)}
\end{aligned}
\end{equation}
Because of $\mathbb{E}\left[n_{\mathbf{x},t+1}^{(i)}-n_{\mathbf{x},t}^{(i)}\right]=0-0=0$, thus
\begin{equation}
\begin{aligned}
\mathbb{E}\left[\mathbf{g}_{t+1}^{(i)^*} - \mathbf{g}_t^{(i)^*}\right]=\nabla_\mathbf{x} f_i\left(\mathbf{x}_{t+1}^{(i)}, \mathbf{y}_{t+1}^{(i)}\right)-\nabla_\mathbf{x} f_i\left(\mathbf{x}_t^{(i)}, \mathbf{y}_t^{(i)}\right)-\beta_\mathbf{x}\left(\mathbf{g}_t^{(i)}-\nabla_\mathbf{x} f_i\left(\mathbf{x}_t^{(i)}, \mathbf{y}_t^{(i)}\right)\right)   
\end{aligned}
\end{equation}
Considering about all the agents, we get:
\begin{equation}
\left\|\mathbb{E}\left[G_{t+1}^*-G_t^*\right]\right\|^2 = \sum_{i=1}^m\bigg\|\nabla_\mathbf{x} f_i\left(\mathbf{x}_{t+1}^{(i)}, \mathbf{y}_{t+1}^{(i)}\right)-\nabla_\mathbf{x} f_i\left(\mathbf{x}_t^{(i)}, \mathbf{y}_t^{(i)}\right)-\beta_\mathbf{x}\left(\mathbf{g}_t^{(i)}-\nabla_\mathbf{x} f_i\left(\mathbf{x}_t^{(i)}, \mathbf{y}_t^{(i)}\right)\right)\bigg\|^2
\end{equation}
Taking expectation on $\mathbf{z}_{t+1}^{(i)}$ the last term of Eq.(\ref{eq11.1}) can be bounded by
\begin{equation}
\begin{aligned}
\label{eq11.3}
& \mathbb{E}\left\langle\left(V_t-\bar{V}_t\right)(W-J),\left(G_{t+1}^*-G_t^*\right)(W-J)\right\rangle \\
&=  \left\langle\left(V_t-\bar{V}_t\right)(W-J), \mathbb{E}\left[G_{t+1}^*-G_t^*\right](W-J)\right\rangle \leq \lambda\left\|V_t-\bar{V}_t\right\|_F \cdot \lambda\left\|\mathbb{E}\left[G_{t+1}^*-G_t^*\right]\right\|_F \\
&\leq  \frac{1-\lambda^2}{4}\left\|V_t-\bar{V}_t\right\|_F^2+\frac{\lambda^4}{1-\lambda^2}\left\|\mathbb{E}\left[G_{t+1}^*-G_t^*\right]\right\|_F^2 \\
&\leq  \frac{1-\lambda^2}{4}\left\|V_t-\bar{V}_t\right\|_F^2+\frac{2 \lambda^4}{1-\lambda^2} \sum_{i=1}^m\left\|\nabla_\mathbf{x} f_i\left(\mathbf{x}_{t+1}^{(i)}, \mathbf{y}_{t+1}^{(i)}\right)-\nabla_\mathbf{x} f_i\left(\mathbf{x}_t^{(i)}, \mathbf{y}_t^{(i)}\right)\right\|^2 \\
& \quad +\frac{2 \lambda^4 \beta_\mathbf{x}^2}{1-\lambda^2} \sum_{i=1}^m\left\|\mathbf{g}_t^{(i)}-\nabla_\mathbf{x} f_i\left(\mathbf{x}_t^{(i)}, \mathbf{y}_t^{(i)}\right)\right\|^2 \\
&\leq  \frac{1-\lambda^2}{4}\left\|V_t-\bar{V}_t\right\|_F^2+\frac{2 \lambda^4 L^2}{1-\lambda^2}\left(\left\|X_{t+1}-X_t\right\|_F^2+\left\|Y_{t+1}-Y_t\right\|_F^2\right) \\
&\quad +\frac{2 \lambda^4 \beta_\mathbf{x}^2}{1-\lambda^2} \sum_{i=1}^m\left\|\mathbf{g}_t^{(i)}-\nabla_\mathbf{x} f_i\left(\mathbf{x}_t^{(i)}, \mathbf{y}_t^{(i)}\right)\right\|^2
\end{aligned}
\end{equation}
Where we use Young's inequality in the second inequality, and then we use Cauchy-Schwartz inequality in the third inequality, and the last inequality is resulted from Assumption \ref{assumption 1}. Besides, applying Cauchy-Schwartz inequality to Eq.(\ref{eq11.2}) we have:
\begin{equation}
\begin{aligned}
\label{eq11.4}
& \mathbb{E}\left\|\mathbf{g}_{t+1}^{(i)^*}-\mathbf{g}_t^{(i)^*}\right\|^2 \\
&\leq 4 \mathbb{E}\left\|\nabla_\mathbf{x} F_i\left(\mathbf{x}_{t+1}^{(i)}, \mathbf{y}_{t+1}^{(i)} ; \mathbf{z}_{t+1}^{(i)}\right)-\nabla_\mathbf{x} F_i\left(\mathbf{x}_t^{(i)}, \mathbf{y}_t^{(i)} ; \mathbf{z}_{t+1}^{(i)}\right)\right\|^2+4 \beta_\mathbf{x}^2 \mathbb{E}\left\|\mathbf{g}_t^{(i)} -\nabla_\mathbf{x} f_i\left(\mathbf{x}_t^{(i)}, \mathbf{y}_t^{(i)}\right)\right\|^2\\
&\quad +4 \beta_\mathbf{x}^2 \mathbb{E}\left\|\nabla_\mathbf{x} F_i\left(\mathbf{x}_t^{(i)}, \mathbf{y}_t^{(i)} ; \mathbf{z}_{t+1}^{(i)}\right)-\nabla_\mathbf{x} f_i\left(\mathbf{x}_t^{(i)}, \mathbf{y}_t^{(i)}\right)\right\|^2+4 \mathbb{E}\left\|n_{\mathbf{x},t+1}^{(i)}-n_{\mathbf{x},t}^{(i)}\right\|^2 \\
& \leq 4 L^2\left(\mathbb{E}\left\|\mathbf{x}_{t+1}^{(i)}-\mathbf{x}_t^{(i)}\right\|^2+\mathbb{E}\left\|\mathbf{y}_{t+1}^{(i)}-\mathbf{y}_t^{(i)}\right\|^2\right)+4 \beta_\mathbf{x}^2 \mathbb{E}\left\|\mathbf{g}_t^{(i)}-\nabla_\mathbf{x} f_i\left(\mathbf{x}_t^{(i)}, \mathbf{y}_t^{(i)}\right)\right\|^2+4 \beta_\mathbf{x}^2 \sigma^2\\
&\quad +4 \mathbb{E}\left\|n_{\mathbf{x},t+1}^{(i)}-n_{\mathbf{x},t}^{(i)}\right\|^2
\end{aligned}
\end{equation}
where in the last inequality we use Assumption \ref{assumption 1} and Assumption \ref{assumption 2}. Combining Eq.(\ref{eq11.1}) (\ref{eq11.3}), (\ref{eq11.4}) and the definition of $\mathcal{N}_{\mathbf{x}, t}$ we can obtain
\begin{equation}
\begin{aligned}
&\mathbb{E}\left\|V_{t+1}-\bar{V}_{t+1}\right\|_F^2 \\
&\leq  \frac{1+\lambda^2}{2} \mathbb{E}\left\|V_t-\bar{V}_t\right\|_F^2+\frac{4 \lambda^4 L^2}{1-\lambda^2}\left(\mathbb{E}\left\|X_{t+1}-X_t\right\|_F^2+\mathbb{E}\left\|Y_{t+1}-Y_t\right\|_F^2\right) \\
& \quad+\frac{4 \lambda^4 \beta_\mathbf{x}^2}{1-\lambda^2} \sum_{i=1}^m \mathbb{E}\left\|\mathbf{g}_t^{(i)}-\nabla_\mathbf{x} f_i\left(\mathbf{x}_t^{(i)}, \mathbf{y}_t^{(i)}\right)\right\|^2+4 L^2\lambda^2\left(\mathbb{E}\left\|X_{t+1}-X_t\right\|^2+\mathbb{E}\left\|Y_{t+1}-Y_t\right\|^2\right)\\
& \quad +4\beta_\mathbf{x}^2\lambda^2\sum_{i=1}^m \mathbb{E}\left\|\mathbf{g}_t^{(i)}-\nabla_\mathbf{x} f_i\left(\mathbf{x}_t^{(i)}, \mathbf{y}_t^{(i)}\right)\right\|^2+4 m \lambda^2 \beta_\mathbf{x}^2 \sigma^2+4 \lambda^2 m \mathbb{E}\left\|\mathcal{N}_{\mathbf{x}, t+1}-\mathcal{N}_{\mathbf{x}, t}\right\|^2\\
&= \frac{1+\lambda^2}{2} \mathbb{E}\left\|V_t-\bar{V}_t\right\|_F^2+\frac{\left(4 \lambda^4 L^2 +4 L^2\lambda^2 -4L^2\lambda^4\right)}{1-\lambda^2}\left(\mathbb{E}\left\|X_{t+1}-X_t\right\|_F^2+\mathbb{E}\left\|Y_{t+1}-Y_t\right\|_F^2\right)\\
&\quad +\frac{\left(4 \lambda^4 \beta_\mathbf{x}^2 +4 \beta_\mathbf{x}^2\lambda^2 - 4\beta_\mathbf{x}^2 \lambda^4\right)}{1-\lambda^2}\sum_{i=1}^m \mathbb{E}\left\|\mathbf{g}_t^{(i)}-\nabla_\mathbf{x} f_i\left(\mathbf{x}_t^{(i)}, \mathbf{y}_t^{(i)}\right)\right\|^2+4 m \lambda^2 \beta_\mathbf{x}^2 \sigma^2\\
&\quad +4 \lambda^2 m \mathbb{E}\left\|\mathcal{N}_{\mathbf{x}, t+1}-\mathcal{N}_{\mathbf{x}, t}\right\|^2\\
\end{aligned}
\end{equation}
Then, we have the result below.
\begin{equation}
\begin{aligned}
&\mathbb{E}\left\|V_{t+1}-\bar{V}_{t+1}\right\|_F^2\\
&\leq\frac{1+\lambda^2}{2} \mathbb{E}\left\|V_t-\bar{V}_t\right\|_F^2+\frac{4 \lambda^2 L^2}{1-\lambda^2}\left(\mathbb{E}\left\|X_{t+1}-X_t\right\|_F^2+\mathbb{E}\left\|Y_{t+1}-Y_t\right\|_F^2\right)\\
&\quad +\frac{4 \lambda^2 \beta_\mathbf{x}^2}{1-\lambda^2}\sum_{i=1}^m \mathbb{E}\left\|\mathbf{g}_t^{(i)}-\nabla_\mathbf{x} f_i\left(\mathbf{x}_t^{(i)}, \mathbf{y}_t^{(i)}\right)\right\|^2+4 m \lambda^2 \beta_\mathbf{x}^2 \sigma^2+4 \lambda^2 m \mathbb{E}\left\|\mathcal{N}_{\mathbf{x}, t+1}-\mathcal{N}_{\mathbf{x}, t}\right\|^2
\end{aligned}
\end{equation}
Using Eq.(\ref{eq9.2}) for substitution: 
\begin{equation}
\begin{aligned}
\label{eq11.5}
& \mathbb{E}\left\|V_{t+1}-\bar{V}_{t+1}\right\|_F^2 \\
\leq & \frac{1+\lambda^2}{2} \mathbb{E}\left\|V_t-\bar{V}_t\right\|_F^2+\frac{12 \lambda^2 L^2}{1-\lambda^2}\left(\mathbb{E}\left\|X_{t+1}-\bar{X}_{t+1}\right\|_F^2+\mathbb{E}\left\|Y_{t+1}-\bar{Y}_{t+1}\right\|_F^2\right) \\
& +\frac{12 \lambda^2 L^2}{1-\lambda^2}\left(\mathbb{E}\left\|X_t-\bar{X}_t\right\|_F^2+\mathbb{E}\left\|Y_t-\bar{Y}_t\right\|_F^2\right)+\frac{12 m \lambda^2 L^2 \eta_\mathbf{y}^2}{1-\lambda^2} \mathbb{E}\left\|\bar{\mathbf{u}}_t\right\|^2 +4 m \lambda^2 \beta_\mathbf{x}^2 \sigma^2\\
& +\frac{12 m \lambda^2 L^2 \eta_\mathbf{x}^2}{1-\lambda^2} \mathbb{E}\left\|\bar{\mathbf{v}}_t\right\|^2+\frac{4 \lambda^2 \beta_\mathbf{x}^2}{1-\lambda^2} \sum_{i=1}^m \mathbb{E}\left\|\mathbf{g}_t^{(i)}-\nabla_\mathbf{x} f_i\left(\mathbf{x}_t^{(i)}, \mathbf{y}_t^{(i)}\right)\right\|^2 \\
&+4 \lambda^2 m \mathbb{E}\left\|\mathcal{N}_{\mathbf{x}, t+1}-\mathcal{N}_{\mathbf{x}, t}\right\|^2
\end{aligned}
\end{equation}
Summing over Eq.(\ref{eq11.5}), we obtain: 
\begin{equation}
\begin{aligned}
&\sum_{s=0}^{t^{\prime}} \mathbb{E}\left\|V_s-\bar{V}_s\right\|_F^2 \\
&\leq \frac{2}{1-\lambda^2} \mathbb{E}\left\|V_0-\bar{V}_0\right\|_F^2+\frac{48 \lambda^2 L^2}{\left(1-\lambda^2\right)^2} \sum_{s=0}^{t^{\prime}}\left(\mathbb{E}\left\|X_s-\bar{X}_s\right\|_F^2+\mathbb{E}\left\|Y_s-\bar{Y}_s\right\|_F^2\right)\\
&\quad +\frac{24 m \lambda^2 L^2}{\left(1-\lambda^2\right)^2} \sum_{s=0}^{t^{\prime}-1} \eta_\mathbf{y}^2 \mathbb{E}\left\|\bar{\mathbf{u}}_s\right\|^2+\frac{24 m \lambda^2 L^2}{\left(1-\lambda^2\right)^2} \sum_{s=0}^{t^{\prime}-1} \eta_\mathbf{x}^2 \mathbb{E}\left\|\bar{\mathbf{v}}_s\right\|^2 +\frac{8 m \lambda^2 \beta_\mathbf{x}^2 \sigma^2 t^{\prime}}{1-\lambda^2}\\
&\quad +\frac{8 \lambda^2 \beta_\mathbf{x}^2}{\left(1-\lambda^2\right)^2} \sum_{s=0}^{t^{\prime}-1} \sum_{i=1}^m \mathbb{E}\left\|\mathbf{g}_s^{(i)}-\nabla_\mathbf{x} f_i\left(\mathbf{x}_s^{(i)}, \mathbf{y}_s^{(i)}\right)\right\|^2
+\frac{8 \lambda^2 m}{1-\lambda^2} \sum_{s=0}^{t^{\prime}}\mathbb{E}\left\|\mathcal{N}_{\mathbf{x},s}-\mathcal{N}_{\mathbf{x},s-1}\right\|^2
\end{aligned}
\end{equation}
for all $t^{\prime} \in\{0,1, \cdots, T-1\}$. Here we should notice that term $\mathbb{E}\left\|X_{t+1}-\bar{X}_{t+1}\right\|_F^2$ in Eq.(\ref{eq11.5}) is summed from $\mathbb{E}\left\|X_1-\bar{X}_1\right\|_F^2$ to $\mathbb{E}\left\|X_{t^{\prime}}-\bar{X}_{t^{\prime}}\right\|_F^2$, while term $\mathbb{E}\left\|X_t-\bar{X}_t\right\|_F^2$ is summed from $\mathbb{E}\left\|X_0-\bar{X}_0\right\|_F^2$ to $\mathbb{E}\left\|X_{t^{\prime}-1}-\bar{X}_{t^{\prime}-1}\right\|_F^2$. As $X_0=\bar{X}_0$, these two terms can be merged together. And it is the same with term $\mathbb{E}\left\|Y_{t+1}-\bar{Y}_{t+1}\right\|_F^2$. Mimic above steps and we can prove the conclusion for $\sum_{s=0}^{t^{\prime}} \mathbb{E}\left\|U_s-\bar{U}_s\right\|_F^2$ in the similar way.

\textbf{Lemma \refstepcounter{lemma}\label{lemma 12}12}. We have the local average gradient estimators $\bar{\mathbf{g}}_t$ and $\bar{\mathbf{h}}_t$ satisfy the following conclusion
\begin{equation}
\begin{aligned}
\sum_{s=0}^t \sum_{i=1}^m \mathbb{E}\left\|\mathbf{g}_s^{(i)}-\nabla_\mathbf{x} f_i\left(\mathbf{x}_s^{(i)}, \mathbf{y}_s^{(i)}\right)\right\|^2 \leq & \frac{m \sigma^2}{\beta_\mathbf{x} b_0}+2 m \beta_\mathbf{x} \sigma^2 t+\frac{12 L^2}{\beta_\mathbf{x}} \sum_{s=0}^t\left(\mathbb{E}\left\|X_s-\bar{X}_s\right\|_F^2\right.\\
&\left.+\mathbb{E}\left\|Y_s-\bar{Y}_s\right\|_F^2\right) +\frac{6 m L^2}{\beta_\mathbf{x}} \sum_{s=0}^{t-1}\left(\eta_\mathbf{x}^2 \mathbb{E}\left\|\bar{\mathbf{v}}_s\right\|^2+\eta_\mathbf{y}^2 \mathbb{E}\left\|\bar{\mathbf{u}}_s\right\|^2\right) \\
\sum_{s=0}^t \sum_{i=1}^m \mathbb{E}\left\|\mathbf{h}_s^{(i)}-\nabla_\mathbf{y} f_i\left(\mathbf{x}_s^{(i)}, \mathbf{y}_s^{(i)}\right)\right\|^2 \leq & \frac{m \sigma^2}{\beta_\mathbf{y} b_0}+2 m \beta_\mathbf{y} \sigma^2 t+\frac{12 L^2}{\beta_\mathbf{y}} \sum_{s=0}^t\left(\mathbb{E}\left\|X_s-\bar{X}_s\right\|_F^2\right.\\
&\left. +\mathbb{E}\left\|Y_s-\bar{Y}_s\right\|_F^2\right) +\frac{6 m L^2}{\beta_\mathbf{y}} \sum_{s=0}^{t-1}\left(\eta_\mathbf{x}^2 \mathbb{E}\left\|\bar{\mathbf{v}}_s\right\|^2+\eta_\mathbf{y}^2 \mathbb{E}\left\|\bar{\mathbf{u}}_s\right\|^2\right)
\end{aligned}
\end{equation}
\textit{Proof:} According to the definition of $\mathbf{g}_t^{(i)}$, we have:
\begin{equation}
\begin{aligned}
\label{eq12.1}
&\mathbf{g}_t^{(i)}-\nabla_\mathbf{x} f_i\left(\mathbf{x}_t^{(i)}, \mathbf{y}_t^{(i)}\right) \\
&= \left(1-\beta_\mathbf{x}\right)\left(\mathbf{g}_{t-1}^{(i)}-\nabla_\mathbf{x} f_i\left(\mathbf{x}_{t-1}^{(i)}, \mathbf{y}_{t-1}^{(i)}\right)\right)+\beta_\mathbf{x}\left(\nabla_\mathbf{x} F_i\left(\mathbf{x}_t^{(i)}, \mathbf{y}_t^{(i)} ; \mathbf{z}_t^{(i)}\right)-\nabla_\mathbf{x} f_i\left(\mathbf{x}_t^{(i)}, \mathbf{y}_t^{(i)}\right)\right) \\
&\quad +\left(1-\beta_\mathbf{x}\right)\left(\nabla_\mathbf{x} F_i\left(\mathbf{x}_t^{(i)}, \mathbf{y}_t^{(i)} ; \mathbf{z}_t^{(i)}\right)-\nabla_\mathbf{x} F_i\left(\mathbf{x}_{t-1}^{(i)}, \mathbf{y}_{t-1}^{(i)} ; \mathbf{z}_t^{(i)}\right)\right.\left. +\nabla_\mathbf{x} f_i\left(\mathbf{x}_{t-1}^{(i)}, \mathbf{y}_{t-1}^{(i)}\right)\right.\\
&\left.\quad-\nabla_\mathbf{x} f_i\left(\mathbf{x}_t^{(i)}, \mathbf{y}_t^{(i)}\right)\right)
\end{aligned}
\end{equation}
The last two terms of Eq.(\ref{eq12.1}) will be 0 if we taking expectation of $\mathbf{z}_t^{(i)}$. 
\begin{equation}
\begin{aligned}
\label{eq12.2}
&\mathbb{E}\left\|\left(1-\beta_\mathbf{x}\right)\left(\nabla_\mathbf{x} F_i\left(\mathbf{x}_t^{(i)}, \mathbf{y}_t^{(i)} ; \mathbf{z}_t^{(i)}\right)-\nabla_\mathbf{x} F_i\left(\mathbf{x}_{t-1}^{(i)}, \mathbf{y}_{t-1}^{(i)} ; \mathbf{z}_t^{(i)}\right)\right.\left. +\nabla_\mathbf{x} f_i\left(\mathbf{x}_{t-1}^{(i)}, \mathbf{y}_{t-1}^{(i)}\right)\right.\right.\\
&\left.\left.\quad-\nabla_\mathbf{x} f_i\left(\mathbf{x}_t^{(i)}, \mathbf{y}_t^{(i)}\right)\right)\right\|^2 \\
&\leq 2 \beta_\mathbf{x}^2 \mathbb{E}\left\|\nabla_\mathbf{x} F_i\left(\mathbf{x}_t^{(i)}, \mathbf{y}_t^{(i)} ; \mathbf{z}_t^{(i)}\right)-\nabla_\mathbf{x} f_i\left(\mathbf{x}_t^{(i)}, \mathbf{y}_t^{(i)}\right)\right\|^2 +2\left(1-\beta_\mathbf{x}\right)^2 \mathbb{E}\left\|\nabla_\mathbf{x} F_i\left(\mathbf{x}_t^{(i)}, \mathbf{y}_t^{(i)} ; \mathbf{z}_t^{(i)}\right)\right.\\
&\left.\quad-\nabla_\mathbf{x} F_i\left(\mathbf{x}_{t-1}^{(i)}, \mathbf{y}_{t-1}^{(i)} ; \mathbf{z}_t^{(i)}\right)\right\|^2 
\end{aligned}
\end{equation}
As Eq.(\ref{eq12.2}) is 0, by using  Cauchy-Schwartz we get:
\begin{equation}
\begin{aligned}
& \mathbb{E}\left\|\mathbf{g}_t^{(i)}-\nabla_\mathbf{x} f_i\left(\mathbf{x}_t^{(i)}, \mathbf{y}_t^{(i)}\right)\right\|^2 \\
&\leq  \left(1-\beta_\mathbf{x}\right)^2 \mathbb{E}\left\|\mathbf{g}_{t-1}^{(i)}-\nabla_\mathbf{x} f_i\left(\mathbf{x}_{t-1}^{(i)}, \mathbf{y}_{t-1}^{(i)}\right)\right\|^2+2 \beta_\mathbf{x}^2 \mathbb{E}\left\|\nabla_\mathbf{x} F_i\left(\mathbf{x}_t^{(i)}, \mathbf{y}_t^{(i)} ; \mathbf{z}_t^{(i)}\right)\right.\\
&\left.\quad -\nabla_\mathbf{x} f_i\left(\mathbf{x}_t^{(i)}, \mathbf{y}_t^{(i)}\right)\right\|^2 +2\left(1-\beta_\mathbf{x}\right)^2 \mathbb{E}\left\|\nabla_\mathbf{x} F_i\left(\mathbf{x}_t^{(i)}, \mathbf{y}_t^{(i)} ; \mathbf{z}_t^{(i)}\right)-\nabla_\mathbf{x} F_i\left(\mathbf{x}_{t-1}^{(i)}, \mathbf{y}_{t-1}^{(i)} ; \mathbf{z}_t^{(i)}\right)\right\|^2 \\
&\leq  \left(1-\beta_\mathbf{x}\right)^2 \mathbb{E}\left\|\mathbf{g}_{t-1}^{(i)}-\nabla_\mathbf{x} f_i\left(\mathbf{x}_{t-1}^{(i)}, \mathbf{y}_{t-1}^{(i)}\right)\right\|^2+2 \beta_\mathbf{x}^2 \sigma^2+2\left(1-\beta_\mathbf{x}\right)^2 L^2\left(\mathbb{E}\left\|\mathbf{x}_t^{(i)}-\mathbf{x}_{t-1}^{(i)}\right\|^2\right.\\
&\left.\quad +\mathbb{E}\left\|\mathbf{y}_t^{(i)}-\mathbf{y}_{t-1}^{(i)}\right\|^2\right)
\end{aligned}
\end{equation}
where we use Assumption \ref{assumption 1} and Assumption \ref{assumption 2} in the last inequality. Sum above inequality from $i=1$ to $m$ and we have:
\begin{equation}
\begin{aligned}
\label{eq12.3}
&\sum_{i=1}^m \mathbb{E}\left\|\mathbf{g}_t^{(i)}-\nabla_\mathbf{x} f_i\left(\mathbf{x}_t^{(i)}, \mathbf{y}_t^{(i)}\right)\right\|^2 \\
&\leq  \left(1-\beta_\mathbf{x}\right)^2 \sum_{i=1}^m \mathbb{E}\left\|\mathbf{g}_{t-1}^{(i)}-\nabla_\mathbf{x} f_i\left(\mathbf{x}_{t-1}^{(i)}, \mathbf{y}_{t-1}^{(i)}\right)\right\|^2\\
&\quad+2 m \beta_\mathbf{x}^2 \sigma^2+2\left(1-\beta_\mathbf{x}\right)^2 L^2\left(\mathbb{E}\left\|X_t-X_{t-1}\right\|^2+\mathbb{E}\left\|Y_t-Y_{t-1}\right\|^2\right)\\
&\leq  \left(1-\beta_\mathbf{x}\right)^2 \sum_{i=1}^m \mathbb{E}\left\|\mathbf{g}_{t-1}^{(i)}-\nabla_\mathbf{x} f_i\left(\mathbf{x}_{t-1}^{(i)}, \mathbf{y}_{t-1}^{(i)}\right)\right\|^2\\
&\quad+2 m \beta_\mathbf{x}^2 \sigma^2 +12\left(1-\beta_\mathbf{x}\right)^2 L^2\left(\mathbb{E}\left\|X_t-\bar{X}_t\right\|_F^2+\mathbb{E}\left\|Y_t-\bar{Y}_t\right\|_F^2\right)\\
& \quad+6 m\left(1-\beta_\mathbf{x}\right)^2 L^2\left(\eta_\mathbf{x}^2 \mathbb{E}\left\|\bar{\mathbf{v}}_{t-1}\right\|^2+\eta_\mathbf{y}^2 \mathbb{E}\left\|\bar{\mathbf{u}}_{t-1}\right\|^2\right)
\end{aligned}
\end{equation}
Applying Lemma \ref{lemma 8} to Eq.(\ref{eq12.3}), we have:
\begin{equation}
\begin{aligned}
& \sum_{s=0}^t \sum_{i=1}^m \mathbb{E}\left\|\mathbf{g}_s^{(i)}-\nabla_\mathbf{x} f_i\left(\mathbf{x}_s^{(i)}, \mathbf{y}_s^{(i)}\right)\right\|^2 \\
&\leq  \frac{1}{\beta_\mathbf{x}} \sum_{i=1}^m \mathbb{E}\left\|\mathbf{g}_0^{(i)}-\nabla_\mathbf{x} f_i\left(\mathbf{x}_0^{(i)}, \mathbf{y}_0^{(i)}\right)\right\|^2+\frac{12 L^2}{\beta_\mathbf{x}} \sum_{s=0}^t\left(\mathbb{E}\left\|X_s-\bar{X}_s\right\|_F^2+\mathbb{E}\left\|Y_s-\bar{Y}_s\right\|_F^2\right) \\
&\quad +\frac{6 m L^2}{\beta_\mathbf{x}} \sum_{s=0}^{t-1}\left(\eta_\mathbf{x}^2 \mathbb{E}\left\|\bar{\mathbf{v}}_s\right\|^2+\eta_\mathbf{y}^2 \mathbb{E}\left\|\bar{\mathbf{u}}_s\right\|^2\right)+2 m \beta_\mathbf{x} \sigma^2 t \\
 &\leq \frac{m \sigma^2}{\beta_\mathbf{x} b_0}+2 m \beta_\mathbf{x} \sigma^2 t+\frac{12 L^2}{\beta_\mathbf{x}} \sum_{s=0}^t\left(\mathbb{E}\left\|X_s-\bar{X}_s\right\|_F^2+\mathbb{E}\left\|Y_s-\bar{Y}_s\right\|_F^2\right) \\
&\quad+\frac{6 m L^2}{\beta_\mathbf{x}} \sum_{s=0}^{t-1}\left(\eta_\mathbf{x}^2 \mathbb{E}\left\|\bar{\mathbf{v}}_s\right\|^2+\eta_\mathbf{y}^2 \mathbb{E}\left\|\bar{\mathbf{u}}_s\right\|^2\right)
\end{aligned}
\end{equation}
for all $t \in\{0,1, \cdots, T-1\}$. Here the last inequality is derived by $\mathbb{E}\left\|\mathbf{g}_0^{(i)}-\nabla_\mathbf{x} f_i\left(\mathbf{x}_0^{(i)}, \mathbf{y}_0^{(i)}\right)\right\|^2 \leq \frac{\sigma^2}{b_0}$ due to Lemma \ref{lemma 3}. The estimation of $\mathbf{h}_t^{(i)}$ can be achieved in the same way as above.

\textbf{Lemma \refstepcounter{lemma}\label{lemma 13}13}. Let $\eta_\mathbf{x} \leq \frac{(1-\lambda)^2}{500 L}$ and $\eta_\mathbf{y} \leq \frac{(1-\lambda)^2}{500 L}$. The consensus error can be bounded by
\begin{equation}
\begin{aligned}
& \sum_{s=0}^t\left(\mathbb{E}\left\|X_s-\bar{X}_s\right\|_F^2+\mathbb{E}\left\|Y_s-\bar{Y}_s\right\|_F^2\right) \\
&\leq  \frac{16 \lambda^2 \eta_\mathbf{x}^2}{\left(1-\lambda^2\right)^3} \mathbb{E}\left\|V_0-\bar{V}_0\right\|_F^2+\frac{16 \lambda^2 \eta_\mathbf{y}^2}{\left(1-\lambda^2\right)^3} \mathbb{E}\left\|U_0-\bar{U}_0\right\|_F^2+\frac{576 m \lambda^4 L^2\left(\eta_\mathbf{x}^2+\eta_\mathbf{y}^2\right)}{\left(1-\lambda^2\right)^4} \sum_{s=0}^{t-2}\left(\eta_\mathbf{x}^2 \mathbb{E}\left\|\bar{\mathbf{v}}_s\right\|^2\right. \\
& \left.\quad+\eta_\mathbf{y}^2 \mathbb{E}\left\|\bar{\mathbf{u}}_s\right\|^2\right)+\frac{64 m \lambda^4\left(\beta_\mathbf{x} \eta_\mathbf{x}^2+\beta_\mathbf{y} \eta_\mathbf{y}^2\right) \sigma^2}{\left(1-\lambda^2\right)^4 b_0}+\frac{192 m \lambda^4\left(\beta_\mathbf{x}^2 \eta_\mathbf{x}^2+\beta_\mathbf{y}^2 \eta_\mathbf{y}^2\right) \sigma^2 t}{\left(1-\lambda^2\right)^4}\\
&\quad+\frac{64 m \lambda^4 \eta_\mathbf{x}^2}{\left(1-\lambda^2\right)^3} \sum_{s=0}^{t-1}\mathbb{E}\left\|\mathcal{N}_{\mathbf{x},s}-\mathcal{N}_{\mathbf{x},s-1}\right\|^2  +\frac{64 m \lambda^4 \eta_\mathbf{x}^2}{\left(1-\lambda^2\right)^3} \mathbb{E}\left\|\mathcal{N}_{\mathbf{y},s}-\mathcal{N}_{\mathbf{y},s-1}\right\|^2
\end{aligned}
\end{equation}
\textit{Proof:} Combining Lemma \ref{lemma 8} and Lemma \ref{lemma 10} , for all $t \in\{0,1, \cdots, T\}$ we have:
\begin{equation}
\begin{aligned}
&\sum_{s=0}^t \left\|X_s - \bar{X}_s\right\|_F^2 \\
&\leq \frac{4 \lambda^2 \eta_\mathbf{x}^2}{\left(1 - \lambda^2\right)^2} \sum_{s=0}^{t-1} \left\|V_s - \bar{V}_s\right\|_F^2\\
&\leq \frac{8 \lambda^2 \eta_\mathbf{x}^2}{\left(1-\lambda^2\right)^3} \mathbb{E}\left\| V_0-\bar{V}_0 \right\|_F^2+\frac{192 \lambda^4 L^2 \eta_\mathbf{x}^2}{\left(1-\lambda^2\right)^4} \sum_{s=0}^{t-1}\left(\mathbb{E}\left\|X_s-\bar{X}_s\right\|_F^2+\mathbb{E}\left\|Y_s-\bar{Y}_s\right\|_F^2\right) \\
& \quad+\frac{96 m \lambda^4 L^2 \eta_\mathbf{x}^2}{\left(1-\lambda^2\right)^4} \sum_{s=0}^{t-2} \eta_\mathbf{y}^2 \mathbb{E}\left\|\bar{\mathbf{u}}_s\right\|^2+\frac{96 m \lambda^4 L^2 \eta_\mathbf{x}^2}{\left(1-\lambda^2\right)^4} \sum_{s=0}^{t-2} \eta_\mathbf{x}^2 \mathbb{E}\left\|\bar{\mathbf{v}}_s\right\|^2 +\frac{32 m \lambda^4 \beta_\mathbf{x}^2 \eta_\mathbf{x}^2 \sigma^2(t-1)}{\left(1-\lambda^2\right)^3}\\
& \quad+\frac{32 m \lambda^4 \eta_\mathbf{x}^2}{\left(1-\lambda^2\right)^3} \sum_{s=0}^{t-1}\mathbb{E}\left\|\mathcal{N}_{\mathbf{x},s}-\mathcal{N}_{\mathbf{x},s-1}\right\|^2+\frac{32 \lambda^4 \beta_\mathbf{x}^2 \eta_\mathbf{x}^2}{\left(1-\lambda^2\right)^4} \sum_{s=0}^{t-2} \sum_{i=1}^m \mathbb{E}\left\|\mathbf{g}_s^{(i)}-\nabla_\mathbf{x} f_i\left(\mathbf{x}_s^{(i)}, \mathbf{y}_s^{(i)}\right)\right\|^2
\end{aligned}
\end{equation}
Where we use Lemma \ref{lemma 11} in the last inequality. Using Lemma \ref{lemma 12} to replace the last term in the result.
\begin{equation}
\begin{aligned}
\label{eq13.1}
& \sum_{s=0}^t \mathbb{E}\left\|X_s-\bar{X}_s\right\|_F^2 \\
&\leq  \frac{8 \lambda^2 \eta_\mathbf{x}^2}{\left(1-\lambda^2\right)^3} \mathbb{E}\left\|V_0-\bar{V}_0\right\|_F^2+\frac{192 \lambda^4 L^2 \eta_\mathbf{x}^2}{\left(1-\lambda^2\right)^4} \sum_{s=0}^{t-1}\left(\mathbb{E}\left\|X_s-\bar{X}_s\right\|_F^2+\mathbb{E}\left\|Y_s-\bar{Y}_s\right\|_F^2\right) \\
& \quad+\frac{96 m \lambda^4 L^2 \eta_\mathbf{x}^2}{\left(1-\lambda^2\right)^4} \sum_{s=0}^{t-2}\left(\eta_\mathbf{x}^2 \mathbb{E}\left\|\bar{\mathbf{v}}_s\right\|^2+\eta_\mathbf{y}^2 \mathbb{E}\left\|\bar{\mathbf{u}}_s\right\|^2\right)+\frac{32 m \lambda^4 \beta_\mathbf{x} \eta_\mathbf{x}^2 \sigma^2}{\left(1-\lambda^2\right)^4 b_0}+\frac{64 m \lambda^4 \beta_\mathbf{x}^3 \eta_\mathbf{x}^2 \sigma^2(t-2)}{\left(1-\lambda^2\right)^4} \\
& \quad+\frac{32 m \lambda^4 \beta_\mathbf{x}^2 \eta_\mathbf{x}^2 \sigma^2(t-1)}{\left(1-\lambda^2\right)^3}+\frac{384 \lambda^4 \beta_\mathbf{x} L^2 \eta_\mathbf{x}^2}{\left(1-\lambda^2\right)^4} \sum_{s=0}^{t-2}\left(\mathbb{E}\left\|X_s-\bar{X}_s\right\|_F^2+\mathbb{E}\left\|Y_s-\bar{Y}_s\right\|_F^2\right) \\
& \quad+\frac{192 m \lambda^4 \beta_\mathbf{x} L^2 \eta_\mathbf{x}^2}{\left(1-\lambda^2\right)^4} \sum_{s=0}^{t-3}\left(\eta_\mathbf{x}^2 \mathbb{E}\left\|\bar{\mathbf{v}}_s\right\|^2+\eta_\mathbf{y}^2 \mathbb{E}\left\|\bar{\mathbf{u}}_s\right\|^2\right)+\frac{32 m \lambda^4 \eta_\mathbf{x}^2}{\left(1-\lambda^2\right)^3} \sum_{s=0}^{t-1}\mathbb{E}\left\|\mathcal{N}_{\mathbf{x},s}-\mathcal{N}_{\mathbf{x},s-1}\right\|^2 \\
&\leq  \frac{8 \lambda^2 \eta_\mathbf{x}^2}{\left(1-\lambda^2\right)^3} \mathbb{E}\left\|V_0-\bar{V}_0\right\|_F^2+\frac{576 \lambda^4 L^2 \eta_\mathbf{x}^2}{\left(1-\lambda^2\right)^4} \sum_{s=0}^{t-1}\left(\mathbb{E}\left\|X_s-\bar{X}_s\right\|_F^2+\mathbb{E}\left\|Y_s-\bar{Y}_s\right\|_F^2\right) \\
&\quad +\frac{288 m \lambda^4 L^2 \eta_\mathbf{x}^2}{\left(1-\lambda^2\right)^4} \sum_{s=0}^{t-2}\left(\eta_\mathbf{x}^2 \mathbb{E}\left\|\bar{\mathbf{v}}_s\right\|^2+\eta_\mathbf{y}^2 \mathbb{E}\left\|\bar{\mathbf{u}}_s\right\|^2\right)+\frac{32 m \lambda^4 \beta_\mathbf{x} \eta_\mathbf{x}^2 \sigma^2}{\left(1-\lambda^2\right)^4 b_0}+\frac{96 m \lambda^4 \beta_\mathbf{x}^2 \eta_\mathbf{x}^2 \sigma^2 t}{\left(1-\lambda^2\right)^4}\\
&\quad+\frac{32 m \lambda^4 \eta_\mathbf{x}^2}{\left(1-\lambda^2\right)^3} \sum_{s=0}^{t-1}\mathbb{E}\left\|\mathcal{N}_{\mathbf{x},s}-\mathcal{N}_{\mathbf{x},s-1}\right\|^2
\end{aligned}
\end{equation}
We use the condition $\beta_\mathbf{x} \leq 1$ in the inequality substitutions to simplify the expressions. Similarly, we can get:
\begin{equation}
\begin{aligned}
\label{eq13.2}
& \sum_{s=0}^t \mathbb{E}\left\|Y_s-\bar{Y}_s\right\|_F^2 \\
&\leq  \frac{8 \lambda^2 \eta_\mathbf{y}^2}{\left(1-\lambda^2\right)^3} \mathbb{E}\left\|U_0-\bar{U}_0\right\|_F^2+\frac{576 \lambda^4 L^2 \eta_\mathbf{y}^2}{\left(1-\lambda^2\right)^4} \sum_{s=0}^{t-1}\left(\mathbb{E}\left\|X_s-\bar{X}_s\right\|_F^2+\mathbb{E}\left\|Y_s-\bar{Y}_s\right\|_F^2\right) \\
&\quad +\frac{288 m \lambda^4 L^2 \eta_\mathbf{y}^2}{\left(1-\lambda^2\right)^4} \sum_{s=0}^{t-2}\left(\eta_\mathbf{x}^2 \mathbb{E}\left\|\bar{\mathbf{v}}_s\right\|^2+\eta_\mathbf{y}^2 \mathbb{E}\left\|\bar{\mathbf{u}}_s\right\|^2\right)+\frac{32 m \lambda^4 \beta_\mathbf{y} \eta_\mathbf{y}^2 \sigma^2}{\left(1-\lambda^2\right)^4 b_0}+\frac{96 m \lambda^4 \beta_\mathbf{y}^2 \eta_\mathbf{y}^2 \sigma^2 t}{\left(1-\lambda^2\right)^4}\\
&\quad+\frac{32 m \lambda^4 \eta_\mathbf{y}^2}{\left(1-\lambda^2\right)^3} \sum_{s=0}^{t-1}\mathbb{E}\left\|\mathcal{N}_{\mathbf{y},s}-\mathcal{N}_{\mathbf{y},s-1}\right\|^2
\end{aligned}
\end{equation}
Add Eq.(\ref{eq13.1}) and (\ref{eq13.2}), we obtain:
\begin{equation}
\begin{aligned}
& \sum_{s=0}^t\left(\mathbb{E}\left\|X_s-\bar{X}_s\right\|_F^2+\mathbb{E}\left\|Y_s-\bar{Y}_s\right\|_F^2\right)
\leq  \frac{8 \lambda^2 \eta_\mathbf{x}^2}{\left(1-\lambda^2\right)^3} \mathbb{E}\left\|V_0-\bar{V}_0\right\|_F^2+\frac{8 \lambda^2 \eta_\mathbf{y}^2}{\left(1-\lambda^2\right)^3} \mathbb{E}\left\|U_0-\bar{U}_0\right\|_F^2\\
&+\frac{576 \lambda^4 L^2\left(\eta_\mathbf{x}^2+\eta_\mathbf{y}^2\right)}{\left(1-\lambda^2\right)^4} \sum_{s=0}^{t-1}\left(\mathbb{E}\left\|X_s-\bar{X}_s\right\|_F^2+\mathbb{E}\left\|Y_s-\bar{Y}_s\right\|_F^2\right)+\frac{32 m \lambda^4\left(\beta_\mathbf{x} \eta_\mathbf{x}^2+\beta_\mathbf{y} \eta_\mathbf{y}^2\right) \sigma^2}{\left(1-\lambda^2\right)^4 b_0}\\
&+\frac{288 m \lambda^4 L^2\left(\eta_\mathbf{x}^2+\eta_\mathbf{y}^2\right)}{\left(1-\lambda^2\right)^4} \sum_{s=0}^{t-2}\left(\eta_\mathbf{x}^2 \mathbb{E}\left\|\bar{\mathbf{v}}_s\right\|^2+\eta_\mathbf{y}^2 \mathbb{E}\left\|\bar{\mathbf{u}}_s\right\|^2\right) +\frac{96 m \lambda^4\left(\beta_\mathbf{x}^2 \eta_\mathbf{x}^2+\beta_\mathbf{y}^2 \eta_\mathbf{y}^2\right) \sigma^2 t}{\left(1-\lambda^2\right)^4}\\
&+\frac{32 m \lambda^4 \eta_\mathbf{x}^2}{\left(1-\lambda^2\right)^3} \sum_{s=0}^{t-1}\mathbb{E}\left\|\mathcal{N}_{\mathbf{x},s}-\mathcal{N}_{\mathbf{x},s-1}\right\|^2+\frac{32 m \lambda^4 \eta_\mathbf{y}^2}{\left(1-\lambda^2\right)^3}\mathbb{E}\left\|\mathcal{N}_{\mathbf{y},s}-\mathcal{N}_{\mathbf{y},s-1}\right\|^2
\end{aligned}
\end{equation}

As $\lambda<1$, when $\eta_\mathbf{x} \leq \frac{(1-\lambda)^2}{500 L}$ and $\eta_\mathbf{y} \leq \frac{(1-\lambda)^2}{500 L}$, it holds that $\frac{576 \lambda^4 L^2\left(\eta_\mathbf{x}^2+\eta_\mathbf{y}^2\right)}{\left(1-\lambda^2\right)^4} \leq \frac{1}{2}$, thus, we can obtain:
\begin{equation}
\begin{aligned}
& \sum_{s=0}^t\left(\mathbb{E}\left\|X_s-\bar{X}_s\right\|_F^2+\mathbb{E}\left\|Y_s-\bar{Y}_s\right\|_F^2\right) \\
&\leq  \frac{16 \lambda^2 \eta_\mathbf{x}^2}{\left(1-\lambda^2\right)^3} \mathbb{E}\left\|V_0-\bar{V}_0\right\|_F^2+\frac{16 \lambda^2 \eta_\mathbf{y}^2}{\left(1-\lambda^2\right)^3} \mathbb{E}\left\|U_0-\bar{U}_0\right\|_F^2+\frac{576 n \lambda^4 L^2\left(\eta_\mathbf{x}^2+\eta_\mathbf{y}^2\right)}{\left(1-\lambda^2\right)^4} \sum_{s=0}^{t-2}\left(\eta_\mathbf{x}^2 \mathbb{E}\left\|\bar{\mathbf{v}}_s\right\|^2\right.\\
&\left.\quad+\eta_\mathbf{y}^2 \mathbb{E}\left\|\bar{\mathbf{u}}_s\right\|^2\right)+\frac{64 m \lambda^4\left(\beta_\mathbf{x} \eta_\mathbf{x}^2+\beta_\mathbf{y} \eta_\mathbf{y}^2\right) \sigma^2}{\left(1-\lambda^2\right)^4 b_0}+\frac{192 m \lambda^4\left(\beta_\mathbf{x}^2 \eta_\mathbf{x}^2+\beta_\mathbf{y}^2 \eta_\mathbf{y}^2\right) \sigma^2 t}{\left(1-\lambda^2\right)^4}\\
&\quad+\frac{64 m \lambda^4 \eta_\mathbf{x}^2}{\left(1-\lambda^2\right)^3} \sum_{s=0}^{t-1}\mathbb{E}\left\|\mathcal{N}_{\mathbf{x},s}-\mathcal{N}_{\mathbf{x},s-1}\right\|^2 +\frac{64 m \lambda^4 \eta_\mathbf{x}^2}{\left(1-\lambda^2\right)^3} \mathbb{E}\left\|\mathcal{N}_{\mathbf{y},s}-\mathcal{N}_{\mathbf{y},s-1}\right\|^2
\end{aligned}
\end{equation}

\subsection{Proof for main Theorems}
Here, we firstly prove the first equation in Theorem \ref{theorem 1}, we set 
\begin{equation}
\beta_\mathbf{x} = \frac{\epsilon \min \{1, m \epsilon\}}{20}, \quad T=\frac{1500 \kappa^3}{(1-\lambda)^2 \epsilon \beta_{\mathbf{x}}}
\end{equation}
Since $\Phi(x)$ is $(\kappa L+L)$-smooth we have:
\begin{equation}
\begin{aligned}
\Phi\left(\bar{\mathbf{x}}_t\right) \leq & \Phi\left(\bar{\mathbf{x}}_{t-1}\right)-\eta_\mathbf{x}\left\langle\bar{\mathbf{v}}_{t-1}, \nabla \Phi\left(\bar{\mathbf{x}}_{t-1}\right)\right\rangle+\eta_\mathbf{x}^2 \kappa L\left\|\bar{\mathbf{v}}_{t-1}\right\|^2 \\
= & \Phi\left(\bar{\mathbf{x}}_{t-1}\right)-\frac{\eta_\mathbf{x}}{2}\left\|\bar{\mathbf{v}}_{t-1}\right\|^2-\frac{\eta_\mathbf{x}}{2}\left\|\nabla \Phi\left(\bar{\mathbf{x}}_{t-1}\right)\right\|^2+\frac{\eta_\mathbf{x}}{2}\left\|\bar{\mathbf{v}}_{t-1}-\nabla \Phi\left(\bar{\mathbf{x}}_{t-1}\right)\right\|^2+\eta_\mathbf{x}^2 \kappa L\left\|\bar{\mathbf{v}}_{t-1}\right\|^2 \\
\end{aligned}
\end{equation}
Then we use Cauchy-Schwartz on above equation, we have:
\begin{equation}
\begin{aligned}
\label{eqt1.1}
\Phi\left(\bar{\mathbf{x}}_t\right)\leq & \Phi\left(\bar{\mathbf{x}}_{t-1}\right)-\frac{\eta_\mathbf{x}}{2}\left\|\nabla \Phi\left(\bar{\mathbf{x}}_{t-1}\right)\right\|^2-\left(\frac{\eta_\mathbf{x}}{2}-\eta_\mathbf{x}^2 \kappa L\right)\left\|\bar{\mathbf{v}}_{t-1}\right\|^2+\eta_\mathbf{x}\left\|\bar{\mathbf{v}}_{t-1}-\nabla_\mathbf{x} f\left(\bar{\mathbf{x}}_{t-1}, \bar{\mathbf{y}}_{t-1}\right)\right\|^2 \\
& +\eta_\mathbf{x}\left\|\nabla \Phi\left(\bar{\mathbf{x}}_{t-1}\right)-\nabla_\mathbf{x} f\left(\bar{\mathbf{x}}_{t-1}, \bar{\mathbf{y}}_{t-1}\right)\right\|^2
\end{aligned}
\end{equation}
Because $\nabla \Phi\left(\bar{\mathbf{x}}_{t-1}\right)=$ $\nabla_\mathbf{x} f\left(\bar{\mathbf{x}}_{t-1}, \hat{\mathbf{y}}_{t-1}\right)$, according to Assumption \ref{assumption 1}, the last term satisfies:
\begin{equation}
\label{eqt1.2}
\left\|\nabla \Phi\left(\bar{\mathbf{x}}_{t-1}\right)-\nabla_\mathbf{x} f\left(\bar{\mathbf{x}}_{t-1}, \bar{\mathbf{y}}_{t-1}\right)\right\|^2 \leq L^2\left\|\hat{\mathbf{y}}_{t-1}-\bar{\mathbf{y}}_{t-1}\right\|^2=L^2 \delta_{t-1}
\end{equation}
Additionally, using Cauchy-Schwartz inequality and Assumption \ref{assumption 1} we have:
\begin{equation}
\begin{aligned}
\label{eqt1.3}
& \left\|\bar{\mathbf{v}}_{t-1}-\nabla_\mathbf{x} f\left(\bar{\mathbf{x}}_{t-1}, \bar{\mathbf{y}}_{t-1}\right)\right\|^2 \\
&\leq  2\left\|\bar{\mathbf{v}}_{t-1}-\frac{1}{m} \sum_{i=1}^m \nabla_\mathbf{x} f_i\left(\mathbf{x}_{t-1}^{(i)}, \mathbf{y}_{t-1}^{(i)}\right)\right\|^2+2\left\|\frac{1}{m} \sum_{i=1}^m \nabla_\mathbf{x} f_i\left(\mathbf{x}_{t-1}^{(i)}, \mathbf{y}_{t-1}^{(i)}\right)-\nabla_\mathbf{x} f\left(\bar{\mathbf{x}}_{t-1}, \bar{\mathbf{y}}_{t-1}\right)\right\|^2 \\
&\leq  2\left\|\bar{\mathbf{v}}_{t-1}-\frac{1}{m} \sum_{i=1}^m \nabla_\mathbf{x} f_i\left(\mathbf{x}_{t-1}^{(i)}, \mathbf{y}_{t-1}^{(i)}\right)\right\|^2+2\left\| \sum_{i=1}^m \frac{1}{m}\left(\nabla_\mathbf{x} f_i\left(\mathbf{x}_{t-1}^{(i)}, \mathbf{y}_{t-1}^{(i)}\right)-\nabla_\mathbf{x} f\left(\bar{\mathbf{x}}_{t-1}, \bar{\mathbf{y}}_{t-1}\right)\right)\right\|^2\\
&\leq  2\left\|\bar{\mathbf{v}}_{t-1}-\frac{1}{m} \sum_{i=1}^m \nabla_\mathbf{x} f_i\left(\mathbf{x}_{t-1}^{(i)}, \mathbf{y}_{t-1}^{(i)}\right)\right\|^2+\frac{2 L^2}{m}\left(\left\|X_{t-1}-\bar{X}_{t-1}\right\|_F^2+\left\|Y_{t-1}-\bar{Y}_{t-1}\right\|_F^2\right)
\end{aligned}
\end{equation}
Combine Eq.(\ref{eqt1.1}) (\ref{eqt1.2}) and (\ref{eqt1.3}) and we can get the inequality:
\begin{equation}
\begin{aligned}
\label{eqt1.4}
&\left\|\nabla \Phi\left(\bar{\mathbf{x}}_{t-1}\right)\right\|^2 \\
&\leq  \frac{2\left(\Phi\left(\bar{\mathbf{x}}_{t-1}\right)-\Phi\left(\bar{\mathbf{x}}_t\right)\right)}{\eta_\mathbf{x}}-\left(1-2 \kappa L \eta_\mathbf{x}\right)\left\|\bar{\mathbf{v}}_{t-1}\right\|^2+2 L^2 \delta_{t-1}+\frac{4 L^2}{m}\left(\left\|X_{t-1}-\bar{X}_{t-1}\right\|_F^2\right. \\
& \left.\quad+\left\|Y_{t-1}-\bar{Y}_{t-1}\right\|_F^2\right)+4\left\|\bar{\mathbf{v}}_{t-1}-\frac{1}{m} \sum_{i=1}^m \nabla_\mathbf{x} f_i\left(\mathbf{x}_{t-1}^{(i)}, \mathbf{y}_{t-1}^{(i)}\right)\right\|^2
\end{aligned}
\end{equation}
Telescoping and taking expectation on Eq.(\ref{eqt1.4}) we have:
\begin{equation}
\begin{aligned}
&\frac{1}{T} \sum_{t=0}^{T-1} \mathbb{E}\left\|\nabla \Phi\left(\bar{\mathbf{x}}_t\right)\right\|^2 \\
&\leq  \frac{2\left(\Phi\left(\mathbf{x}_0\right)-\mathbb{E} \Phi\left(\bar{\mathbf{x}}_t\right)\right)}{\eta_\mathbf{x} T}-\frac{\left(1-2 \kappa L \eta_\mathbf{x}\right)}{T} \sum_{t=0}^{T-1} \mathbb{E}\left\|\bar{\mathbf{v}}_t\right\|^2+\frac{2 L^2}{T} \sum_{t=0}^{T-1} \mathbb{E}\delta_t \\
& \quad+\frac{4 L^2}{m T} \sum_{t=0}^{T-1}\left(\mathbb{E}\left\|X_t-\bar{X}_t\right\|_F^2+\mathbb{E}\left\|Y_t-\bar{Y}_t\right\|_F^2\right)+\frac{4}{T} \sum_{t=0}^{T-1} \mathbb{E}\left\|\bar{\mathbf{v}}_t-\frac{1}{m} \sum_{i=1}^m \nabla_\mathbf{x} f_i\left(\mathbf{x}_t^{(i)}, \mathbf{y}_t^{(i)}\right)\right\|^2
\end{aligned}
\end{equation}
Using Lemma \ref{lemma 6} to replace \begin{equation}
\begin{aligned}
&\frac{1}{T} \sum_{t=0}^{T-1} \mathbb{E}\left\|\nabla \Phi\left(\bar{\mathbf{x}}_t\right)\right\|^2 \\
&\leq  \frac{2\left(\Phi\left(\mathbf{x}_0\right) - \Phi^*\right)}{\eta_\mathbf{x} T} - \left(1 - 2 \kappa L \eta_\mathbf{x} - \frac{40 \kappa^4 \eta_\mathbf{x}^2}{\eta_\mathbf{y}^2}\right) \frac{1}{T} \sum_{t=0}^{T-1} \mathbb{E}\left\|\bar{\mathbf{v}}_t\right\|^2 + \frac{8 \kappa L^2 \delta_0}{T L \eta_\mathbf{y}}\\
&\quad - \frac{28 \kappa L \eta_\mathbf{y}}{5 T} \sum_{t=0}^{T-1} \left(1 - \left(1 - \frac{\mu \eta_\mathbf{y}}{4}\right)^{T-t}\right) \mathbb{E}\left\|\bar{\mathbf{u}}_t\right\|^2  + \frac{84 \kappa^2 L^2}{m T} \sum_{t=0}^{T-1} \left(\mathbb{E}\left\|X_t - \bar{X}_t\right\|_F^2 \right.\\
&\left.\quad+ \mathbb{E}\left\|Y_t - \bar{Y}_t\right\|_F^2\right)+ \frac{4}{T} \sum_{t=0}^{T-1} \mathbb{E}\left\|\bar{\mathbf{v}}_t - \frac{1}{m} \sum_{i=1}^m \nabla_\mathbf{x} f_i\left(\mathbf{x}_t^{(i)}, \mathbf{y}_t^{(i)}\right)\right\|^2 \\
&\quad + \frac{20 \kappa L \eta_\mathbf{y}}{T} \sum_{t=1}^{T-1} \left(1 - \frac{\mu \eta_\mathbf{y}}{4}\right)^{T-t-1} \sum_{s=0}^{t-1} \mathbb{E}\left\|\bar{\mathbf{u}}_t - \frac{1}{m} \sum_{i=1}^m \nabla f_i\left(\mathbf{x}_s^{(i)}, \mathbf{y}_s^{(i)}\right)\right\|^2  
\end{aligned}
\end{equation}
And using Lemma \ref{lemma 9} to replace the last two terms.
\begin{equation}
\begin{aligned}
&\frac{1}{T} \sum_{t=0}^{T-1} \mathbb{E}\left\|\nabla \Phi\left(\bar{\mathbf{x}}_t\right)\right\|^2\\
&\leq  \frac{2\left(\Phi\left(\mathbf{x}_0\right)-\Phi^*\right)}{\eta_\mathbf{x} T}-\left(1-2 \kappa L \eta_\mathbf{x}-\frac{40 \kappa^4 \eta_\mathbf{x}^2}{\eta_\mathbf{y}^2}\right) \frac{1}{T} \sum_{t=0}^{T-1} \mathbb{E}\left\|\bar{\mathbf{v}}_t\right\|^2+\frac{8 \kappa L^2 \delta_0}{T L \eta_\mathbf{y}}+\frac{8 \sigma^2}{m b_0 T}\left(\frac{1}{\beta_\mathbf{x}}+\frac{20 \kappa^2}{\beta_\mathbf{y}}\right) \\
& \quad+\frac{8 \sigma^2}{m}\left(\beta_\mathbf{x}+20 \kappa^2 \beta_\mathbf{y}\right)+\frac{4 L^2}{m T}\left(21 \kappa^2+\frac{12}{m \beta_\mathbf{x}}+\frac{240 \kappa^2}{m \beta_\mathbf{y}}\right) \sum_{t=0}^{T-1}\left(\mathbb{E}\left\|X_t-\bar{X}_t\right\|_F^2+\mathbb{E}\left\|Y_t-\bar{Y}_t\right\|_F^2\right) \\
& \quad+\frac{24 L^2}{m \beta_\mathbf{x} T} \sum_{t=0}^{T-1}\left(1-\left(1-\beta_\mathbf{x}\right)^{T-t}\right)\left(\eta_\mathbf{x}^2 \mathbb{E}\left\|\bar{\mathbf{v}}_t\right\|^2+\eta_\mathbf{y}^2 \mathbb{E}\left\|\bar{\mathbf{u}}_t\right\|^2\right)+\frac{480 \kappa^2 L^2}{m \beta_\mathbf{y} T} \sum_{t=0}^{T-1}\\
& \quad\left(1-\left(1-\frac{\mu \eta_\mathbf{y}}{4}\right)^{T-t}\right) \left(\eta_\mathbf{x}^2 \mathbb{E}\left\|\bar{\mathbf{v}}_t\right\|^2+\eta_\mathbf{y}^2 \mathbb{E}\left\|\bar{\mathbf{u}}_t\right\|^2\right)-\frac{28\kappa L \eta_\mathbf{y}}{5 T} \sum_{t=0}^{T-1}\left(1-\left(1-\frac{\mu \eta_\mathbf{y}}{4}\right)^{T-t}\right) \mathbb{E}\left\|\bar{\mathbf{u}}_t\right\|^2\\
&\quad+\frac{8}{T} \sum_{s=0}^{T-1} \mathbb{E}\left\|\mathcal{N}_{\mathbf{x}, s}- \left(1 - \beta_\mathbf{x}\right)\mathcal{N}_{\mathbf{x}, s-1}\right\|^2+\frac{40 \kappa L \eta_{\mathbf{y}}}{T} \sum_{t=1}^{T-1}\left(1-\frac{\mu \eta_{\mathbf{y}}}{4}\right)^{T-t-1}\sum_{s=0}^{t-1} \mathbb{E}\left\|\mathcal{N}_{\mathbf{y}, s}- \left(1 - \beta_\mathbf{y}\right)\mathcal{N}_{\mathbf{y}, s-1}\right\|^2
\end{aligned}
\end{equation}
For the sum of $1-\beta_\mathbf{x}$
\begin{equation}
\label{eqt1.5}
\frac{1}{\beta_\mathbf{x}}\left(1-\left(1-\beta_\mathbf{x}\right)^{T-t}\right)=\sum_{s=0}^{T-t-1}\left(1-\beta_\mathbf{x}\right)^s
\end{equation}
we know Eq.(\ref{eqt1.5}) is increasing when $\beta_\mathbf{x}$ is decreasing. \\Hence $\frac{1}{\beta_\mathbf{x}}\left(1-\left(1-\beta_\mathbf{x}\right)^{T-t}\right) \leq \frac{300 \kappa^2}{(1-\lambda)^2 \beta_\mathbf{x}} \left (1-(1-\frac{(1-\lambda)^2 \beta_\mathbf{x}}{300 \kappa^2})^{T-t}\right)$. According to the definition of $\beta_\mathbf{x}$ and $\eta_\mathbf{y}$, we have $\frac{(1-\lambda)^2 \beta_\mathbf{x}}{300 \kappa^2} \leq \frac{\mu \eta_\mathbf{y}}{4}$ and
$
\frac{24 L^2}{m \beta_\mathbf{x} T}\left(1-\left(1-\beta_\mathbf{x}\right)^{T-t}\right) \leq \frac{7200 L^2 \kappa^2}{m(1-\lambda)^2 \beta_\mathbf{x} T}\left(1-\left(1-\frac{\mu \eta_\mathbf{y}}{4}\right)^{T-t}\right)
$. 
Therefore, using the definition of $\beta_\mathbf{x}, \beta_\mathbf{y}$ and $\eta_\mathbf{y}$ we obtain:

\begin{equation}
\begin{aligned}
& \frac{1}{T} \sum_{t=0}^{T-1} \mathbb{E}\left\|\nabla \Phi\left(\bar{\mathbf{x}}_t\right)\right\|^2 \\
&\leq  \frac{2\left(\Phi\left(\mathbf{x}_0\right)-\Phi^*\right)}{\eta_\mathbf{x} T}-\left(1-2 \kappa L \eta_\mathbf{x}-\frac{40 \kappa^4 \eta_\mathbf{x}^2}{\eta_\mathbf{y}^2}\right) \frac{1}{T} \sum_{t=0}^{T-1} \mathbb{E}\left\|\bar{\mathbf{v}}_t\right\|^2+\frac{8 \kappa L^2 \delta_0}{T L \eta_\mathbf{y}}+\frac{8 \sigma^2}{m b_0 T}\left(\frac{1}{\beta_\mathbf{x}}+\frac{20 \kappa^2}{\beta_\mathbf{y}}\right) \\
& \quad+\frac{8 \sigma^2}{m}\left(\beta_\mathbf{x}+20 \kappa^2 \beta_\mathbf{y}\right)+\frac{4 L^2}{m T}\left(21 \kappa^2+\frac{12}{m \beta_\mathbf{x}}+\frac{240 \kappa^2}{m \beta_\mathbf{y}}\right) \sum_{t=0}^{T-1}\left(\mathbb{E}\left\|X_t-\bar{X}_t\right\|_F^2+\mathbb{E}\left\|Y_t-\bar{Y}_t\right\|_F^2\right) \\
& \quad+\left(\frac{24 L^2 \eta_\mathbf{x}^2}{m \beta_\mathbf{x}}+\frac{480 \kappa^2 L^2 \eta_\mathbf{x}^2}{m \beta_\mathbf{y}}\right) \frac{1}{T} \sum_{t=0}^{T-1} \mathbb{E}\left\|\bar{\mathbf{v}}_t\right\|^2-\frac{\kappa L \eta_\mathbf{y}}{T} \sum_{t=0}^{T-1} \mathbb{E}\left\|\bar{\mathbf{u}}_t\right\|^2+\frac{8}{T} \sum_{s=0}^{T-1} \mathbb{E}\left\|\mathcal{N}_{\mathbf{x}, s}- \left(1 - \beta_\mathbf{x}\right)\mathcal{N}_{\mathbf{x}, s-1}\right\|^2\\
&\quad+\frac{160\kappa^2}{T} \sum_{s=0}^{T-1} \mathbb{E}\left\|\mathcal{N}_{\mathbf{y}, s}- \left(1 - \beta_\mathbf{y}\right)\mathcal{N}_{\mathbf{y}, s-1}\right\|^2
\end{aligned}
\end{equation}
Using Lemma \ref{lemma 13} to replace $ \mathbb{E}\left\|X_s - \bar{X}_s\right\|_F^2 + \mathbb{E}\left\|Y_s - \bar{Y}_s\right\|_F^2$
\begin{equation}
\begin{aligned}
& \frac{1}{T} \sum_{t=0}^{T-1} \mathbb{E}\left\|\nabla \Phi\left(\bar{\mathbf{x}}_t\right)\right\|^2 \\
&\leq \frac{2\left(\Phi\left(\mathbf{x}_0\right)-\Phi^*\right)}{\eta_\mathbf{x} T}-\left(1-2 \kappa L \eta_\mathbf{x}-\frac{40 \kappa^4 \eta_\mathbf{x}^2}{\eta_\mathbf{y}^2}\right) \frac{1}{T} \sum_{t=0}^{T-1} \mathbb{E}\left\|\bar{\mathbf{v}}_t\right\|^2+\frac{8 \kappa L^2 \delta_0}{T L \eta_\mathbf{y}}+\frac{8 \sigma^2}{m b_0 T}\left(\frac{1}{\beta_\mathbf{x}}+\frac{20 \kappa^2}{\beta_\mathbf{y}}\right) \\
&\quad +\frac{8 \sigma^2}{m}\left(\beta_\mathbf{x}+20 \kappa^2 \beta_\mathbf{y}\right)+\frac{4 L^2}{m T}\left(21 \kappa^2+\frac{12}{m \beta_\mathbf{x}}+\frac{240 \kappa^2}{m \beta_\mathbf{y}}\right)\left(\frac{16 \lambda^2 \eta_\mathbf{x}^2}{\left(1-\lambda^2\right)^3} \mathbb{E}\left\|V_0-\bar{V}_0\right\|_F^2\right. \\
&\quad \left.+\frac{16 \lambda^2 \eta_\mathbf{y}^2}{\left(1-\lambda^2\right)^3} \mathbb{E}\left\|U_0-\bar{U}_0\right\|_F^2+\frac{64 m \lambda^4\left(\beta_\mathbf{x} \eta_\mathbf{x}^2+\beta_\mathbf{y} \eta_\mathbf{y}^2\right) \sigma^2}{\left(1-\lambda^2\right)^4 b_0}+\frac{192 m \lambda^4\left(\beta_\mathbf{x}^2 \eta_\mathbf{x}^2+\beta_\mathbf{y}^2 \eta_\mathbf{y}^2\right) \sigma^2 T}{\left(1-\lambda^2\right)^4}\right) \\
&\quad+\frac{4 L^2}{m T}\left(21 \kappa^2+\frac{12}{m \beta_\mathbf{x}}+\frac{240 \kappa^2}{m \beta_\mathbf{y}}\right) \frac{576 m \lambda^4 L^2\left(\eta_\mathbf{x}^2+\eta_\mathbf{y}^2\right)}{\left(1-\lambda^2\right)^4} \sum_{t=0}^{T-1}\left(\eta_\mathbf{x}^2 \mathbb{E}\left\|\bar{\mathbf{v}}_t\right\|^2+\eta_\mathbf{y}^2 \mathbb{E}\left\|\bar{\mathbf{u}}_t\right\|^2\right) \\
&\quad +\left(\frac{24 L^2 \eta_\mathbf{x}^2}{m \beta_\mathbf{x}}+\frac{480 \kappa^2 L^2 \eta_\mathbf{x}^2}{m \beta_\mathbf{y}}\right) \frac{1}{T} \sum_{t=0}^{T-1} \mathbb{E}\left\|\bar{\mathbf{v}}_t\right\|^2-\frac{\kappa L \eta_\mathbf{y}}{T} \sum_{t=0}^{T-1} \mathbb{E}\left\|\bar{\mathbf{u}}_t\right\|^2+\frac{8}{T} \sum_{s=0}^{T-1} \mathbb{E}\left\|\mathcal{N}_{\mathbf{x}, s}- \left(1 - \beta_\mathbf{x}\right)\mathcal{N}_{\mathbf{x}, s-1}\right\|^2\\
&\quad+\frac{160\kappa^2}{T} \sum_{s=0}^{T-1} \mathbb{E}\left\|\mathcal{N}_{\mathbf{y}, s}- \left(1 - \beta_\mathbf{y}\right)\mathcal{N}_{\mathbf{y}, s-1}\right\|^2+\frac{256 L^2 \lambda^4 \eta_\mathbf{x}^2}{ T \left(1-\lambda^2\right)^3}\left(21 \kappa^2+\frac{12}{m \beta_\mathbf{x}}+\frac{240 \kappa^2}{m \beta_\mathbf{y}}\right)\\
&\quad \sum_{s=0}^{T-1}\mathbb{E}\left\|\mathcal{N}_{\mathbf{x}, s}-\mathcal{N}_{\mathbf{x}, s-1}\right\|^2 +\frac{256 L^2 \lambda^4 \eta_\mathbf{y}^2}{ T \left(1-\lambda^2\right)^3}\left(21 \kappa^2+\frac{12}{m \beta_\mathbf{x}}+\frac{240 \kappa^2}{m \beta_\mathbf{y}}\right)\sum_{s=0}^{T-1}\mathbb{E}\left\|\mathcal{N}_{\mathbf{y}, s}-\mathcal{N}_{\mathbf{y}, s-1}\right\|^2\\
\end{aligned}
\end{equation}
When $\beta_\mathbf{x}, \beta_\mathbf{y}, \eta_\mathbf{x}$ and $\eta_\mathbf{y}$ are defined as Theorem \ref{theorem 1} , we have
\begin{equation}
\frac{4 L^2}{m T}\left(21 \kappa^2+\frac{12}{m \beta_\mathbf{x}}+\frac{240 \kappa^2}{m \beta_\mathbf{y}}\right) \frac{576 m \lambda^4 L^2\left(\eta_\mathbf{x}^2+\eta_\mathbf{y}^2\right)}{\left(1-\lambda^2\right)^4} \eta_\mathbf{y}^2 \leq \frac{\kappa L \eta_\mathbf{y}}{2T}
\end{equation}
and 
\begin{equation}
\begin{aligned}
&1-2 \kappa L \eta_\mathbf{x} -\frac{40 \kappa^4 \eta_\mathbf{x}^2}{\eta_\mathbf{y}^2}-\frac{24 L^2 \eta_\mathbf{x}^2}{m \beta_\mathbf{x}}-\frac{480 \kappa^2 L^2 \eta_\mathbf{x}^2}{m \beta_\mathbf{y}}\\
&-\frac{4 L^2}{m}\left(21 \kappa^2+\frac{12}{m \beta_\mathbf{x}}+\frac{240 \kappa^2}{m \beta_\mathbf{y}}\right) \frac{576 m \lambda^4 L^2\left(\eta_\mathbf{x}^2+\eta_\mathbf{y}^2\right)}{\left(1-\lambda^2\right)^4} \eta_\mathbf{x}^2
\geq \frac{2}{5} 
\end{aligned}
\end{equation}
Therefore, subtracting the terms containing these two quantities will not affect the validity of the inequality. This simplification is achieved by using this scaling method.
\begin{equation}
\begin{aligned}
& \frac{1}{T} \sum_{t=0}^{T-1} \mathbb{E}\left\|\nabla \Phi\left(\bar{\mathbf{x}}_t\right)\right\|^2 \\
&\leq \frac{2\left(\Phi\left(\mathbf{x}_0\right)-\Phi^*\right)}{\eta_\mathbf{x} T}+\frac{8 \kappa L^2 \delta_0}{T L \eta_\mathbf{y}}+\frac{8 \sigma^2}{m b_0 T}\left(\frac{1}{\beta_\mathbf{x}}+\frac{20 \kappa^2}{\beta_\mathbf{y}}\right)+\frac{8 \sigma^2}{m}\left(\beta_\mathbf{x}+20 \kappa^2 \beta_\mathbf{y}\right) \\
& \quad+\frac{4 L^2}{m T}\left(21 \kappa^2+\frac{12}{m \beta_\mathbf{x}}+\frac{240 \kappa^2}{m \beta_\mathbf{y}}\right)\left(\frac{16 \lambda^2 \eta_\mathbf{x}^2}{\left(1-\lambda^2\right)^3} \mathbb{E}\left\|V_0-\bar{V}_0\right\|_F^2+\frac{16 \lambda^2 \eta_\mathbf{y}^2}{\left(1-\lambda^2\right)^3} \mathbb{E}\left\|U_0-\bar{U}_0\right\|_F^2\right. \\
&\quad \left.+\frac{64 m \lambda^4\left(\beta_\mathbf{x} \eta_\mathbf{x}^2+\beta_\mathbf{y} \eta_\mathbf{y}^2\right) \sigma^2}{\left(1-\lambda^2\right)^4 b_0}+\frac{192 m \lambda^4\left(\beta_\mathbf{x}^2 \eta_\mathbf{x}^2+\beta_\mathbf{y}^2 \eta_\mathbf{y}^2\right) \sigma^2 T}{\left(1-\lambda^2\right)^4}\right)+\frac{8}{T} \sum_{s=0}^{T-1} \mathbb{E}\left\|\mathcal{N}_{\mathbf{x}, s}- \left(1 - \beta_\mathbf{x}\right)\mathcal{N}_{\mathbf{x}, s-1}\right\|^2\\
&\quad+\frac{160\kappa^2}{T} \sum_{s=0}^{T-1} \mathbb{E}\left\|\mathcal{N}_{\mathbf{y}, s}- \left(1 - \beta_\mathbf{y}\right)\mathcal{N}_{\mathbf{y}, s-1}\right\|^2+\frac{256 L^2 \lambda^4 \eta_\mathbf{x}^2}{ T \left(1-\lambda^2\right)^3}\left(21 \kappa^2+\frac{12}{m \beta_\mathbf{x}}+\frac{240 \kappa^2}{m \beta_\mathbf{y}}\right)\sum_{s=0}^{T-1} \\
&\quad\quad\mathbb{E}\left\|\mathcal{N}_{\mathbf{x}, s}-\mathcal{N}_{\mathbf{x}, s-1}\right\|^2+\frac{256 L^2 \lambda^4 \eta_\mathbf{y}^2}{ T \left(1-\lambda^2\right)^3}\left(21 \kappa^2+\frac{12}{m \beta_\mathbf{x}}+\frac{240 \kappa^2}{m \beta_\mathbf{y}}\right)\sum_{s=0}^{T-1}\mathbb{E}\left\|\mathcal{N}_{\mathbf{y}, s}-\mathcal{N}_{\mathbf{y}, s-1}\right\|^2
\end{aligned}
\end{equation}
By Assumption \ref{assumption 4} and Cauchy-Schwartz inequality we also have
\begin{equation}
\mathbb{E}\left\|V_0-\bar{V}_0\right\|_F^2=\mathbb{E}\left\|G_0(W-J)\right\|_F^2 \leq \lambda^2 \mathbb{E}\left\|G_0\right\|_F^2 \leq \frac{2 m \lambda^2 \sigma^2}{b_0}+2 \lambda^2 \sum_{i=1}^m\left\|\nabla_\mathbf{x} f_i\left(\mathbf{x}_0, \mathbf{y}_0\right)\right\|^2
\end{equation}
Similarly, we have
\begin{equation}
\mathbb{E}\left\|U_0-\bar{U}_0\right\|_F^2 \leq \frac{2 m \lambda^2 \sigma^2}{b_0}+2 \lambda^2 \sum_{i=1}^m\left\|\nabla_\mathbf{y} f_i\left(\mathbf{x}_0, \mathbf{y}_0\right)\right\|^2
\end{equation}
With the definition of $\beta_\mathbf{x}, \beta_\mathbf{y}, \eta_\mathbf{x}$ and $\eta_\mathbf{y}$ are given in Theorem \ref{theorem 1}, therefore we get:
\begin{equation}
\begin{aligned}
\frac{256 L^2 \lambda^4 \eta_\mathbf{x}^2}{T\left(1-\lambda^2\right)^3}\left(21 \kappa^2+\frac{12}{n \beta_\mathbf{x}}+\frac{240 \kappa^2}{n \beta_\mathbf{y}}\right) \leq   \frac{\lambda^4(1-\lambda)}{T}\left(m^2\epsilon^2+26\kappa^2\right)\\
\frac{256 L^2 \lambda^4 \eta_\mathbf{y}^2}{T\left(1-\lambda^2\right)^3}\left(21 \kappa^2+\frac{12}{n \beta_\mathbf{x}}+\frac{240 \kappa^2}{n \beta_\mathbf{y}}\right) \leq  \frac{\lambda^4(1-\lambda)}{T}\left(m^2\epsilon^2+26\kappa^2\right)
\end{aligned}
\end{equation}
We know that the maximum of $\lambda^4(1-\lambda)$ is $\frac{256}{3075} < 1$, meanwhile, by the definition of $\kappa=\frac{L}{\mu}\leq 1$ thus, we have:
\begin{equation}
\begin{aligned}
\frac{256 L^2 \lambda^4 \eta_\mathbf{x}^2}{T\left(1-\lambda^2\right)^3}\left(21 \kappa^2+\frac{12}{n \beta_\mathbf{x}}+\frac{240 \kappa^2}{n \beta_\mathbf{y}}\right) \leq   \frac{\left(m^2\epsilon^2+3\kappa^2\right)}{T}\\
\frac{256 L^2 \lambda^4 \eta_\mathbf{y}^2}{T\left(1-\lambda^2\right)^3\kappa^2}\left(21 \kappa^2+\frac{12}{n \beta_\mathbf{x}}+\frac{240 \kappa^2}{n \beta_\mathbf{y}}\right) \leq  \frac{\left(m^2\epsilon^2+3\right)}{T}
\end{aligned}
\end{equation}
Combine above three inequalities and substitute the parameters with their definitions. We achieve
\begin{equation}
\begin{aligned}
\label{eqt1.6}
&\frac{1}{T} \sum_{t=0}^{T-1} \mathbb{E}\left\|\nabla \Phi\left(\bar{\mathbf{x}}_t\right)\right\|^2 \\
\leq &  L\left(\Phi\left(\mathbf{x}_0\right)-\Phi^*\right) \epsilon^2+2 L^2 \delta_0 \epsilon^2+2 \sigma^2 \epsilon^2+ \frac{\epsilon^2}{m} \sum_{i=1}^m\left\|\nabla_\mathbf{x} f_i\left(\mathbf{x}_0, \mathbf{y}_0\right)\right\|^2 \\
& + \frac{\epsilon^2}{m} \sum_{i=1}^m\left\|\nabla_\mathbf{x} f_i\left(\mathbf{x}_0, \mathbf{y}_0\right)\right\|^2+ \frac{\left(m^2\epsilon^2+3\kappa^2\right)}{T} \sum_{s=0}^{T-1} \left(\mathbb{E}\left\|\mathcal{N}_{\mathbf{x}, s}-\mathcal{N}_{\mathbf{x}, s-1}\right\|^2+\mathbb{E}\left\|\mathcal{N}_{\mathbf{y}, s}-\mathcal{N}_{\mathbf{y}, s-1}\right\|^2\right)\\
&+\frac{8}{T} \sum_{s=0}^{T-1} \mathbb{E}\left\|\mathcal{N}_{\mathbf{x}, s}- \left(1 - \beta_\mathbf{x}\right)\mathcal{N}_{\mathbf{x}, s-1}\right\|^2 +\frac{160\kappa^2}{T} \sum_{s=0}^{T-1} \mathbb{E}\left\|\mathcal{N}_{\mathbf{y}, s}- \left(1 - \beta_\mathbf{y}\right)\mathcal{N}_{\mathbf{y}, s-1}\right\|^2
\end{aligned}
\end{equation}
Now, review the definition of $\mathcal{N}_{\mathbf{x}, s}$ and $\mathcal{N}_{\mathbf{x}, s}$, we can obtain that:
\begin{equation}
\mathcal{N}_{\mathbf{x}, t} \sim \mathcal{N}\left(0, \frac{\sigma_\mathbf{x}^2}{m} I_{d_1}\right), \mathcal{N}_{y, t} \sim \mathcal{N}\left(0, \frac{\sigma_\mathbf{y}^2}{m} I_{d_2}\right)
\end{equation}
\(\mathcal{N}_{\mathbf{x}, s}\) and \(\mathcal{N}_{\mathbf{x}, s-1}\) are independent normally distributed random variables because the noises \(n_{\mathbf{x}, s}^{(i)}\) and \(n_{\mathbf{x}, s-1}^{(i)}\) generated at times \(s\) and \(s-1\) are independent. Therefore, the distribution of \(\mathcal{N}_{\mathbf{x}, s} - \mathcal{N}_{\mathbf{x}, s-1}\) is also normally distributed, with mean 0 and a covariance matrix that is the sum of the covariances of the two independent normal distributions.
\begin{equation}
\mathcal{N}_{\mathbf{x}, s}-\mathcal{N}_{\mathbf{x}, s-1} \sim \mathcal{N}\left(0,2  \frac{\sigma_\mathbf{x}^2}{m} I_{d_1}\right)
\end{equation}
Therefore:
\begin{equation}
\mathbb{E}\left\|\mathcal{N}_{\mathbf{x}, s}-\mathcal{N}_{\mathbf{x}, s-1}\right\|^2=\operatorname{Tr}\left(2 \cdot \frac{\sigma_\mathbf{x}^2}{m} I_{d_1}\right)=2  \frac{\sigma_\mathbf{x}^2}{m}  d_1
\end{equation}
Sum up from 0 to $T-1$:
\begin{equation}
\sum_{s=0}^{T-1} \mathbb{E}\left\|\mathcal{N}_{\mathbf{x}, s}-\mathcal{N}_{\mathbf{x}, s-1}\right\|^2=\sum_{s=0}^{T-1} 2 \cdot \frac{\sigma_\mathbf{x}^2}{m} \cdot d_1=2  \frac{\sigma_\mathbf{x}^2}{m}  d_1  T
\end{equation}
Similarly, we can get:
\begin{equation}
\sum_{s=0}^{T-1} \mathbb{E}\left\|\mathcal{N}_{\mathbf{y}, s}-\mathcal{N}_{\mathbf{y}, s-1}\right\|^2=2  \frac{\sigma_\mathbf{y}^2}{m}  d_2  T
\end{equation}
Mimic the process above, we know that:
\begin{equation}
\mathcal{N}_{\mathbf{x}, s}-\left(1-\beta_{\mathbf{x}}\right) \mathcal{N}_{\mathbf{x}, s-1} \sim \mathcal{N}\left(0, \frac{\sigma_{\mathbf{x}}^2}{m}\left(I_{d_1}+\left(1-\beta_{\mathbf{x}}\right)^2 I_{d_1}\right)\right)
\end{equation}
Therefore, we have:
\begin{equation}
\begin{aligned}
& \sum_{s=0}^{T-1} \mathbb{E}\left\|\mathcal{N}_{\mathbf{x}, s}-\left(1-\beta_{\mathbf{x}}\right) \mathcal{N}_{\mathbf{x}, s-1}\right\|^2=T  \frac{\sigma_{\mathbf{x}}^2}{m}  d_1 \left(1+\left(1-\beta_{\mathbf{x}}\right)^2\right) \\
& \sum_{s=0}^{T-1} \mathbb{E}\left\|\mathcal{N}_{\mathbf{y}, s}-\left(1-\beta_{\mathbf{y}}\right) \mathcal{N}_{\mathbf{y}, s-1}\right\|^2=T  \frac{\sigma_{\mathbf{y}}^2}{m} d_2 \left(1+\left(1-\beta_{\mathbf{y}}\right)^2\right)
\end{aligned}
\end{equation}
Therefore, Eq.(\ref{eqt1.6}) can be written as:
\begin{equation}
\begin{aligned}
\frac{1}{T} \sum_{t=0}^{T-1} \mathbb{E}\left\|\nabla \Phi\left(\bar{\mathbf{x}}_t\right)\right\|^2 \leq & L\left(\Phi\left(\mathbf{x}_0\right)-\Phi^*\right) \epsilon^2+2 L^2 \delta_0 \epsilon^2+2 \sigma^2 \epsilon^2+\frac{\epsilon^2}{m} \sum_{i=1}^m\left\|\nabla_{\mathbf{x}} f_i\left(\mathbf{x}_0, \mathbf{y}_0\right)\right\|^2 \\
& +\frac{\epsilon^2}{m} \sum_{i=1}^m\left\|\nabla_{\mathbf{x}} f_i\left(\mathbf{x}_0, \mathbf{y}_0\right)\right\|^2+\frac{(2m^2\epsilon^2+6\kappa^2)}{m} \left(\sigma_\mathbf{x}^2 d_1 +\sigma_\mathbf{y}^2 d_2\right)\\
& + \frac{16\sigma_\mathbf{x}^2d_1}{m} +\frac{320\kappa^2\sigma_\mathbf{y}^2d_2}{m}
\end{aligned}
\end{equation}
where we use following inequalities for simplification.
\begin{equation}
\begin{gathered}
\beta_\mathbf{x} \geq \beta_\mathbf{y}, 
4 L^2\left(21 \kappa^2+\frac{12}{m \beta_\mathbf{x}}+\frac{240 \kappa^2}{m \beta_\mathbf{y}}\right) \leq 100 L^2 \kappa^2+\frac{1000 L^2 \kappa^2}{m \beta_\mathbf{y}} \\
\frac{L^2 \beta_\mathbf{x} \eta_\mathbf{x}^2}{(1-\lambda)^4 b_0 T} \leq \frac{\epsilon(\min \{1, m \epsilon\})^5 \epsilon^2}{20 \cdot 400 \kappa \cdot 30000 \kappa^3\left(15000 \kappa^3\right)^2}, \frac{L^2 \beta_\mathbf{y} \eta_\mathbf{y}^2}{(1-\lambda)^4 b_0 T} \leq \frac{\epsilon(\min \{1, m \epsilon\})^5 \epsilon^2}{500 \cdot 400 \kappa \cdot 30000 \kappa^3(1500 \kappa)^2} \\
\frac{L^2 \beta_\mathbf{x}^2 \eta_\mathbf{x}^2}{(1-\lambda)^4} \leq \frac{\epsilon^2(\min \{1, m \epsilon\})^4}{400\left(15000 \kappa^3\right)^2}, \frac{L^2 \beta_\mathbf{y}^2 \eta_\mathbf{y}^2}{(1-\lambda)^4} \leq \frac{\epsilon^2(\min \{1, m \epsilon\})^4}{\left(500 \kappa^2\right)^2(1500 \kappa)^2}
\end{gathered}
\end{equation}

Therefore, if $T$ is determined by $\epsilon$, we have the first conclusion in Theorem \ref{theorem 1}:
\begin{equation}
\begin{aligned}
\frac{1}{T} \sum_{t=0}^{T-1} \mathbb{E}\left\|\nabla \Phi\left(\bar{\mathbf{x}}_t\right)\right\|^2 = & \mathcal{O}\left(\epsilon^2\right)+\mathcal{O}\left(m\epsilon^2\right)+\mathcal{O}\left(\sigma_\mathbf{x}^2d_1+\sigma_\mathbf{y}^2d_2\right) 
\end{aligned}
\end{equation}

In the above proof, we have established the convergence result when $T$ is determined by $\epsilon$. Next, we analyze the case when $T$ is uncertain, for which we provide the following proof. Before presenting our proof, we first provide some definitions regarding $T$.
\begin{equation}
\label{eqt2.1}
T_0 \geq 10 m^2, \quad T=\frac{30000 \kappa^3 T_0}{(1-\lambda)^2}
\end{equation}

Similarly, we can obtain:

\begin{equation}
\begin{aligned}
& \frac{1}{T} \sum_{t=0}^{T-1} \mathbb{E}\left\|\nabla \Phi\left(\bar{\mathbf{x}}_t\right)\right\|^2 \\
&\leq \frac{2\left(\Phi\left(\mathbf{x}_0\right)-\Phi^*\right)}{\eta_\mathbf{x} T}+\frac{8 \kappa L^2 \delta_0}{T L \eta_\mathbf{y}}+\frac{8 \sigma^2}{m b_0 T}\left(\frac{1}{\beta_\mathbf{x}}+\frac{20 \kappa^2}{\beta_\mathbf{y}}\right)+\frac{8 \sigma^2}{m}\left(\beta_\mathbf{x}+20 \kappa^2 \beta_\mathbf{y}\right) \\
& \quad+\frac{4 L^2}{m T}\left(21 \kappa^2+\frac{12}{m \beta_\mathbf{x}}+\frac{240 \kappa^2}{m \beta_\mathbf{y}}\right)\left(\frac{16 \lambda^2 \eta_\mathbf{x}^2}{\left(1-\lambda^2\right)^3} \mathbb{E}\left\|V_0-\bar{V}_0\right\|_F^2+\frac{16 \lambda^2 \eta_\mathbf{y}^2}{\left(1-\lambda^2\right)^3} \mathbb{E}\left\|U_0-\bar{U}_0\right\|_F^2\right. \\
&\quad \left.+\frac{64 m \lambda^4\left(\beta_\mathbf{x} \eta_\mathbf{x}^2+\beta_\mathbf{y} \eta_\mathbf{y}^2\right) \sigma^2}{\left(1-\lambda^2\right)^4 b_0}+\frac{192 m \lambda^4\left(\beta_\mathbf{x}^2 \eta_\mathbf{x}^2+\beta_\mathbf{y}^2 \eta_\mathbf{y}^2\right) \sigma^2 T}{\left(1-\lambda^2\right)^4}\right)+\frac{8}{T} \sum_{s=0}^{T-1} \mathbb{E}\left\|\mathcal{N}_{\mathbf{x}, s}-\mathcal{N}_{\mathbf{x}, s-1}\right\|^2\\
&\quad+\frac{160\kappa^2}{T} \sum_{s=0}^{T-1} \mathbb{E}\left\|\mathcal{N}_{\mathbf{y}, s}-\mathcal{N}_{\mathbf{y}, s-1}\right\|^2+\frac{256 L^2 \lambda^4 \eta_\mathbf{x}^2}{ T \left(1-\lambda^2\right)^3}\left(21 \kappa^2+\frac{12}{m \beta_\mathbf{x}}+\frac{240 \kappa^2}{m \beta_\mathbf{y}}\right)\sum_{s=0}^{T-1} \\
&\quad\quad\mathbb{E}\left\|\mathcal{N}_{\mathbf{x}, s}-\mathcal{N}_{\mathbf{x}, s-1}\right\|^2+\frac{256 L^2 \lambda^4 \eta_\mathbf{y}^2}{ T \left(1-\lambda^2\right)^3}\left(21 \kappa^2+\frac{12}{m \beta_\mathbf{x}}+\frac{240 \kappa^2}{m \beta_\mathbf{y}}\right)\sum_{s=0}^{T-1}\mathbb{E}\left\|\mathcal{N}_{\mathbf{y}, s}-\mathcal{N}_{\mathbf{y}, s-1}\right\|^2
\end{aligned}
\end{equation}

Substituting the parameter values given in above Eq.(\ref{eqt2.1}) and the relationships between all these parameters in Theorem \ref{theorem 1}, and using the scaling method to simplify the calculations, we can get:
\begin{equation}
\begin{aligned}
\frac{1}{T} \sum_{t=0}^{T-1} \mathbb{E}\left\|\nabla \Phi\left(\bar{\mathbf{x}}_t\right)\right\|^2 \leq & \frac{L\left(\Phi\left(\mathbf{x}_0\right)-\Phi^*\right)+2\sigma^2+2L^2 \delta_0}{\left(m T_0\right)^{2 / 3}}+\frac{\frac{1}{m} \sum_{i=1}^m \mathbb{E}\left\|\nabla_\mathbf{x} f_i\left(\mathbf{x}_0, \mathbf{y}_0\right)\right\|^2}{T_0} \\
& +\frac{\frac{1}{m} \sum_{i=1}^m \mathbb{E}\left\|\nabla_\mathbf{y} f_i\left(\mathbf{x}_0, \mathbf{y}_0\right)\right\|^2}{T_0}+\left(\frac{2m^{2/3}}{T_0^{2/3}}+\frac{3\kappa^2}{m}\right) \left(\sigma_\mathbf{x}^2 d_1+\sigma_\mathbf{y}^2 d_2\right)\\
& +\frac{16 \sigma_{\mathbf{x}}^2 d_1}{m}+\frac{320 \kappa^2 \sigma_{\mathbf{y}}^2 d_2}{m}
\end{aligned}
\end{equation}
Therefore, if the number of iteration is not fixed, we have the second conclusion in Theorem \ref{theorem 1}, we have:
\begin{equation}
\begin{aligned}
\frac{1}{T} \sum_{t=0}^{T-1} \mathbb{E}\left\|\nabla \Phi\left(\bar{\mathbf{x}}_t\right)\right\|^2 = & \mathcal{O}\left(\frac{1}{\left(m T_0\right)^{2 / 3}}\right)+\mathcal{O}\left(\frac{1}{T_0}\right)+\mathcal{O}\left(\frac{m^{1/3}}{T_0^{2/3}}\right)+\mathcal{O}\left(\sigma_\mathbf{x}^2d_1+\sigma_\mathbf{y}^2d_2\right) 
\end{aligned}
\end{equation}

\section{Proof of privacy guarantee}\label{sec:apendixb}
Before we start our privacy analysis, let's introduce moments accountant method \cite{abadi2016deep}.\\
\textbf{Definition \refstepcounter{definition}\label{definition 2}2.} [Privacy Loss \citep{abadi2016deep}] For adjacent datasets $D$ and $D^{\prime}$, mechanism $\mathcal{M}$, and output $o \in \mathbb{R}$, the privacy loss at $o$ is defined as:
\begin{equation}
\begin{aligned}
c\left(o ; \mathcal{M}, D, D^{\prime}\right) = \log \left(\frac{\mathbb{P}[\mathcal{M}(D) = o]}{\mathbb{P}\left[\mathcal{M}\left(D^{\prime}\right) = o\right]}\right)
\end{aligned}
\end{equation}
\textbf{Definition \refstepcounter{definition}\label{definition 3}3.} [Moment \citep{abadi2016deep}] For a mechanism $\mathcal{M}$ and the privacy loss at output $o$, the $\lambda$-th moment is defined as:
\begin{equation}
\begin{aligned}
\alpha_{\mathcal{M}}\left(\lambda ; D, D^{\prime}\right) = \log \left(\mathbb{E}_{o \sim \mathcal{M}(D)} \left[\exp \left(\lambda c\left(o ; \mathcal{M}, D, D^{\prime}\right)\right)\right]\right)
\end{aligned}
\end{equation}
with the upper bound given by:
\begin{equation}
\begin{aligned}
\alpha_{\mathcal{M}}(\lambda) = \max_{D, D^{\prime}} \alpha_{\mathcal{M}}\left(\lambda ; D, D^{\prime}\right)
\end{aligned}
\end{equation}
\textbf{Lemma \refstepcounter{lemma}\label{lemma 14}14.} [Composability \citep{abadi2016deep}] Let $\alpha_{\mathcal{M}}(\lambda)$ be defined as above. Suppose $\mathcal{M}$ is composed of several mechanisms $\mathcal{M}_1, \dots, \mathcal{M}_k$, where $\mathcal{M}_i$ depends on $\mathcal{M}_1, \dots, \mathcal{M}_{i-1}$. Then, for any $\lambda$:
\begin{equation}
\begin{aligned}
\alpha_{\mathcal{M}}(\lambda) \leq \sum_{i=1}^k \alpha_{\mathcal{M}_i}(\lambda)
\end{aligned}
\end{equation}
\textbf{Lemma \refstepcounter{lemma}\label{lemma 15}15.} [Tail Bound \citep{abadi2016deep}] For any $\theta > 0$, mechanism $\mathcal{M}$ is $(\theta, \gamma)$-differentially private if:
\begin{equation}
\begin{aligned}
\gamma = \min_{\lambda} \exp \left(\alpha_{\mathcal{M}}(\lambda) - \lambda \theta\right)
\end{aligned}
\end{equation}
\textbf{Definition \refstepcounter{definition}\label{definition 4}4.} [Rényi Divergence \citep{bun2016concentrated}] Let $P$ and $Q$ be probability distributions. For $\rho \in (1, \infty)$, the Rényi Divergence of order $\rho$ between $P$ and $Q$ is defined as:
\begin{equation}
\begin{aligned}
D_\rho(P \| Q) = \frac{1}{\rho - 1} \log \left(\mathbb{E}_{\mathbf{x} \sim P}\left[\left(\frac{P(\mathbf{x})}{Q(\mathbf{x})}\right)^{\rho - 1}\right]\right)
\end{aligned}
\end{equation}
\textbf{Lemma \refstepcounter{lemma}\label{lemma 16}16.} For Gaussian distributions $\mathcal{N}(\mu, \sigma^2 I_p)$ and $\mathcal{N}(\nu, \sigma^2 I_p)$, where $\mu, \nu \in \mathbb{R}^p$, $\sigma \in \mathbb{R}$, and $\rho \in (1, \infty)$, the Rényi Divergence of order $\rho$ is given by:
\begin{equation}
\begin{aligned}
D_\rho\left(\mathcal{N}\left(\mu, \sigma^2 I_p\right) \| \mathcal{N}\left(\nu, \sigma^2 I_p\right)\right) = \frac{\rho \|\mu - \nu\|_2^2}{2 \sigma^2}
\end{aligned}
\end{equation}
Firstly, let's review training process in DP-DM-HSGD:
\begin{equation}
\begin{aligned}
\bar{\mathbf{x}}_{t+1} & =\bar{\mathbf{x}}_t-\eta_\mathbf{x}\left(\bar{\mathbf{g}}_t + \mathcal{N}_{\mathbf{x}, t}\right) \\
\bar{\mathbf{y}}_{t+1} & =\bar{\mathbf{y}}_t+\eta_\mathbf{y}\left(\bar{\mathbf{h}}_t + \mathcal{N}_{y, t}\right)
\end{aligned}
\end{equation}
where $\mathcal{N}_{\mathbf{x}, t} \sim \mathcal{N}\left(0, \frac{\sigma_\mathbf{x}^2}{m} I_{d_1}\right)$, $\mathcal{N}_{\mathbf{y}, t} \sim \mathcal{N}\left(0, \frac{\sigma_\mathbf{y}^2}{m} I_{d_2}\right)$, and 
$
\sigma_\mathbf{x} = c\left(\frac{L_g \sqrt{\left(\frac{8 T(T+1)(2 T+1)}{3}+4 T\right) \log (1 / \gamma)}}{2 \theta \sqrt{m}}\right)
$

Now, we will present the full proof for Theorem \ref{theorem 2}. We first analyze parameter $\mathbf{x}$.\\
When updating $\mathbf{x}$, at iteration $t$, the randomized mechanism $\mathcal{M}_t$ which may disclose privacy is
\begin{equation}
\begin{aligned}
\mathcal{M}_t&=\bar{\mathbf{g}}_t + \mathcal{N}_{\mathbf{x}, t}\\
&=\frac{1}{m}  \sum_{j=1}^{m}\bigg(\sum_{k=0}^{t}\left(1-\beta_\mathbf{x}\right)^{t-k}\left[\nabla_\mathbf{x} F_j\left(\mathbf{x}_k^{(i)}, \mathbf{y}_k^{(i)} ; \mathbf{z}_k^{(i)}\right) -\nabla_\mathbf{x} F_j\left(\mathbf{x}_k^{(i)}, \mathbf{y}_k^{(i)} ; \mathbf{z}_{k+1}^{(i)}\right)\right]\\
&\quad+\nabla_\mathbf{x} F_j\left(\mathbf{x}_t^{(i)}, \mathbf{y}_t^{(i)} ; \mathbf{z}_{t+1}^{(i)}\right)\bigg) +\mathcal{N}_{\mathbf{x}, t}
\end{aligned}
\end{equation}
We set probability distribution of $\mathcal{M}_t$ over adjacent datasets $D, D^{\prime}$ as $P$ and $Q$, respectively, also, we assume the single different data sample is on the $m^{\text {th }}$one, and we obtain:
\begin{equation}
\begin{aligned}
P &=\frac{1}{m} \sum_{i=1}^{m-1}\bigg(\sum_{k=0}^{t}\left(1-\beta_\mathbf{x}\right)^{t-k}\left[\nabla_\mathbf{x} F_j\left(\mathbf{x}_k^{(i)}, \mathbf{y}_k^{(i)} ; \mathbf{z}_k^{(i)}\right) -\nabla_\mathbf{x} F_j\left(\mathbf{x}_k^{(i)}, \mathbf{y}_k^{(i)} ; \mathbf{z}_{k+1}^{(i)}\right)\right]\\
&\quad+\nabla_\mathbf{x} F_j\left(\mathbf{x}_t^{(i)}, \mathbf{y}_t^{(i)} ; \mathbf{z}_{t+1}^{(i)}\right)\bigg)+\frac{1}{m} \bigg(\sum_{k=0}^{t}\left(1-\beta_\mathbf{x}\right)^{t-k}\left[\nabla_\mathbf{x} F_j\left(\mathbf{x}_k^{(m)}, \mathbf{y}_k^{(m)} ; \mathbf{z}_k^{(m)}\right) \right.\\
&\left.\quad-\nabla_\mathbf{x} F_j\left(\mathbf{x}_k^{(m)}, \mathbf{y}_k^{(m)} ; \mathbf{z}_{t+1}^{(m)}\right)\right]+\nabla_\mathbf{x} F_j\left(\mathbf{x}_t^{(m)}, \mathbf{y}_t^{(m)} ; \mathbf{z}_{t+1}^{(m)}\right)\bigg)+\mathcal{N}_{\mathbf{x}, t} \\
Q &=\frac{1}{m} \sum_{i=1}^{m-1} \bigg(\sum_{k=0}^{t}\left(1-\beta_\mathbf{x}\right)^{t-k}\left[\nabla_\mathbf{x} F_j\left(\mathbf{x}_k^{(i)}, \mathbf{y}_k^{(i)} ; \mathbf{z}_k^{(i)}\right) -\nabla_\mathbf{x} F_j\left(\mathbf{x}_k^{(i)}, \mathbf{y}_k^{(i)} ; \mathbf{z}_{k+1}^{(i)}\right)\right]\\
&\quad+\nabla_\mathbf{x} F_j\left(\mathbf{x}_t^{(i)}, \mathbf{y}_t^{(i)} ; \mathbf{z}_{t+1}^{(i)}\right)\bigg)+\frac{1}{m} \bigg(\sum_{k=0}^{t}\left(1-\beta_\mathbf{x}\right)^{t-k}\left[\nabla_\mathbf{x} F_j\left(\mathbf{x}_k^{(m)}, \mathbf{y}_k^{(m)} ; \mathbf{z}_k^{(m)\prime}\right) \right.\\
&\left.\quad-\nabla_\mathbf{x} F_j\left(\mathbf{x}_k^{(m)}, \mathbf{y}_k^{(m)} ; \mathbf{z}_{k+1}^{(m)\prime}\right)\right]+\nabla_\mathbf{x} F_j\left(\mathbf{x}_t^{(m)}, \mathbf{y}_t^{(m)} ; \mathbf{z}_{t+1}^{(m)\prime}\right)\bigg)+\mathcal{N}_{\mathbf{x}, t} 
\end{aligned}
\end{equation}
For the simplicity of the next steps, we set 
\begin{equation}
\begin{aligned}
\mathcal{I}&=\frac{1}{m} \sum_{i=1}^{m-1}\bigg(\sum_{k=0}^{t}\left(1-\beta_\mathbf{x}\right)^{t-k}\left[\nabla_\mathbf{x} F_j\left(\mathbf{x}_k^{(i)}, \mathbf{y}_k^{(i)} ; \mathbf{z}_k^{(i)}\right) -\nabla_\mathbf{x} F_j\left(\mathbf{x}_k^{(i)}, \mathbf{y}_k^{(i)} ; \mathbf{z}_{k+1}^{(i)}\right)\right]\\
&\quad+\nabla_\mathbf{x} F_j\left(\mathbf{x}_t^{(i)}, \mathbf{y}_t^{(i)} ; \mathbf{z}_{t+1}^{(i)}\right)\bigg)+\frac{1}{m} \bigg(\sum_{k=0}^{t}\left(1-\beta_\mathbf{x}\right)^{t-k}\left[\nabla_\mathbf{x} F_j\left(\mathbf{x}_k^{(m)}, \mathbf{y}_k^{(m)} ; \mathbf{z}_k^{(m)}\right) \right.\\
&\left.\quad-\nabla_\mathbf{x} F_j\left(\mathbf{x}_k^{(m)}, \mathbf{y}_k^{(m)} ; \mathbf{z}_{t+1}^{(m)}\right)\right]+\nabla_\mathbf{x} F_j\left(\mathbf{x}_t^{(m)}, \mathbf{y}_t^{(m)} ; \mathbf{z}_{t+1}^{(m)}\right)\bigg)\\
\mathcal{I^\prime}&=\frac{1}{m} \sum_{i=1}^{m-1} \bigg(\sum_{k=0}^{t}\left(1-\beta_\mathbf{x}\right)^{t-k}\left[\nabla_\mathbf{x} F_j\left(\mathbf{x}_k^{(i)}, \mathbf{y}_k^{(i)} ; \mathbf{z}_k^{(i)}\right) -\nabla_\mathbf{x} F_j\left(\mathbf{x}_k^{(i)}, \mathbf{y}_k^{(i)} ; \mathbf{z}_{k+1}^{(i)}\right)\right]\\
&\quad+\nabla_\mathbf{x} F_j\left(\mathbf{x}_t^{(i)}, \mathbf{y}_t^{(i)} ; \mathbf{z}_{t+1}^{(i)}\right)\bigg)+\frac{1}{m} \bigg(\sum_{k=0}^{t}\left(1-\beta_\mathbf{x}\right)^{t-k}\left[\nabla_\mathbf{x} F_j\left(\mathbf{x}_k^{(m)}, \mathbf{y}_k^{(m)} ; \mathbf{z}_k^{(m)\prime}\right) \right.\\
&\left.\quad-\nabla_\mathbf{x} F_j\left(\mathbf{x}_k^{(m)}, \mathbf{y}_k^{(m)} ; \mathbf{z}_{k+1}^{(m)\prime}\right)\right]+\nabla_\mathbf{x} F_j\left(\mathbf{x}_t^{(m)}, \mathbf{y}_t^{(m)} ; \mathbf{z}_{t+1}^{(m)\prime}\right)\bigg)
\end{aligned}
\end{equation}

As we define $\mathbf{z}^{(i)}$ as an index sample in local dataset $\mathcal{Z}$, therefore, according to Assumption \ref{assumption 7}:
\begin{equation}
\left\|\nabla_\mathbf{x} F_i\left(\mathbf{x}_t^{(i)}, \mathbf{y}_t^{(i)} ; \mathbf{z}_t^{(i)}\right) -\nabla_\mathbf{x} F_i\left(\mathbf{x}_t^{(i)}, \mathbf{y}_t^{(i)} ; \mathbf{z}_t^{(i)\prime}\right)\right\|_2 \leq 2L_g
\end{equation}
The inequality above stands because $\mathbf{z}_t^{(m)}$ and $\mathbf{z}_t^{(m)\prime}$ are adjacent data samples. Now, we can get:
\begin{equation}
\begin{aligned}
&\left\|\mathcal{I}-\mathcal{I^\prime}\right\|_2^2\\
&=\bigg\|\frac{1}{m} \bigg(\sum_{k=0}^{t}\left(1-\beta_\mathbf{x}\right)^{t-k}\left[\nabla_\mathbf{x} F_j\left(\mathbf{x}_k^{(m)}, \mathbf{y}_k^{(m)} ; \mathbf{z}_k^{(m)}\right) -\nabla_\mathbf{x} F_j\left(\mathbf{x}_k^{(m)}, \mathbf{y}_k^{(m)} ; \mathbf{z}_{k+1}^{(m)}\right)\right]\\\
&\quad+\nabla_\mathbf{x} F_j\left(\mathbf{x}_t^{(m)}, \mathbf{y}_t^{(m)} ; \mathbf{z}_{t+1}^{(m)}\right)\bigg) -\frac{1}{m} \bigg(\sum_{k=0}^{t}\left(1-\beta_\mathbf{x}\right)^{t-k}\left[\nabla_\mathbf{x} F_j\left(\mathbf{x}_k^{(m)}, \mathbf{y}_k^{(m)} ; \mathbf{z}_k^{(m)\prime}\right) \right.\\
&\left.\quad-\nabla_\mathbf{x} F_j\left(\mathbf{x}_k^{(m)}, \mathbf{y}_k^{(m)} ; \mathbf{z}_{k+1}^{(m)\prime}\right)\right]+\nabla_\mathbf{x} F_j\left(\mathbf{x}_t^{(m)}, \mathbf{y}_t^{(m)} ; \mathbf{z}_{t+1}^{(m)\prime}\right)\bigg)\bigg\|_2^2\\
&\leq \frac{2}{m^2} \bigg( \sum_{k=0}^{t}\left(1-\beta_\mathbf{x}\right)^{t-k}\left(\left\|\nabla_\mathbf{x} F_j\left(\mathbf{x}_k^{(m)}, \mathbf{y}_k^{(m)} ; \mathbf{z}_k^{(m)}\right) -\nabla_\mathbf{x} F_j\left(\mathbf{x}_k^{(m)}, \mathbf{y}_k^{(m)} ; \mathbf{z}_k^{(m)\prime}\right)\right\|_2\right.\\
&\left.\quad-\left\|\nabla_\mathbf{x} F_j\left(\mathbf{x}_k^{(m)}, \mathbf{y}_k^{(m)} ; \mathbf{z}_{k+1}^{(m)}\right) -\nabla_\mathbf{x} F_j\left(\mathbf{x}_k^{(m)}, \mathbf{y}_k^{(m)} ; \mathbf{z}_{k+1}^{(m)\prime}\right)\right\|_2\right)\bigg)^2\\
&\quad+\frac{2}{m^2 } \left(\left\|\nabla_\mathbf{x} F_j\left(\mathbf{x}_t^{(m)}, \mathbf{y}_t^{(m)} ; \mathbf{z}_{t+1}^{(m)}\right)-\nabla_\mathbf{x} F_j\left(\mathbf{x}_t^{(m)}, \mathbf{y}_t^{(m)} ; \mathbf{z}_{t+1}^{(m)\prime}\right)\right\|_2\right)^2 \\
&\leq \frac{2}{m^2} \left( \sum_{k=0}^{t}\left(1-\beta_\mathbf{x}\right)^{t-k} 4L_g\right)^2+ \frac{8L_g^2}{m^2} \\
&\leq \frac{2L_g^2}{m^2} \left(16 \left(\sum_{k=0}^{t}\left(1-\beta_\mathbf{x}\right)^{t-k}\right)^2+ 4 \right) \leq \frac{2L_g^2}{m^2}\left(16t^2+4\right)=\frac{L_g^2}{m^2} \left(32t^2+8\right)
\end{aligned}
\end{equation}
where we use Young's inequality in the first inequality, and simplify $\sum_{k=0}^{t}\left(1-\beta_\mathbf{x}\right)^{t-k}$ to $\sum_{k=0}^{t} 1$  in last inequality as we have a bound 
$ 0<\beta_x<1 $. Noting that $\mathcal{N}_{\mathbf{x}, t} \sim \mathcal{N}\left(0, \frac{\sigma_\mathbf{x}^2}{m} I_p\right)$, we have
\begin{equation}
\begin{aligned}
P\sim\mathcal{N}\left(\mathcal{I}, \frac{\sigma_\mathbf{x}^2}{m} I_p\right), 
Q\sim\mathcal{N}\left(\mathcal{I^\prime}, \frac{\sigma_\mathbf{x}^2}{m} I_p\right)
\end{aligned}
\end{equation}
With Definition \ref{definition 2} and \ref{definition 3}, we obtain:
\begin{equation}
\begin{aligned}
\alpha_{\mathcal{M}_t}\left(\lambda ; D, D^{\prime}\right)=\log \left(\mathbb{E}_{o \sim P}\left[\exp \left(\lambda \log \left(\frac{P}{Q}\right)\right)\right]\right)=\log \left(\mathbb{E}_{o \sim P}\left[\left(\frac{P}{Q}\right)^\lambda\right]\right)=\lambda D_{\lambda+1}(P \| Q)
\end{aligned}
\end{equation}
Where we use Definition \ref{definition 4} to get the last inequality. From lemma \ref{lemma 16}, we have:
\begin{equation}
\begin{aligned}
\alpha_{\mathcal{M}_t}\left(\lambda ; D, D^{\prime}\right)=\frac{m\lambda(\lambda+1)\left\|\mathcal{I}-\mathcal{I^\prime}\right\|_2^2}{2 \sigma^2} \leq \frac{\left(16t^2+4\right) L_g^2 \lambda(\lambda+1)}{m \sigma^2}=\alpha_{\mathcal{M}_t}(\lambda)
\end{aligned}
\end{equation}
 The inequality holds because $F_i(\cdot, \cdot ; \cdot)$ is $L$-Lipschitz, and the last step holds because of Definition \ref{definition 3}. By Lemma \ref{lemma 14}, there are $T$ iterations, so we have
\begin{equation}
\begin{aligned}
\alpha_{\mathcal{M}}(\lambda) \leq \sum_{t=1}^T \alpha_{\mathcal{M}_t}(\lambda) \leq \frac{2\left(16t^2+4\right) L_g^2 \lambda^2 }{m \sigma^2}
\end{aligned}
\end{equation}
where the last inequality holds because $\lambda \in(1, \infty)$. We also assume that the maximum value of $\alpha_{\mathcal{M}_t}(\lambda)$ is no greater than twice the average of its sum.\\
Taking $\sigma_\mathbf{x}=c \frac{L_g \sqrt{\left(\frac{8 T(T+1)(2 T+1)}{3}+4 T\right) \log (1 / \gamma)}}{2 \theta \sqrt{m}}$, we can guarantee $\alpha_{\mathcal{M}}(\lambda) \leq \lambda \theta / 2$, and as the consequence, by Lemma \ref{lemma 16}, we obtain $\gamma \leq \exp (-\lambda \theta / 2)$, and in this way, it leads $(\theta, \gamma)$-DP to parameter $\bar{\mathbf{x}}$. Similarly, we have the same proof for $\bar{\mathbf{y}}_t$, if $\mathcal{N}_{y, t} \sim \mathcal{N}\left(0, \frac{\sigma_\mathbf{y}^2}{m} I_p\right)$ with $\sigma_\mathbf{y}=c \frac{L_g \sqrt{\left(\frac{8 T(T+1)(2 T+1)}{3}+4 T\right) \log (1 / \gamma)}}{2 \theta \sqrt{m}}$ is used when updating $\bar{\mathbf{y}}_t$, then $(\theta, \gamma)$-DP can be guaranteed.
The proof is completed.

\section{Additional Experiments}
\label{sec:appendixC}

\subsection{Gradient clipping}
We conducted experiments on \alg and DM-HSGD with a clipping threshold set to clip the top 20\% of gradients, which yielded similar results, further confirming the reliability of our original findings. See in Table \ref{table3}, here $\sigma$ represents the intensity of the noise added in the DPMixSGD algorithm.

\begin{table}[h]
\centering
\caption{AUC Performance of DPMixSGD and DM-HSGD under Different Noise Levels with Gradient Clipping Comparison}
\resizebox{\columnwidth}{!}{%
\begin{tabular}{|c|c|c|c|c|c|}
\hline
\multirow{2}{*}{\textbf{Dataset}} & \multirow{2}{*}{\textbf{DM-HSGD}} & \multicolumn{2}{c|}{$\sigma = 0.5$} & \multicolumn{2}{c|}{$\sigma = 1$} \\
\cline{3-6}
& & \textbf{DPMixSGD} & \textbf{DPMixSGD (Clip)} & \textbf{DPMixSGD} & \textbf{DPMixSGD (Clip)} \\
\hline
MNIST & 0.9937 & \cellcolor{greyL}0.9897 & \cellcolor{greyL}0.9733 & \cellcolor{greyL}0.9796 & \cellcolor{greyL}0.9548 \\
\hline
Fashion\_MNIST & 0.9859 & \cellcolor{greyL}0.9757 & \cellcolor{greyL}0.9493 & \cellcolor{greyL}0.9627 & \cellcolor{greyL}0.9184 \\
\hline
ljcnn1 & 0.9984 & \cellcolor{greyL}0.9962 & \cellcolor{greyL}0.9889 & \cellcolor{greyL}0.9901 & \cellcolor{greyL}0.9711 \\
\hline
\end{tabular}
}

\label{table3}
\end{table}

\subsection{Decentralized min-max problem in multilayer perceptron of image classification problem}

This experiment focuses on image classification of the Fashion-MNIST~\citep{xiao2017fashion} dataset using a multilayer perceptron (MLP) model. We introduce corresponding dual variables to formulate a min-max problem. Additionally, we also compare the AUROC performance of the \alg, DM-HSGD, SGDA, and DP-SGDA algorithms across different scenarios.
In this problem, we consider a distributed network composed of $m$ agents. Each agent $i$ possesses its own model parameter $\mathbf{x}_i$ as well as a set of dual variables $y_{a,i}$, $y_{b,i}$, and $y_{w,i}$. These dual variables are typically employed to handle constraints or to model adversarial factors. The optimization objective of the entire MLP system is defined as follows:
\begin{align}
\min_{\{\mathbf{x}_i\}_{i=1}^m} \max_{\{y_{a,i}, y_{b,i}, y_{w,i}\}_{i=1}^m} \Phi\left(\{\mathbf{x}_i\}, \{y_{a,i}, y_{b,i}, y_{w,i}\}\right),
\end{align}
where $\Phi$ is the global objective function, defined as the average of all agents' local objective functions as the following:
\begin{align}
\Phi\left(\{\mathbf{x}_i\}, \{y_{a,i}, y_{b,i}, y_{w,i}\}\right) = \frac{1}{m} \sum_{i=1}^{m} \phi_i\left(\mathbf{x}_i, y_{a,i}, y_{b,i}, y_{w,i}\right).
\end{align}

Each agent $i$ has a local optimization function $\phi_i$ defined as:
\begin{align}
&\phi_i\left(\mathbf{x}_i, y_{a,i}, y_{b,i}, y_{w,i}\right)  \nonumber \\
&=\mathcal{L}\left(\mathbf{x}_i; \mathcal{D}_i\right) + y_{a,i} \cdot f_a\left(\mathbf{x}_i\right) 
 + y_{b,i} \cdot f_b\left(\mathbf{x}_i\right) + y_{w,i} \cdot f_w\left(\mathbf{x}_i\right),
\end{align}
where $\mathcal{L}\left(\mathbf{x}_i; \mathcal{D}_i\right)$ is the primary loss function based on the local dataset $\mathcal{D}_i$ (e.g., cross-entropy loss).
$f_a\left(\mathbf{x}_i\right)$, $f_b\left(\mathbf{x}_i\right)$, and $f_w\left(\mathbf{x}_i\right)$ are auxiliary functions associated with the dual variables, introduce to impose additional constraints or model adversarial factors.
$y_{a,i}$, $y_{b,i}$, and $y_{w,i}$ are the corresponding dual variables, typically acting as lagrange multipliers to balance the primary loss with the auxiliary terms.

For the image classification algorithms \alg, DM-HSGD, SGDA, and DP-SGDA, we conduct extensive experimental validations and compare their AUROC metrics. The primary parameters involved in the experiments are as follows: the learning rates for the model parameters $\mathbf{x}$ and their dual variables $\mathbf{y}$ are selected from the set $\{0.01, 0.001, 0.0001\}$. The mini-batch size is set to $64$. Specifically, for the \alg and DM-HSGD algorithms, the initial batch size is set to $b_0 = 64$. The gradient weight adjustment parameters $\beta_x$ and $\beta_y$ are chosen from the set $\{0.1, 0.01\}$. Table~\ref{table2} illustrates the AUROC results over epochs for each algorithm during the image classification experiments. In all compared groups, our proposed method surpasses existing algorithms, because the introduced noise aids in escaping saddle points while expediting the model's training process.

\begin{table}[t]
\centering
\caption{AUROC results over epochs for each algorithm during the image classification experiments on Fashion-MNIST dataset.}
 % \begin{small}
  \begin{normalsize}
  \centering
  \subfloat[Impact of total number of agents $m$.]
{
  \begin{tabular}{|c|c|c|c|c|}
    \hline
    $m$ & $m=5$ & $m=10$ & $m=15$ & $m=20$ \\
    \hline
    SGDA & 0.7978 & 0.7227 & 0.5602 & 0.5503 \\ \hline
    DP-SGDA & 0.7754 & 0.7251 & 0.5367 & 0.5506 \\ \hline
    DM-HSGD & 0.9352 & 0.9179 & 0.9345 & 0.9087 \\ \hline
\rowcolor{greyL}
    \algns & 0.9311 & 0.9310 & 0.9296 & 0.9317 \\
    \hline
  \end{tabular}
  }
\\
  \subfloat[Impact of sparsity level $p$.]
{
  \begin{tabular}{|c|c|c|c|c|}
    \hline
    $p$ & $p=0.2$ & $p=0.5$ & $p=0.8$ & $p=1$ \\
    \hline
    SGDA & 0.7881 & 0.7978 & 0.7978 & 0.7971 \\ \hline
    DP-SGDA & 0.7816 & 0.7796 & 0.7754 & 0.7769 \\ \hline
    DM-HSGD & 0.9359 & 0.9357 & 0.9352 & 0.9329 \\ \hline
\rowcolor{greyL}
    DPMixSGD & 0.9328 & 0.9373 & 0.9311 & 0.9369 \\
    \hline
  \end{tabular}
}
\\
  \subfloat[Impact of $\theta$.]
{
  \begin{tabular}{|c|c|c|c|c|}
    \hline
    $\theta$ & $\theta=0.005$ & $\theta=0.01$ &$\theta=0.05$ & $\theta=0.1$ \\
    \hline
    SGDA & 0.7978 & 0.7978 & 0.7978 & 0.7978 \\ \hline
    DP-SGDA & 0.6637 & 0.5773 & 0.7066 & 0.7542 \\ \hline
    DM-HSGD & 0.9351 & 0.9351 & 0.9351 & 0.9351 \\ \hline 
\rowcolor{greyL}
    \algns & 0.9048 & 0.9213 & 0.9356 & 0.9355 \\ 
    \hline
  \end{tabular}
  }
  \\
  \subfloat[Impact of $\gamma$.]
{
  \begin{tabular}{|c|c|c|c|c|}
    \hline
    $\gamma$ & $\gamma=\frac{1}{60000}$ & $\gamma=\frac{1}{30000}$ & $\gamma=\frac{1}{5000}$ & $\gamma=\frac{1}{1000}$ \\
    \hline
    \!SGDA \!& 0.7978 & 0.7978 & 0.7978 & 0.7978 \\ \hline
    \!DP-SGDA\! & 0.5795 & 0.5773 & 0.5732 & 0.5725 \\ \hline
    \!DM-HSGD \!& 0.9351 & 0.9351 & 0.9351 & 0.9351 \\ \hline
\rowcolor{greyL}
    \algns & 0.9206 & 0.9213 & 0.9237 & 0.9264 \\ \hline
  \end{tabular}
  }
  %\end{small}
 \end{normalsize}
  \label{table2}
\end{table}

\end{document}